\documentclass[12pt]{article}
\usepackage[margin=1in]{geometry}                
\usepackage{graphicx}
\usepackage{amssymb,amsmath,amsthm,amsfonts,bbm,bm,mathrsfs,xcolor}
\usepackage{mhequ}
\usepackage{mathabx}
\usepackage{comment}
\usepackage[numbers]{natbib}
\usepackage[colorlinks,citecolor=blue,urlcolor=blue,filecolor=blue,backref=page]{hyperref}
\usepackage{accents,url}
\usepackage[colorinlistoftodos,prependcaption,textsize=tiny]{todonotes}

\DeclareMathOperator*{\argmin}{arg\,min}

\DeclareMathOperator{\TV}{TV}

\def\m{\mathcal}
\def\mb{\mathbb}
\def\mr{\mathrm}

\def\ind{\mathbbm{1}}

\def\T{{\mathrm{\scriptscriptstyle T} }}
\def\D{{\mathrm{D} }}

\def\MF{{\mathrm{\scriptscriptstyle MF} }}
\def\DY{{\mathrm{\scriptscriptstyle DY} }}
\def\LV{{\mathrm{\scriptscriptstyle LV} }}

\newcommand{\blds}[1]{\mbox{\scriptsize \boldmath $#1$}}

\newcommand{\be}{\begin{equs}}
\newcommand{\ee}{\end{equs}}

\numberwithin{equation}{section}
\theoremstyle{plain}
\newtheorem{theorem}{Theorem}

\newtheorem{remark}{Remark}
\newtheorem{corollary}{Corollary}
\newtheorem{lemma}{Lemma}
\newtheorem{proposition}{Proposition}

\title{On the Convergence of Coordinate Ascent Variational Inference}
\author{Anirban Bhattacharya, Debdeep Pati and Yun Yang}
\date{\vspace{-1.5em}}                                           

\begin{document}
\maketitle

\begin{abstract}
    As a computational alternative to Markov chain Monte Carlo approaches, variational inference (VI) is becoming more and more popular for approximating intractable posterior distributions in large-scale Bayesian models due to its comparable efficacy and superior efficiency. Several recent works provide theoretical justifications of VI by proving its statistical optimality for parameter estimation under various settings; meanwhile, formal analysis on the algorithmic convergence aspects of VI is still largely lacking. In this paper, we consider the common coordinate ascent variational inference (CAVI) algorithm for implementing the mean-field (MF) VI towards optimizing a Kullback--Leibler divergence objective functional over the space of all factorized distributions. Focusing on the two-block case, we analyze the convergence of CAVI by leveraging the extensive toolbox from functional analysis and optimization. We provide general conditions for certifying global or local exponential convergence of CAVI. Specifically, a new notion of generalized correlation for characterizing the interaction between the constituting blocks in influencing the VI objective functional is introduced, which according to the theory, quantifies the algorithmic contraction rate of two-block CAVI. As illustrations, we apply the developed theory to a number of examples, and derive explicit problem-dependent upper bounds on the algorithmic contraction rate.
\end{abstract}


\section{Introduction}
Variational inference (VI,~\citep{jordan1999introduction,bishop2006pattern,wainwright2008graphical}) has now become a commonly used alternative to Markov chain Monte Carlo (MCMC) for approximating intractable posterior distributions in complicated Bayesian models with massive data due to its statistical efficiency and computational scalability. A number of recent works have investigated large-sample properties of VI, and proved its statistical optimality for point estimation in various settings. For example, see~\citep{ray2022variational,bickel2013asymptotic,hall2011theory,hall2011asymptotic,ormerod2012gaussian,titterington2006convergence} for theoretical results of VI applied to concrete models; and~\citep{alquier2020concentration,pati2018statistical,yang2020alpha,zhang2020convergence,wang2019frequentist} for general theoretical treatments of VI. 

Among various approximation schemes of VI, the mean-field (MF) approximation, which originates from statistical mechanics and uses the approximating family consisting of all factorized density functions over (blocks of) variables of interest, is the most widely used
and representative instance of VI.
A standard computational method for implementing MF is the coordinate ascent variational inference (CAVI) algorithm, which optimizes each component in the factorized density function while holding the others fixed. Refer to comprehensive review articles \citep{ormerod2010explaining,blei2017variational} and \citep[Chapter 10]{bishop2006pattern} for a book-level treatment. Depending on the execution order, a CAVI algorithm can be realized through either a sequential scheme where one component is updated at a time according to a deterministic order or a randomized order, or a parallel scheme where all components are simultaneously updated. Due to the lack of tractability of the updating formula involving unwieldy normalization constants and the technical challenge of dealing with optimization over infinite-dimensional distributions, a general theoretical treatment on analyzing the convergence of CAVI is missing in the literature. Study of the algorithmic landscape by explicitly analyzing dynamical systems in conditionally conjugate models contains both positive and negative results \citep{wang2006convergence,zhang2017theoretical,mukherjee2018mean,ghorbani2018instability,plummer2020dynamics}.

In this article, we aim to address the theoretical question of certifying the convergence of CAVI under general sufficient conditions, as well as quantifying its algorithmic contraction rate. 
Specifically, our analysis focuses on the two-block CAVI for technical simplicity, where the MF approximation family is composed of all distributions that can be factorized into the product of distributions over two pre-specified blocks of variables. Our novel contribution is to introduce a new notion of generalized correlation for characterizing the interaction between the constituting blocks in influencing the VI objective functional. We show that such a generalized correlation certifies the convergence of CAVI if sufficiently small and quantifies the algorithmic contraction rate. Our analysis leverages the extensive toolbox from functional analysis and optimization to avoid directly handling the normalization constants in the updating formula.  As an important implication of the derived contraction rate, CAVI tends to converge faster if each constituting component in the MF is less affected by the status of others when being individually optimized as in the CAVI algorithm. We apply the developed theory to a number of concrete Bayesian models. By examining the sharpness of the implied results in these examples, we find that our sufficient conditions based on the generalized correlation precisely captures the algorithmic hardness in the two-block MF variational approximation.

The rest of this article is organized as follows. In Section~\ref{sec:background}, we introduce some necessary background; briefly review the mean-field VI, the CAVI algorithm, and some results from constrained optimization in function spaces. In Section~\ref{sec:conv_two_block}, we present our main theoretical results on the convergence of CAVI with two blocks. In Section~\ref{sec:examples}, we apply the developed theory to a number of representative Bayesian models. Section \ref{sec:gtr_2b} offers a preliminary extension to CAVI with more than 2 blocks. Proofs of main results are provided in Section~\ref{sec:proofs}; proofs related to the applications and some other technical results are deferred to the appendices. Section~\ref{sec:discussion} concludes the article with some discussions.

\section{Preliminary Background}\label{sec:background}
We begin this section with introducing notation and recalling some useful probability identities and inequalities related to the Kullback–Leibler (KL) divergence. After that, we briefly review key elements of mean-field variational inference and the associated coordinate ascent variational inference (CAVI) algorithm along with some of its variants. We end this section with some useful results from constrainted optimization in function spaces that will be used to analyze the one-step contraction of CAVI.

\subsection{Notation}
Given probability measures $Q$ and $P$ on $\mb R^d$ with $Q \ll P$ and $f = dQ/dP$ the Radon--Nikodym derivative, the Kullback--Leibler (KL) divergence  from $Q$ to $P$ is defined as
\begin{align}\label{eq:KL_def}
D_{\rm KL}(Q\,||\,P) :\,= 
\begin{cases}
\int f \log(f) \, dP & \text{ when } f \log_+(f) \in L_1(P), \\
+\infty & \text { otherwise.}
\end{cases}
\end{align}
We record the following useful variational characterization of the KL divergence; see, for example, Corollary 4.15 of \cite{boucheron2013concentration}:
\begin{align}\label{eq:KL_var_char}
D_{\rm KL}(Q||P) = \sup_{Z}\big\{\mb E_Q[Z] - \log \mb E_P[e^Z]\big\},
\end{align}
where the supremum is taken over all random variable $Z$ such that $\mb E_P[e^Z]<\infty$.

If $P$ and $Q$ have densities $p$ and $q$ with respect to a dominating measure, we shall interchangeably use $D_{\rm KL}(Q\,||\,P)$ and $D_{\rm KL}(q\,||\,p)$. We shall throughout use the convention of identifying densities $p$ and $q$ with $D_{\rm KL}(q\,||\,p) = 0$. 
For $\alpha \in [0, 1]$, define a family of weighted divergences 
\begin{align}\label{eq:wt_KL_def}
D_{\rm KL, \alpha}(q \,|| \, p) = \alpha D_{\rm KL}(q \,||\, p) + (1-\alpha) D_{\rm KL}(p \,|| \, q). \end{align}
When $\alpha = 1$ (resp. $0$), this weighted divergence equals $D_{\rm KL}(q\,||\,p)$ (resp. $D_{\rm KL}(p\,||\,q)$), and $\alpha = 1/2$ corresponds to (half of) the symmetrized KL divergence \citep{kullback1951information}. Observe that $D_{\rm KL, \alpha}(p \,||\, q) = D_{\rm KL, 1-\alpha}(q \,||\, p)$, and $D_{\rm KL, \alpha}(q \,||\, p) + D_{\rm KL, 1-\alpha}(q \,||\, p) = 2 D_{\rm KL, 1/2}(q \,||\, p)$. 

A real random variable $Z$ is called sub-Gaussian if there exists $\nu > 0$ such that $\mb E [e^{t(Z-\mb E(Z))}] \le e^{\nu^2 t^2/2}$ for all $t \in \mb R$. The smallest such $\nu$ is called the sub-Gaussian norm (or Orlicz norm with Orlicz function $\psi_2(x)=e^{x^2}-1$) and is denoted as $\|Z\|_{\psi_2}$. We record the following transport inequality that we use frequently to control the difference in the expectations of two distributions via their KL divergence. 
\begin{lemma}[Transportation cost inequality]\label{lem:tce}
If $Z$ is sub-Gaussian under probability measure $P$ on $\mb R$, i.e.~$\mb E_{P} [e^{t(Z-\mb E_{P} [Z])}]\leq e^{\nu^2 t^2/2}$ for all $t\in\mb R$, then for any other probability measure $Q \ll P$ on $\mb R$, we have
\begin{align*}
\big| \mb E_{Q} [Z] -  \mb E_{P} [Z]\big| \leq \sqrt{2\nu^2 D_{\rm KL}(Q||P)}.
\end{align*}
\end{lemma}
The bound on $\mb E_{Q} [Z] -  \mb E_{P} [Z]$ is from Lemma 4.18 of \cite{boucheron2013concentration}. To get the other inequality, simply note that if $Z$ is sub-Gaussian with respect to $P$, so is $-Z$, and $\|Z\|_{\psi_2} = \|-Z\|_{\psi_2}$.

For $\mu \in \mb R^d$ and $\Sigma$ a $d \times d$ positive definite matrix, we use $N_d(\mu, \Sigma)$ to denote a multivariate normal distribution with mean $\mu$ and covariance matrix $\Sigma$. We also use $N_d(x; \mu, \Sigma)$ to denote the corresponding probability density function evaluated at $x \in \mb R^d$. Unless otherwise specified, $\|x\|$ denotes the Euclidean norm of a vector $x$, and $\|A\|_2$ denotes the operator norm of matrix $A$, i.e., its largest singular value. 

\subsection{Mean-field variational inference and CAVI algorithm}
Let $\pi_n$ denote our target distribution defined on set $\m X \subseteq \mb R^m$. With a slight abuse of notation, we shall continue to use $\pi_n$ to denote its density with respect to a dominating measure, which for our purpose will be either the Lebesgue measure or the counting measure in this article. In Bayesian statistics, $\pi_n$ corresponds to a posterior distribution, $\pi_n(u) \,\propto\, e^{-U_n(u)} \, \pi(u)$, where $U_n$ is the negative log-likelihood function with $n$ data points and $\pi$ the prior distribution on the parameter space $\m X$. More generally, we shall continue to write $\pi_n(u) = e^{-V_n(u)}/Z_n$ in a Gibbs distribution form, with the subscript $n$ indicating some evolving aspect of the distribution, e.g., the sequence of distributions $\{\pi_n\}$ gets increasingly concentrated as $n$ increases. In the Bayesian example, we have $V_n(u) = U_n(u) + \log \pi(u)$ and the normalizing constant $Z_n$ is the marginal likelihood or evidence.

Mean-field variational inference \cite[Chapter 10]{bishop2006pattern} aims to approximate $\pi_n$ with a product density. Specifically, suppose $\m X = \m X_1 \times \ldots \times \m X_d$ with $\m X_j \subseteq \mb R^{m_j}$ and $\sum_{j=1}^d m_j = m$. Let 
\begin{align*}
\m Q_{\MF} :\,= \big\{q = q_1 \otimes \ldots \otimes q_d \, : \, q \ll \pi_n \text{ and } D_{\rm KL}(q \,||\, \pi_n) < \infty \big\}
\end{align*}
denote the class of probability densities absolutely continuous with respect to the target density $\pi_n$ which decompose into a product form, with $q_j$ a probability density on $\m X_j$ for each $j \in [d]$. Mean-field variational inference minimizes some notion of distance or divergence between $q \in \m Q_{\MF}$ and $\pi_n$ to arrive at a best proxy for $\pi_n$ from the mean-field family $\m Q_{\MF}$. The Kullback--Leibler (KL) divergence $D_{\rm KL}(q \,||\, \pi_n)$ has been traditionally used as a measure of discrepancy, with its origin in statistical physics \cite{parisi1988statistical}, and continues to remain one of the popular choices. In recent years, a number of other divergence or distance measures have been considered in the literature, such as R\'{e}nyi divergence~\citep{li2016renyi}, reverse KL divergence~\citep{minka2013expectation}, and Wasserstein distance~\citep{ambrogioni2018wasserstein}, among many others.

The coordinate ascent variational inference (CAVI) algorithm is one of the most commonly used iterative procedures for the optimization problem $\min D_{\rm KL}(q_1 \otimes \ldots \otimes q_d \,||\, \pi_n)$. CAVI is an alternating minimization algorithm which iteratively updates a single coordinate\footnote{we shall henceforth refer to $q_j$ as the $j$th coordinate of $q = q_1 \otimes \ldots \otimes q_d$, although $q_j$ itself could be a multivariable density} $q_j$ at each step, keeping the remaining coordinates fixed. We shall consider several variants of the CAVI, and we introduce some notation to describe them succinctly. Let $F(q) :\,=  D_{\rm KL}(q \,||\, \pi_n)$ denote our objective function. For $q \in \m Q_{\MF}$ and any $j \in [d]$, let $q_{-j}$ denote the product density $\otimes_{k \ne j} q_k$ defined on $\bigtimes_{k \ne j} \m X_k$. We shall write $F(q)$ equivalently as $F(q_j \otimes q_{-j})$ to emphasize the dependence on the $j$th coordinate. Similarly, also define $q_{<j} = \otimes_{k < j} q_k$ and $q_{>j} = \otimes_{k > j} q_k$, and write $F(q) = F(q_{<j} \otimes q_j \otimes q_{>j})$; in the corner cases $j \in \{1, d\}$, only two of the three terms remain.

\begin{figure}[h!]
\centering
\includegraphics[width=0.70\textwidth]{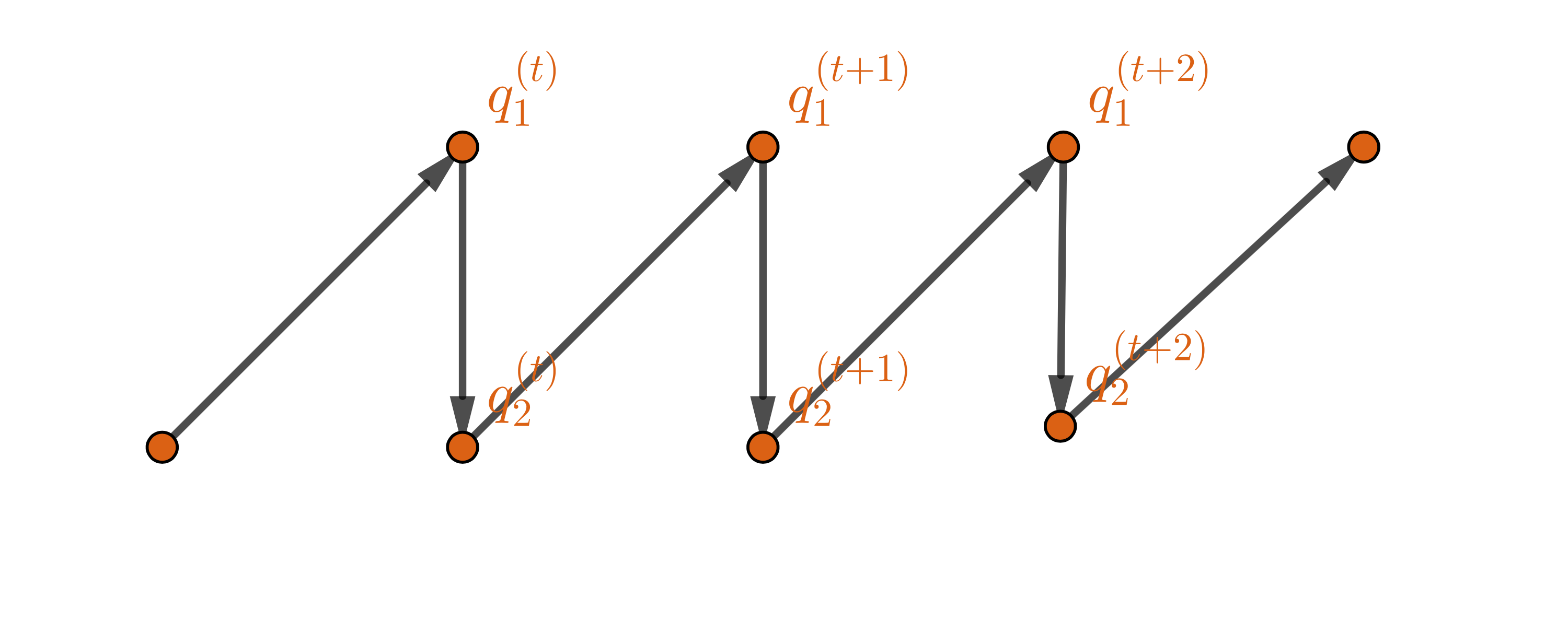} 
\vspace{-0.5in}
\caption{{\em Sequential dynamics in 2d with $q_1$ updated first, i.e., $q_1^{(t+1)} = \argmin_{q_1} F(q_1 \otimes q_2^{(t)})$ and $q_2^{(t+1)} = \argmin_{q_2} F(q_1^{(t+1)} \otimes q_2)$.}}\label{fig:seq}
\end{figure}

A commonly employed version is the sequential CAVI, which proceeds for $t=0,1,2,\ldots$ as 
\be\label{eq:seq_CAVI_gen}
q_j^{(t+1)} = \argmin_{q_j} F(q_{< j}^{(t+1)} \otimes q_j \otimes q_{>j}^{(t)}), \quad j \in [d]. 
\ee
A single pass of the sequential CAVI going from $q^{(t)} \rightarrow q^{(t+1)}$ consists of $d$ intermediate steps, updating the coordinates one-at-a-time (in a pre-fixed order), 
{\small\be
(q_1^{(t)}, q_2^{(t)}, \ldots, q_d^{(t)}) \rightarrow (q_1^{(t+1)}, q_2^{(t)}, \ldots, q_d^{(t)}) \rightarrow  (q_1^{(t+1)}, q_2^{(t+1)}, \ldots, q_d^{(t)}) \rightarrow \cdots \rightarrow (q_1^{(t+1)}, q_2^{(t+1)}, \ldots, q_d^{(t+1)}). 
\ee}
Figure \ref{fig:seq} shows a schematic of the sequential CAVI for two blocks, i.e., $d = 2$. 

\begin{figure}[h!]
\centering
\includegraphics[width=0.60\textwidth]{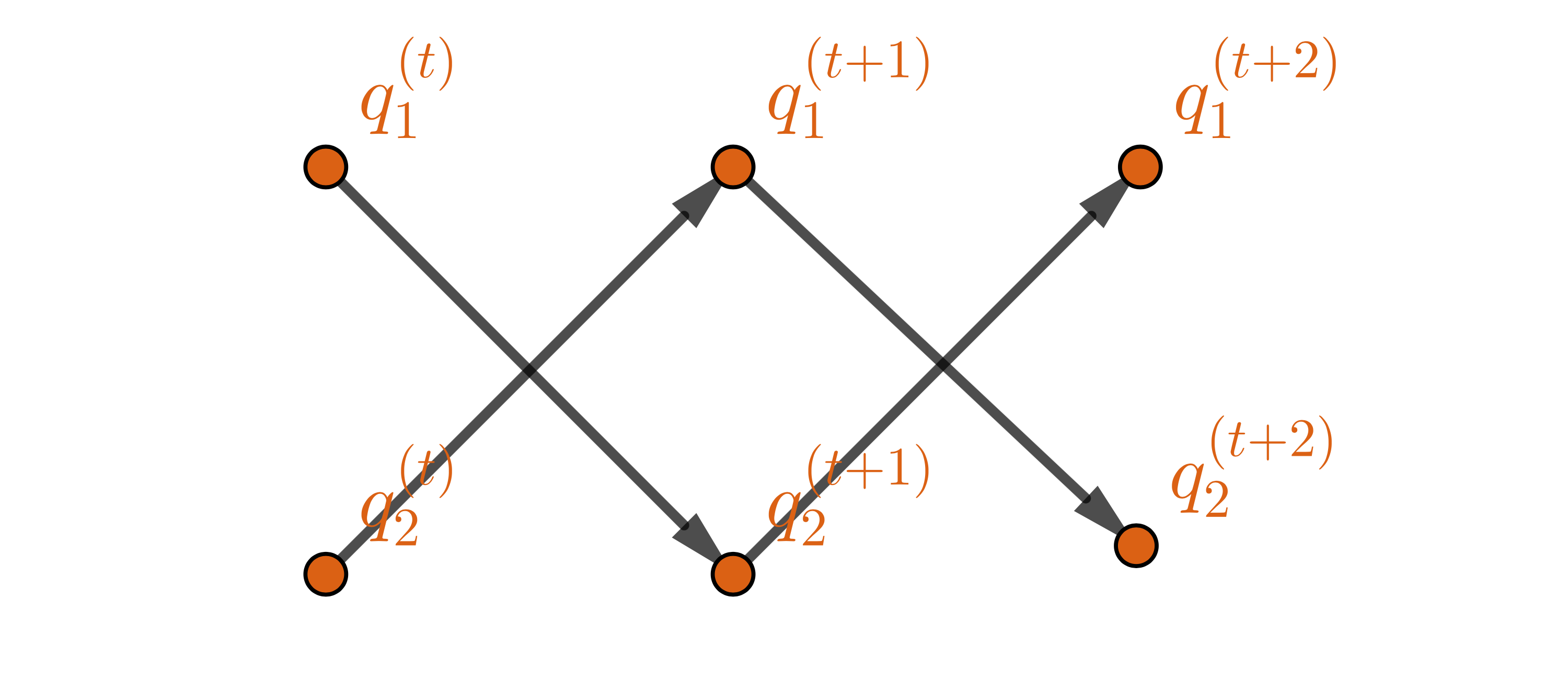} 
\vspace{-0.3in}
\caption{{\em Parallel dynamics in 2d with $q_1^{(t+1)} = \argmin_{q_1} F(q_1 \otimes q_2^{(t)})$ and $q_2^{(t+1)} = \argmin_{q_2} F(q_1^{(t)} \otimes q_2)$.}}\label{fig:par}
\end{figure}

We also consider a parallel version of the CAVI algorithm which proceeds as 
\be\label{eq:par_CAVI_gen}
q_j^{(t+1)} = \argmin_{q_j} F(q_j \otimes q_{-j}^{(t)}), \quad j \in [d]. 
\ee
Figure \ref{fig:par} shows a schematic diagram of the parallel update for $d = 2$. 

Finally, we also analyze a randomized algorithm where at each pass, a coordinate is selected at random and updated, keeping the other coordinates unchanged. Specifically, at time $t+1$, pick coordinate $h$ uniformly at random from $[d]$, and set
\be\label{eq:seq_CAVI_gen_rand}
q_h^{(t+1)} = \argmin_{q_h} F(q_h \otimes q_{-h}^{(t)}), \quad q_k^{(t+1)} = q_k^{(t)} \text{ for } k \ne h. 
\ee

Irrespective of the version employed, one needs to solve the optimization problem $\argmin_{\rho_j} F(\rho_j \otimes q_{-j})$ at each step. This sub-problem is convex with a unique solution 
\be \label{eq:cavi_gen_form}
q_j \,\propto\, \exp \bigg(\int_{\m X_{-j}} q_{-j} \, \log \pi_n \bigg). 
\ee
In conditionally conjugate models, this generally leads to tractable updates without any further restrictions on the $q_k$'s; see \cite[Chapter 10]{bishop2006pattern} for numerous examples.

\subsection{Constrained optimization in function spaces}\label{sec:cons_fn}
We recall some facts about constrained convex optimization of functionals \cite{peypouquet2015convex}. 
Recall that a functional $F: W \to \mb R$ on a normed linear space $(W, \|\cdot\|)$ is called Fr{\'e}chet differentiable at $w \in W$ if there exists a bounded linear operator $\D F(w): W \to \mb R$ such that 
\be \label{eq:Frechet}
\lim_{\|h\| \to 0} \frac{ |F(w+h) - F(w) - \D F(w)[h]|}{\|h\|} = 0.
\ee
The linear map $\D F(w)$ is called the Fr{\'e}chet derivative at $w$. When the Fr{\'e}chet derivative exists at $w$, one has 
\be\label{eq:Gateaux}
\D F(w)[h] = \lim_{t \to 0} \frac{F(w + t h) - F(w)}{t}, \quad h \in W. 
\ee
The quantity in the right hand side is called the G{\^a}teaux derivative of $F$ at $w$ in the direction $h$.  The G{\^a}teaux derivative generalizes the notion of a directional derivative to normed linear spaces. 

For a functional $F$ which is convex and Fr{\'e}chet differentiable, consider its restriction to a convex sub-domain $W_c \subset W$. The {\em first order condition} for $v \in W_c$ to be a minimum point of $F$ over $W_c$ is 
\be \label{eq:foc}
\D F(v)[w-v] \ge 0, \quad w \in W_c,
\ee
or equivalently, 
\be \label{eq:foc2}
\D F(v)[h] \ge 0, \quad h \in T_{W_c}(v),
\ee
where $T_{W_c}(v) = \{w-v:\, w\in W_c\}$ is the tangent cone of $W_c$ at $v$.
This immediately follows using the convexity of $F$; for any $w \in W_c$, we have $F(w) \ge F(v) + \D F(v)[w-v]$, and the above condition implies $F(v) \le F(w)$ for any $w \in W_c$. 
When $W_c \subset W = \mb R^d$, the condition \eqref{eq:foc2} reduces to the familiar first order condition of optimality $h'\nabla F(v) \ge 0$ for all $h \in T_{W_c}(v)$. More generally, if $W$ is a Hilbert space, then by the Riesz representation theorem, there exists a unique 
element $\dot{F}(v) \in W$ such that $\D F(v)[h] = \langle \dot{F}(v), h \rangle$ for all $h \in W$. Thus, the first order condition \eqref{eq:foc2} can again be written as $\langle \dot{F}(v), h \rangle \ge 0$ for all $h \in T_{W_c}(v)$. 

To see the relevance in the present context, let $\m Q$ denote the collection of probability densities $q$ on $\m X$ absolutely continuous with respect to $\pi_n$ such that $D_{\rm KL}(q \,||\, \pi_n) < \infty$. We shall view $\m Q$ as a subset of $L^2(\m X)$ equipped with the usual $L^2$ inner product $\langle f, g \rangle = \int_{\m X} f g$. Since $D_{\rm KL}(\alpha q + (1-\alpha) \tilde{q}\,||\, \pi_n) \leq  \alpha D_{\rm KL}(q \,||\, \pi_n) + (1- \alpha) D_{\rm KL}(\tilde{q} \,||\, \pi_n)$ by convexity of the KL divergence \cite{van2014renyi}, $\m Q$ is a convex subset of $L^2(\m X)$. Define $F(f) = \int f \log f/\pi_n$ so that the restriction of $F$ to $\m Q$ is $D_{\rm KL}(\cdot\,||\, \pi_n)$. For any $q\in \m Q$, the tangent cone of $\m Q$ at $q$ is $T_{\m Q}(q) = \{t(p-q):\, p\in\m Q,\, t\in[0,1]\}$.
It is straightforward to check (see Appendix \ref{sec:pf_app_aux} for details) that for any $q \in \m Q$ and $h \in T_{\m Q}(q)\subset L^2(\m X)$, 
\be \label{eq:KL_frechet}
\D F(q)[h] = \langle \log(q/\pi_n), h \rangle = \int h \log (q/\pi_n). 
\ee
Thus, in our earlier notation, we may take $\dot{F}(q) = \log(q/\pi_n)$. $\dot{F}(q)$ is also called the first variation of functional $F(\cdot)$ at $q$, and it is worth mentioning that $\dot{F}(q)$ is uniquely defined up to a constant shift: for any constant $C\in\mb R$, $\log(q/\pi_n)+C$ is also a valid first variation since $\int h=0$ for any $h \in T_{\m Q}(q)$. This arbitrary constant does not interfere with any of our subsequent analysis as it always gets annihilated, and we thus set it to a default value of zero while deriving first variations. 

The above in particular implies\footnote{As noted above, the constant $C$ disappears below since $\int C (q - \tilde{q}) = 0$ for any $q, \tilde{q} \in \m Q$.}
\begin{align}\label{eq:conv_F}
\begin{aligned}
F(q) - F(\tilde{q}) - \langle \dot{F}(\tilde{q}), q - \tilde{q} \rangle
&= \int q \log \frac{q}{\pi_n} - \int \tilde{q} \log \frac{\tilde{q}}{\pi_n} - \int (q - \tilde{q})  \log \frac{\tilde{q}}{\pi_n}  \\
&= D_{\rm KL}(q \,||\, \tilde{q}),
\end{aligned}
\end{align}
for any $q, \tilde{q} \in \m Q$; a fact we shall crucially exploit in the sequel. 

\section{Convergence analysis: two block case}\label{sec:conv_two_block}
We begin with a two-block decomposition of $q$, i.e., $d = 2$ and $q = q_1 \otimes q_2$. We first present some preliminary results in Section~\ref{se:2b_pre} describing the geometry of the KL objective function $D_{\rm KL}(q_1 \otimes q_2 \,||\, \pi_n)$ under the mean-field family relative to the KL divergence disrepancy measure. Then, motivated by a local ``convexity" type decomposition due to this geometry, we define an important quantity, called generalized correlation, to capture the intervention between the two blocks in the parallel CAVI, which further leads to our main result about the global convergence of the algorithm. In Section~\ref{sec:2b_ext}, we discuss extensions of the theorem to: 1.~local convergence; 2.~randomized CAVI; 3.~sequential CAVI.  In Section~\ref{sec:examples} to folloow, we illustrate the application of the theory to a number of examples. 

\subsection{Preliminaries}\label{se:2b_pre}
Let $q_1^\star \otimes q_2^\star$ denote a global minimum of the variational objective function, i.e.,  
\be \label{eq:CAVI_sol}
F(q_1^\star \otimes q_2^\star) = \min_{q_1 \otimes q_2 \in \m Q_{\MF}} F(q_1 \otimes q_2), \quad F(q_1 \otimes q_2) = D_{\rm KL}(q_1 \otimes q_2 \,||\, \pi_n).
\ee
Our theoretical result below provides a sufficient condition to guarantee the uniqueness of a global minima as a byproduct; see Remark~\ref{rem:global_opt}. 
We record some properties of $q_1^\star \otimes q_2^\star$ in Proposition \ref{prop:LSC} below. 
\begin{lemma}\label{prop:LSC}
The optimality of $q_1^\star$ and $q_2^\star$ implies
\be \label{eq:CAVI_opt}
q_j^\star = (Z_j^\star)^{-1}\exp\Big\{\int q_{-j}^\star \log\pi_n\Big\}, \ Z_j^\star =\int_{\m X_j} \exp\Big\{\int q_{-j}^\star \log\pi_n\Big\}, \ j=1,2.
\ee
For any $q_1 \otimes q_2 \in \m Q_{\MF}$, this gives, 
\begin{align}
F(q_1 \otimes q_2) - F(q_1^\star \otimes q_2^\star) 
=D_{\rm KL}(q_1\,||\,q_1^\star) + D_{\rm KL}(q_2\,||\,q_2^\star) 
- \Delta_n(q_1, q_2), \label{Eqn:LSC}
\end{align}
where 
\be \label{eq:Deltaq}
\Delta_n(q_1, q_2) = \int (q_1-q_1^\star)\otimes(q_2- q_2^\star )\log \pi_n.
\ee
In particular, it immediately follows that 
\begin{align}\label{Eqn:LSC_spec}
\begin{aligned}
F(q_1 \otimes q_2^\star) - F(q_1^\star \otimes q_2^\star) &= D_{\rm KL}(q_1\,||\,q_1^\star), 
\\
F(q_1^\star \otimes q_2) - F(q_1^\star \otimes q_2^\star) &= D_{\rm KL}(q_2\,||\,q_2^\star).
\end{aligned}
\end{align}
\end{lemma}
The interaction term $\Delta_n(q_1, q_2)$ reflects the degree of dependence between 
the two blocks
under the joint distribution $\pi_n(u_1,u_2)$. For example, in the extreme case when $\pi_n$ can be factorized into $q_1^\star\otimes q_2^\star$, we have $\Delta_n(q_1, q_2)\equiv 0$. In addition, since then $F(q_1\otimes q_2) = D_{\rm KL}(q_1\,||\,q_1^\star) + D_{\rm KL}(q_2\,||\,q_2^\star)$,
both the sequential and the parallel versions of the CAVI boil down to the independent updating rules on $q_1$ and $q_2$. In general, if we can decompose $\log\pi_n(u_1,u_2) = V_{n,1}(u_1) + V_{n,2}(u_2) + V_{n,12}(u_1,u_2)$ into two marginal terms plus an interaction term characterizing the dependence, then only the interaction term contributes to the overall integral:
\be \label{eq:int_survives}
\Delta_n(q_1, q_2) = - \int V_{n, 12}(\theta_1, \theta_2)[q_1(\theta_1) - q_1^\star(\theta_1)] \, [q_2(\theta_2) - q_2^\star(\theta_2)] d\theta_1d\theta_2. 
\ee
In particular, $\Delta_n$ does not depend on the normalizing constant of the target density, which makes it practicable to compute this integral in applications. We explicitly compute $\Delta_n$ along the solution path for a variety of statistical models in the sequel. 

In the CAVI updates, $\Delta_n(q_1, q_2)$ elucidates how sensitive the time $(t+1)$ iterate $q_j^{(t+1)}$ is to the time $t$ iterate $q_{-j}^{(t)}$ for $j=1,2$. In other words, a small $\Delta_n(q_1,q_2)$ along the optimization solution path implies that the $j$-th ($j=1,2$) coordinate $\{q_j^{(t)}:\,t\geq 0\}$ is close to the ideal convex optimization problem of 
\be\label{eq:ideal_op}
\min_{q_j} F(q_j \otimes q_{-j}^\star),
\ee
where $q_{-j}$ is fixed at its global solution $q_{-j}^\star$. 
Consequently, appropriate control on the magnitude of $\Delta_n(q_1, q_2)$ plays a key role in our analysis to quantify how the actual solution path $\{q_j^{(t)}:\,t\geq 0\}$
deviates from the solution path of the idealized problem~\eqref{eq:ideal_op}. 
Following this high-level intuition, we prepare to state our first convergence result. 
We consider the parallel CAVI algorithm \eqref{eq:par_CAVI_gen} first, and later generalize to the sequential and randomized updating schemes. It is worthy mentioning that the convergence rate derived based on this interaction term $\Delta_n(q_1,q_2)$ is sharp for Gaussian targets; see Section~\ref{sec:2b_examples}.

\subsection{Global convergence}\label{sec:global_conv}
We begin by setting up some notation. Recall the weighted KL divergence $D_{\rm KL, \alpha}(p\,||\,q)$ from \eqref{eq:wt_KL_def}. For any $\alpha\in[0,1]$, we will call the following quantity as the $\alpha$-generalized correlation within $\pi_n$ with respect to the decomposition $\m X = \m X_1 \times \m X_2$ over families $\m Q_1$ and $\m Q_2$:
\be\label{eq:Gcorr_alpha}
\mbox{\rm GCorr}_\alpha(\pi_n) = \sup_{q_j \in \m Q_j \backslash \{q_j^\star\}} \frac{|\Delta_n(q_1,\,q_2)|}{\sqrt{D_{\rm KL, \alpha}(q_1 \,||\, q_1^\star) }\, \sqrt{D_{\rm KL, 1-\alpha}(q_2\,||\, q_2^\star) }}.
\ee
We also define the associated generalized correlation as
\be \label{eq:GCorr}
\mbox{\rm GCorr}(\pi_n) = \inf_{\alpha\in[0,1]} \bar{\gamma}_{n, \alpha}, \quad \bar{\gamma}_{n,\alpha} = \max \big\{ \mbox{\rm GCorr}_\alpha(\pi_n), \mbox{\rm GCorr}_{1-\alpha}(\pi_n) \big\}.
\ee
We have the necessary ingredients to state our first convergence result below. 
\begin{theorem}[Two-block parallel CAVI]\label{thm:CAVI_2block_par_alpha}
Suppose the target density $\pi_n$ satisfies $\mbox{\rm GCorr}(\pi_n) \in (0, 2)$.
Then, for any initialization $q^{(0)} = q_1^{(0)} \otimes q_2^{(0)} \in \m Q$ of the parallel {\rm CAVI} algorithm, one has a contraction
\be \label{eq:par_cavi_exp_cgence}
D_{\rm KL,1/2}(q^{(t+1)}\,||\,q^\star)  \le \kappa_n D_{\rm KL, 1/2}(q^{(t)}\,||\,q^\star),
\ee
for any $t \ge 0$, where the contraction constant $\kappa_n :\,= \mbox{\rm GCorr}^2(\pi_n)/4 \in (0, 1)$.
\end{theorem}
Iterating \eqref{eq:par_cavi_exp_cgence}, we have for any $t \ge 1$ that $D_{\rm KL, 1/2}(q^{(t)}\,||\,q^\star) \le \kappa_n^t D_{\rm KL, 1/2}(q^{(0)}\,||\,q^\star)$, implying exponential convergence in the symmetrized KL divergence. This in particular implies a unique global minima, explicated in the following remark. 

\begin{remark}\label{rem:global_opt}
In Theorem \ref{thm:CAVI_2block_par_alpha}, $q^\star = q_1^\star \otimes q_2^\star$ is the unique global minima of $F$ within $\m Q$. To see this, suppose $\tilde{q}^\star = \tilde{q}_1^\star \otimes \tilde{q}_2^\star$ is a global minimum of $F$ in $\m Q$ distinct from $q^\star$. Then, initializing the algorithm at $\tilde{q}^\star$, one has $q^{(t)} = \tilde{q}^\star$ for all $t$ from the stationarity condition \eqref{eq:CAVI_opt}. Substituting in \eqref{eq:par_cavi_exp_cgence}, this implies $D_{\rm KL, 1/2}(\tilde{q}^\star\,||\,q^\star) = 0$, arriving at a contradiction. 
\end{remark}

Verifying the condition $\mbox{\rm GCorr}(\pi_n) \in (0, 2)$ entails demonstrating $\alpha \in [0, 1]$ such that $\bar{\gamma}_{n, \alpha} \in (0, 2)$. In particular, for $\alpha=1/2$, $\bar{\gamma}_{n, \alpha} \in (0, 2)$ is equivalent to showing that there exists a constant $\tilde{\gamma}_n \in (0,2)$ such that
\begin{align}
|\Delta_n(q_1, q_2)| \le \tilde{\gamma}_n \sqrt{ D_{\rm KL,1/2}(q_1^\star\,||\, q_1) } \, \sqrt{ D_{\rm KL, 1/2}(q_2^\star\,||\, q_2) }, \label{eq:Gcorr_alpha_half}
\end{align}
for all $q_j \in \m Q_j, j = 1, 2$. Similarly, for $\alpha = 0$ or $1$, it amounts to showing the existence of $\gamma_n \in (0, 2)$ such that 
\begin{align}\label{eq:Gcorr_alpha_1}
\begin{aligned}
|\Delta_n(q_1, q_2)| & \le \gamma_n \sqrt{D_{\rm KL}(q_1^\star\,||\, q_1)} \sqrt{D_{\rm KL}(q_2 \,||\, q_2^\star)}, \\
|\Delta_n(q_1, q_2)| & \le \gamma_n \sqrt{D_{\rm KL}(q_1\,||\, q_1^\star)} \sqrt{D_{\rm KL}(q_2^\star \,||\, q_2)},
\end{aligned}
\end{align}
for all $q_j \in \m Q_j, j = 1, 2$. We shall henceforth refer to \eqref{eq:Gcorr_alpha_half} and \eqref{eq:Gcorr_alpha_1} as the {\em symmetric} and {\em asymmetric} KL condition, respectively. 

\begin{remark}\label{rem:restrict}
For most common applications of CAVI, the iterates $\{q_j^{(t)}:\,t\geq 1\}$ ($j=1,2$) lie in some exponential family $\bar{\m Q}_j \subset \m Q_j$ because of the natural conditional conjugacy, under mild conditions on the initialization. For example, if the target distribution $\pi_n(\theta_1, \theta_2)$ is bivariate Gaussian, and $q(\theta) = q_1(\theta_1) q_2(\theta_2)$, then $q_j^{(t)}$ lie in the univariate Gaussian family whenever the algorithm is initialized at any $q^{(0)} = q_1^{(0)} \otimes q_2^{(0)}$ with $q_j^{(0)}, j=1, 2$ having finite second moments. In such situations, we may restrict the supremum in the definition of $\mbox{\rm GCorr}_\alpha(\pi_n)$ to $\bar{\m Q}_j \backslash \{q_j^\star\} \, (j = 1, 2)$. Then, the conclusion of Theorem \ref{thm:CAVI_2block_par_alpha} continues to hold with the modified definition of $\mbox{\rm GCorr}(\pi_n)$ as long as the algorithm is initialized inside $\bar{\m Q}_j$. 
\end{remark}
Remark \ref{rem:restrict} immediately follows from the proof of Theorem \ref{thm:CAVI_2block_par_alpha} as it suffices to bound $|\Delta_n(q_1^{(r)}, q_2^{(t)})|$ for $|r-t| = 1$ inside the proof. 
The utility of Remark \ref{rem:restrict} is that $D_{\rm KL, \alpha}(q_j\,||\,q_j^\star)$ can be explicitly computed when $q_j, q_j^\star \in \bar{\m Q}_j$ using standard properties of exponential families. Moreover, $\Delta_n(q_1, q_2)$ can be conveniently expressed in terms of the parameters of the exponential family. 

\subsection{Local Convergence}\label{sec:2b_ext}
In general, we cannot expect global convergence as in Theorem \ref{thm:CAVI_2block_par_alpha}.
For example, due to the label switching issue in Gaussian mixture models, global convergence cannot be generally guaranteed, and one can at best expect local convergence to the closest global minimum to the initialization. 
We formalize this intuition in the following local convergence result. 
\begin{theorem}[Two-block parallel CAVI: local contraction]\label{thm:CAVI_2block_par_alpha_init}
In the definition \eqref{eq:Gcorr_alpha} of $\mbox{\rm GCorr}_\alpha(\pi_n)$, replace $\m Q_j$ by $\m Q_j^\star(r_0) :\, = \{q_j \in \m Q_j : D_{\rm KL}(q_j^\star \,||\, q_j) \le r_0\}$ for $r_0 > 0$, and call the resulting quantity $\mbox{\rm GCorr}_\alpha(\pi_n; r_0)$. Using $\mbox{\rm GCorr}_\alpha(\pi_n; r_0)$ instead of $\mbox{\rm GCorr}_\alpha(\pi_n)$ in \eqref{eq:GCorr}, define the restricted generalized correlation $\mbox{\rm GCorr}(\pi_n; r_0)$. 

Suppose there exists $r_0 > 0$ such that $\mbox{\rm GCorr}(\pi_n; r_0) \in (0, 2)$. Assume that the initialization satisfies $D_{\rm KL,1/2}(q_j^{(0)}\,||\,q_j^\star) \le r_0/2$ for $j = 1,2$. Then, for any $t \ge 0$,
\be \label{eq:par_cavi_exp_cgence_local}
D_{\rm KL,1/2}(q^{(t+1)}\,||\,q^\star)  \le \kappa_n D_{\rm KL, 1/2}(q^{(t)}\,||\,q^\star),
\ee
with $\kappa_n :\,= \mbox{\rm GCorr}^2(\pi_n; r_0)/4 \in (0, 1)$. 
\end{theorem}
Iterating \eqref{eq:par_cavi_exp_cgence_local}, one gets exponential convergence to $q^\star$ as before. A crucial component of Theorem \ref{thm:CAVI_2block_par_alpha_init} is the interplay between the choice of the local neighborhood $\m Q_j^\star(r_0)$ around $q_j^\star$ to define the restricted generalized correlation $\mbox{\rm GCorr}(\pi_n; r_0)$, and the initialization neighborhood $\m Q_j^{init} :\,= \{q_j \in \m Q_j : D_{\rm KL,1/2}(q_j\,||\,q_j^\star) \le r_0/2\}$. We show in the proof that if $q_j^{(t)} \in Q_j^{init}, (j = 1, 2)$ for any $t \ge 0$, then $q_j^{(t+1)} \in \m Q_j^\star(r_0), (j = 1, 2)$. This provides control over $|\Delta_n(q_1^{(r)}, q_2^{(t)})|$ for $|r-t| = 1$, which is crucially used inside the proof. As a parallel to Remark \ref{rem:global_opt}, the conclusion of the theorem forces $q^\star$ to be the unique global minimizer of $F$ within $\m Q_1^{init} \otimes \m Q_2^{init}$. 
\begin{remark}\label{rem:restrict_init}
As in Remark \ref{rem:restrict}, when the updates for $q_j$ lie inside an exponential family $\bar{\m Q}_j$, one may replace $\m Q_j^\star(r_0)$ with $\widebar{\m Q^\star}_j(r_0) :\,= \m Q_j^\star(r_0) \bigcap \bar{\m Q}_j$. Specifically, if $\bar{\m Q}_j$ is an exponential family, and $q_j \in \bar{\m Q}_j$ has natural parameter $\psi_j$, then $\widebar{\m Q^\star}_j(r_0)$ can be conveniently expressed as $\{d(\psi_j, \psi_j^\star) \le r_0\}$, where $d(\psi_j, \psi_j^\star) = D_{\rm KL}\big(q(\cdot \mid \psi_j^\star) \,||\, q(\cdot \mid \psi_j)\big)$.
\end{remark}
\begin{remark}
    In some applications of Theorem~\ref{thm:CAVI_2block_par_alpha_init} (see Section~\ref{subsec:local}), we can only control the restricted generalized correlation on a high-probability set under the true data generating model where the mean-field approximation $q_1^\star$, $q_2^\star$ have good concentration properties when sample size is large; see the related works mentioned in the introduction for concrete large-sample properties of VI. Then, the conclusion of Theorem~\ref{thm:CAVI_2block_par_alpha_init} remains valid as long as we are conditioning on this high-probability set at the beginning of the analysis.
\end{remark}

\subsection{Randomized and sequential updates}
We record versions of Theorem \ref{thm:CAVI_2block_par_alpha_init} for the randomized and sequential updates below. Throughout, $\mbox{GCorr}(\pi_n; r_0)$ is defined as in Theorem \ref{thm:CAVI_2block_par_alpha_init}.
\begin{theorem}[Two-block randomized CAVI: local contraction]\label{thm:CAVI_2block_rand_alpha_init}
Consider the randomized {\rm CAVI} algorithm \eqref{eq:seq_CAVI_gen_rand} for $d=2$. 
Suppose there exists $r_0 > 0$ sufficiently small such that $\mbox{\rm GCorr}(\pi_n; r_0) \in (0, 2)$, and the initialization satisfies $D_{\rm KL,1/2}(q_j^{(0)}\,||\,q_j^\star) \le r_0/2, (j= 1,2)$. Let $\m F_t$ denote the sigma-algebra generated by all random coordinate choices up to time $t$.  
Then, for any $t$, we have,
\be 
\mb E [D_{\rm KL, 1/2}(q^{(t+1)}\,||\,q^\star) \mid \m F_t] \le 
\bar{\kappa}_n \,D_{\rm KL, 1/2}(q^{(t)}\,||\,q^\star),
\ee
where $\bar{\kappa}_n = (1 + \kappa_n)/2 \in (0, 1)$, where $\kappa_n$ is as in Theorem \ref{thm:CAVI_2block_par_alpha_init}. As a consequence, we obtain
\be
\mb E [D_{\rm KL}(q^{(t)}\,||\,q^\star)] \le 
\bar{\kappa}_n^t \,\mb E[D_{\rm KL}(q^{(0)}\,||\,q^\star)].
\ee
\end{theorem}
Since the randomized algorithm updates only one coordinate at each pass, it is not surprising that the contraction constant $\bar{\kappa}_n$ is slower than that of the parallel update $\kappa_n$. In fact, if we consider two passes of the randomized algorithm to constitute one epoch, so that it has the same computational complexity as one epoch of the parallel algorithm, we get a contraction constant of $\bar{\kappa}_n^2$ which is still slower than $\kappa_n$. This can be possibly due to the chance of updating the same coordinate twice, so that the algorithm is not improving the other coordinate. Although randomized CAVI is a descent algorithm\footnote{A descent algorithm means that the objective value is non-increasing as the iteration increases.} regardless of the value of $\kappa_n$, an upper bound requirement on $\kappa_n$ as in the theorem is still necessary. This is because a descent algorithm may oscillate among several points without convergence; a simple example is CAVI for a bivariate Gaussian target with correlation coefficient $\rho=-1$, where the corresponding $\kappa_n=1$ and CAVI oscillates; see Section~\ref{sec:2b_examples} for related calculations for a general Gaussian target. 

Next, we consider the sequential update $(q_1^{(t)}, q_2^{(t)}) \mapsto (q_1^{(t+1)}, q_2^{(t)}) \mapsto (q_1^{(t+1)}, q_2^{(t+1)})$, where $q_1$ is updated first. 
\begin{theorem}[Two-block sequential CAVI: local contraction]\label{thm:CAVI_2block_seq_alpha_init}
Suppose there exists $r_0 > 0$ sufficiently small such that $\mbox{\rm GCorr}(\pi_n; r_0) \in (0, 2)$. Assume that the initialization for $q_1$ satisfies $D_{\rm KL,1/2}(q_1^{(0)}\,||\,q_1^\star) \le r_0/2$, and we prepare $q_2^{(0)} :\, = \argmin_{q_2} F(q_1^{(0)} \otimes q_2)$. Then, for any $t \ge 0$, one has 
\begin{align}\label{eq:seq_reinf}
\begin{aligned}
D_{\rm KL, 1/2}(q_1^{(t+1)}\,||\,q_1^\star) &\le \kappa_n D_{\rm KL, 1/2}(q_2^{(t)}\,||\, q_2^\star), \\
D_{\rm KL, 1/2}(q_2^{(t+1)}\,||\,q_2^\star) &\le \kappa_n D_{\rm KL, 1/2}(q_1^{(t+1)}\,||\, q_1^\star),
\end{aligned}
\end{align}
with $\kappa_n :\,= \mbox{\rm GCorr}^2(\pi_n; r_0)/4 \in (0, 1)$. 
Iterating, we get, for any $t \ge 1$,
\be 
D_{\rm KL,1/2}(q^{(t)}\,||\,q^\star) \le \kappa_n^{2t} D_{\rm KL,1/2}(q^{(0)}\,||\,q^\star). 
\ee
\end{theorem}
Theorem \ref{thm:CAVI_2block_seq_alpha_init} implies that under identical condition, the sequential update needs at most half the number of iterates for the parallel CAVI to achieve the same error tolerance. As a side note, the Jacobi and Gauss--Seidel methods for solving linear systems can be viewed as parallel and sequential versions of the same fixed point equation respectively, and under appropriate conditions, the speed of convergence of Gauss--Seidel can be shown to be twice as fast as the Jacobi method; see Chapter 4 of~\cite{quarteroni2006iterative} for more details. 

If the two equations of \eqref{eq:seq_reinf} are satisfied with $\kappa_{1n}$ and $\kappa_{2n}$ respectively, then we only need $\kappa_{1n} \kappa_{2n} < 1$ to obtain a contraction overall. This is a useful feature, especially in latent variable models, as one of the two steps need not be a contraction as long as the combined effect is. We shall illustrate this through an example in Section~\ref{sec:seq_eg}.

\section{Applications}\label{sec:examples}
In this section, we apply Theorems \ref{thm:CAVI_2block_par_alpha},   \ref{thm:CAVI_2block_par_alpha_init} and \ref{thm:CAVI_2block_rand_alpha_init} to a number of examples, and assess the sharpness of the derived contraction rate.

\subsection{Examples with global convergence}\label{sec:2b_examples}
We first provide some illustrations of Theorem \ref{thm:CAVI_2block_par_alpha} in concrete examples where global convergence holds.
\\[1ex]
{\bf 2d discrete distribution.} Let $\pi_n$ be the discrete distribution supported on \\
$\{(0,0), (0,1), (1,0), (1,1)\}$ with p.m.f. table
\begin{equation*}
\begin{bmatrix}
(0,0) & (0,1) & (1,0) & (1,1) \\
(1-p)/2 &p/2 & p/2& (1-p)/2
\end{bmatrix},
\end{equation*}
where $p \in (0, 1)$. Clearly, the marginals are $\mbox{Bernoulli}(0.5)$ each. It is known \citep{plummer2020dynamics} that the parallel CAVI system is globally convergent at $q_1^\star = q_2^\star =\mbox{Bernoulli}(0.5)$ provided $|\mbox{logit}(p)| < 2$. Indeed, the target density here can be viewed as an Ising model on two nodes, and the condition $|\mbox{logit}(p)| < 2$ coincides with the Dobrushin regime. Turning to verifying Theorem \ref{thm:CAVI_2block_par_alpha}, we can directly compute $\Delta_n(q_1, q_2) = 2 \{q_1(0) - q_1^\star(0)\}\{q_2(0) - q_2^\star(0)\} \log \{(1-p)/p\}$ for any probability measures $q_1, q_2$ supported on $\{0, 1\}$. This gives $|\Delta_n(q_1, q_2)| = 2 \|q_1 - q_1^\star\|_{\TV} \|q_2 - q_2^\star\|_{\TV} |\mbox{logit}(p)|$. Now use Pinsker's inequality to obtain $2 \|q_i - q_i^\star\|_{\TV}^2 \le \min\{D_{\rm KL}(q_i \,||\, q_i^\star), D_{\rm KL}(q_i^\star \,||\, q_i)\} \le D_{\rm KL, \alpha}(q_i \,||\, q_i^\star)$ for any $\alpha \in [0,1]$, implying $\mbox{GCorr}(\pi_n) \le |\mbox{logit}(p)|$. Hence, the condition $\mbox{GCorr}(\pi_n) < 2$ in Theorem \ref{thm:CAVI_2block_par_alpha} is implied by $|\mbox{logit}(p)| < 2$, and thus Theorem \ref{thm:CAVI_2block_par_alpha} certifies global convergence in the Dobrushin regime. \qed
\\[1ex]
{\bf Gaussian target.} Suppose $\pi_n \equiv N_p(\theta_0, (n Q)^{-1})$ where $Q$ is a fixed positive definite matrix. Consider a mean-field decomposition $q(\theta) = q_1(\theta_1) \, q_2(\theta_2)$ where we decompose $\theta = (\theta_1, \theta_2)'$ with $\theta_i \in \mb R^{p_i}$. Partition $\theta_0 = (\theta_{01}, \theta_{02})'$ and 
\be \label{eq:two_block_prec}
Q =  
\begin{bmatrix}
Q_{11} & Q_{12}\\
Q_{21} & Q_{22}\\
\end{bmatrix}
\ee
into the corresponding blocks. The parallel CAVI updates are
\be 
q_1^{(t+1)}(\theta_1) = \m N(\theta_1; m_1^{(t+1)}, (n Q_{11})^{-1}), \quad q_2^{(t+1)}(\theta_2) = \m N(\theta_2; m_2^{(t+1)}, (n Q_{22})^{-1}),
\ee
with 
\be 
m_1^{(t+1)} = \theta_{01} - Q_{11}^{-1} Q_{12} \big(E_{q_2^{(t)}}(\theta_2) - \theta_{02}\big),  \, m_2^{(t+1)} = \theta_{02} - Q_{22}^{-1} Q_{21} \big(E_{q_1^{(t)}}(\theta_1) - \theta_{01}\big). 
\ee
Therefore, we have the explicit dynamic $(m_1^{(t+2)} - \theta_{01}) = Q_{11}^{-1}Q_{12}Q_{22}^{-1}Q_{21}(m_1^{(t)} - \theta_{01})$. This dynamic is a contraction if and only if the spectral radius of $A=Q_{11}^{-1}Q_{12}Q_{22}^{-1}Q_{21} = Q_{11}^{-1/2}
BB^T Q_{11}^{1/2}$ is less than one, where $B= Q_{11}^{-1/2} Q_{12} Q_{22}^{-1/2}$. Note that the spectral radius of $A$ also quantifies the exact contraction rate of CAVI.  Since the spectral radius of $A$ is the same as the spectral radius of $BB^T$, which equals to the square of the operator norm $\|B\|_2$ of $B$. The next result below shows that the contraction rate derived from our general Theorem~\ref{thm:CAVI_2block_par_alpha} is sharp for Gaussian targets.

\begin{proposition}\label{prop:two_block_g}
For a Gaussian target $\pi_n \equiv N_p(\theta_0, (n Q)^{-1})$ with $Q$ positive definite, and a mean-field decomposition as described above, $\mbox{\rm GCorr}(\pi_n) = 2 \|B\|_2 < 2$. 
\end{proposition}
It is immediate from the nature of the updates that the mean-field solution will be $q_j^\star \equiv N(m_j^\ast, (n Q_{jj})^{-1})$ with $m_j^\ast = \theta_{0j}$. Moreover, we can set $\bar{Q}_j = \{ N(m_j, (n Q_{jj})^{-1}) \,:\, m_j \in \mb R^{p_j} \}$ in lieu of Remark \ref{rem:restrict}. 
One can then compute
$
\Delta_n(q_1, q_2) = - n \delta_1' Q_{12} \delta_2
$
where $\delta_j = \mb E_{q_j}[\theta_j] - \mb E_{q_j^\star}[\theta_j] = m_j - m_j^\star$ for $j = 1, 2$. Also, for any $\alpha \in [0, 1]$ and $q_j \equiv N(m_j, (n Q_{jj})^{-1}) \in \bar{Q}_j$, we have $D_{\rm KL, \alpha}(q_j\,||\,q_j^\star) = \|Q_{jj}^{1/2} \delta_j\|^2/2$ for $j = 1, 2$. We can then show; see proof of Proposition \ref{prop:two_block_g}; that $\mbox{\rm GCorr}(\pi_n) = 2 \|Q_{11}^{-1/2} Q_{12} Q_{22}^{-1/2}\|_2 < 2$. Thus, the conclusion of Theorem \ref{thm:CAVI_2block_par_alpha} holds for any non-singular Gaussian distribution.  \qed
\\[1ex]
{\bf Target with Gaussian conditionals.} Consider the target distribution:
\be\label{eq:den_gauss_conds} 
\pi(u_1, u_2) = Z^{-1} \exp \left[- \frac{1}{2}\big(u_1^2 + u_2^2 + u_1^2 u_2^2 \big) \right], \quad u_1, u_2 \in \mb R.
\ee
This is an example of a bivariate non-Gaussian joint distribution whose both conditionals are univariate Gaussian \cite[Chapter 30]{mackay2003information}. It is straightforward to see that the parallel CAVI updates are given by: 
\be 
q_j^{(t+1)} \equiv N\big(0,1/\tau_j^{(t+1)}\big), \quad (j = 1, 2), 
\ee
where 
\be 
\tau_1^{(t+1)} &= 1 + \mb E_{q_2^{(t)}}(U_2^2) = 1 + \frac{1}{\tau_2^{(t)}}, \\
\tau_2^{(t+1)} &= 1 + \mb E_{q_1^{(t)}}(U_1^2) = 1 + \frac{1}{\tau_1^{(t)}}.
\ee
\begin{proposition}\label{prop:gauss_conds}
For $\pi$ as in \eqref{eq:den_gauss_conds}, we have $\mathrm{GCorr}(\pi) \le 4/(1+\sqrt{5}) \approx 1.24$.  
\end{proposition}

\noindent {\bf Probit regression.} Finally, we provide an example involving latent variables. Suppose $y_i \mid x_i, \beta \overset{ind.} \sim \mbox{Bernoulli}(\Phi(x_i'\beta))$ independently for $i \in [n]$, and let $X \in \mb R^{n \times d}$ and $y \in \{0, 1\}^n$ denote the design matrix and response vector, respectively. Assume prior $\beta \sim N(0, \kappa^{-1} I_p)$ for some $\kappa > 0$. To enable closed-form updates, consider an Albert--Chib  data augmentation \citep{albert1993bayesian,ormerod2010explaining} to introduce latent variables $z = (z_1, \ldots, z_n)$ with $y_i = \ind(z_i > 0)$ and $z_i \overset{ind.}\sim N(x_i'\beta, 1)$. The augmented likelihood
\be 
L(\beta, z) = \bigg[\prod_{i=1}^n \big\{ y_i \ind(z_i > 0) + (1-y_i) \ind(z_i \le 0) \big\} \bigg] \, \bigg[\prod_{i=1}^n e^{-(z_i - x_i'\beta)^2/2} \bigg].
\ee
Consider the mean-field decomposition 
\be 
q(\beta, z) = q_{\blds \beta}(\beta) \, q_{\blds z}(z).
\ee
Let $TN_1$ and $TN_0$ respectively denote univariate truncated normals with truncation region $(0, \infty)$ and $(-\infty, 0)$. The parallel updates then take the form
\be
q_{\blds \beta}^{(t+1)}(\beta) = \m N_p(\beta; m^{(t+1)}, \Sigma); \quad q_{\blds z}^{(t+1)}(z) = \prod_{i=1}^n q_i^{(t+1)}(z_i), \ q_i^{(t+1)}(z_i) \equiv N_{y_i}(z_i; \alpha_i^{(t+1)},1)
\ee
with $\Sigma = (X'X + \kappa \mr I_p)^{-1}$, and 
\be \label{eq:probit_params_update}
\alpha^{(t+1)} = X m^{(t)}, \quad m^{(t+1)} = \Sigma X' E_{q_{\blds z}^{(t)}}(z) = \Sigma X' b^{(t)},
\ee
where $\alpha^{(t)} = (\alpha_1^{(t)}, \ldots, \alpha_n^{(t)})$ and $b^{(t)} = (b_1^{(t)}, \ldots, b_n^{(t)})$ with $b_i^{(t)} = E_{q_i^{(t)}}(z_i)$. 
\begin{proposition}\label{prop:probit}
We have, 
\be 
\mbox{\rm GCorr}(\pi_n) \le 2 \lambda_{\max}^{1/2}\big((X'X + \kappa \mr I_p)^{-1/2} (X'X)(X'X + \kappa \mr I_p)^{-1/2}\big),
\ee
where $\lambda_{\max}(A)$ denotes the largest eigenvalue of a positive semi-definite matrix $A$. 
\end{proposition}
Since $(X'X + \kappa \mr I_p)^{-1/2} (X'X)(X'X + \kappa \mr I_p)^{-1/2} \prec \mr I_p$ in the positive semi-definite sense for any $\kappa > 0$, we have that $\mbox{\rm GCorr}(\pi_n) < 2$. In particular, if $X'X = n \mr I_p$, then we have the simplified bound $2 \sqrt{n/(n+\kappa)}$. We note that the expression for $\gamma_n$ remains valid if $X$ is rank-deficient or $p > n$.

\subsection{Examples with local convergence}\label{subsec:local}
We now consider various statistical examples involving models without and with latent variables to illustrate applications of Theorem \ref{thm:CAVI_2block_par_alpha_init} to show local convergence (under suitable condition on the initialization). 
\\[1ex]
{\bf Gaussian model with unknown mean and precision.} Consider $x_1, \ldots, x_n \mid \mu, \tau \overset{ind.} \sim N(\mu, \tau^{-1})$, and consider independent priors $\mu \sim N(0, \kappa^{-1})$ and $\tau \sim \mbox{Gamma}(a_0, b_0)$. Assume a mean-field decomposition $q(\mu, \tau) = q_{\blds \mu}(\mu) \, q_{\blds \tau}(\tau)$. The parallel updates are given by
\be \label{eq:gauss_mp_1}
q_{\blds \mu}^{(t+1)}(\mu) = N(\mu; m^{(t+1)}, 1/s^{(t+1)}), \ q_{\blds \tau}^{(t+1)}(\tau) = \mbox{Gamma}(\tau; n/2+a_0, b^{(t+1)}), 
\ee
where 
\be 
s^{(t+1)} = n E_{q_{\blds \tau}^{(t)}}(\tau) + \kappa, \, m^{(t+1)} = \frac{n E_{q_{\blds \tau}^{(t)}}(\tau) \, \bar{x}}{n E_{q_{\blds \tau}^{(t)}}(\tau) + \kappa}, \, b^{(t+1)} = \frac{1}{2} E_{q_{\blds \mu}^{(t)}} \|x - \mu 1_n\|_2^2 + b_0.
\ee
Given the nature of the updates, we can set $\bar{\m Q}_{\blds \mu} = \{q_{\blds \mu} \equiv N(m,s^{-1}) \,:\, (m, s) \in \mb R \times \mb (0, \infty)\}$, and $\bar{\m Q}_{\blds \tau} = \{q_{\blds \tau} \equiv \mbox{Gamma}(n/2+a_0,b) \,:\, b \in (0, \infty)\}$. Also, evidently,  
$q_{\blds \mu}^\star \equiv N(m^\star, 1/s^\star)$ and $q_{\blds \tau} \equiv \mbox{Gamma}(n/2+a_0,b^\star)$. Moreover, it follows from the concentration of the VB posterior that $|m^\star - \bar{x}| = O(1/\sqrt{n})$, $s^\star \in [c_1 n, c_2 n], b^\star \in [c_3 n, c_4 n]$ with high probability under the true data generating model, where the $c_i$s are positive constants. We shall operate within this high-probability set below. We then have the following result. 
\begin{proposition}\label{prop:gauss_meanprec}
Let $\omega \in (0, 1)$ be a small enough constant such that $(2 \sqrt{2e}/c_3) \omega < 1$. Set $r_0 = W_0(\omega^2n)/2$, where for $x > 0$, $W_0(x)$ is the unique solution to the equation $y e^y = x$ in terms of $y$. Then, for $n$ sufficiently large, we have ${\rm GCorr}(\pi_n, r_0) \le (C n^{-1/2} + 1)$ for a global constant $C$ free of $n$. 
\end{proposition}
\noindent {\bf General parametric models.} Consider the general setup of a parametric model $x_1, \ldots, x_n \mid \theta \overset{ind.} \sim f(\cdot \mid \theta)$, where $\theta = (\theta_1, \theta_2) \in \mb R^2$. 
Let $\ell_n(\theta) = n^{-1} \sum_{i=1}^n \log f(x_i \mid \theta)$ denote the average log-likelihood function. We assume the usual regularity conditions on the likelihood surface to ensure the existence of a unique maximum likelihood estimator $\widehat{\theta}_n = (\widehat{\theta}_{1n}, \widehat{\theta}_{2n})'$, and that the average observed Fisher information matrix evaluated at the maximum likelihood estimator, $\widehat{I}_n = - \nabla^2 \ell_n(\widehat{\theta}_n)$, is positive definite. In addition, we also assume $\ell_n$ to be thrice continuously differentiable, and the mixed partial derivatives of order 3 to be bounded in magnitude by a global constant, i.e., that there exists a constant $C > 0$ such that for any $\theta$, 
\be \label{eq:third_order_der_bd}
\max \bigg\{ \left| \frac{\partial^3 \ell_n(\theta)}{\partial^2 \theta_1 \partial \theta_2} \right|, \left| \frac{\partial^3 \ell_n(\theta)}{\partial \theta_1 \partial^2 \theta_2} \right| \bigg\} \le C.
\ee
This is readily satisfied (given the other assumptions) if the parameter space for $\theta$ is compact. 

Assume independent priors $\theta_j \sim \pi_j$ for $j = 1, 2$, so that the posterior distribution 
$\pi_n(\theta) \,\propto\, e^{n \ell_n(\theta)} \, \pi_1(\theta_1) \pi_2(\theta_2)$. Consider a mean-field decomposition $q(\theta) = q_1(\theta_1) \, q_2(\theta_2)$. Let $q_1^\star \otimes q_2^\star$ be a global minimizer of the variational objective $F$. We assume that \\
{\bf Assumption A1.} $q_j^\star$ is sub-Gaussian, with its squared sub-Gaussian norm bounded above by $1/(n \widehat{I}_{n,jj}), (j= 1, 2)$. 
\\
{\bf Assumption A2.} There exists a constant $C$ such that the parallel CAVI iterates for $q_j$ all lie in $\bar{\m Q}_j(C)$, with 
$
\bar{\m Q}_j(C) = \big\{ q_j \,:\, \mb E_{q_j}(\theta_j - \mb E_{q_j}(\theta_j))^4 \le C [\mbox{var}_{q_j}(\theta_j)]^2 \big\}
$
the collection of densities whose fourth central moment is bounded by $C$ times the squared variance. 

In regular parametric models, Bernstein--von Mises (BvM) type results have been established \citep{wang2019frequentist} for the mean-field approximation under which $q_j^\star$ is approximately distributed as $N(\widehat{\theta}_{jn}, (n \widehat{I}_{n, jj})^{-1/2})$ for large $n$. In A1, we refrain from making any assumptions on the shape of $q_j^\star$ and only assume it to be sub-Gaussian with the sub-Gaussian norm bounded by the asymptotic standard deviation. Assumptions A2 is rather mild; we only assume the fourth central moment is bounded by a global constant times the squared variance across the solution path. For example, if the iterates lie in a Gaussian family, the constant $C = 3$. In particular, we do not assume the CAVI iterates lie in an exponential family or some parameterized family of densities, or even that they are sub-Gaussian.

To obtain a convergence result under these assumptions, we first state a corollary of Theorems \ref{thm:CAVI_2block_par_alpha} and \ref{thm:CAVI_2block_par_alpha_init}. 
\begin{corollary}[Two-block parallel CAVI: one-sided version]\label{cor:2block_par_onesided}
Suppose there exists $r_0 > 0$ and $\gamma_n \in (0,1)$ such that for any $t \ge 0$, $D_{\rm KL}(q_j^{(t)}\,||\,q_j^\star) \le r_0, (j=1, 2)$ implies 
\begin{align}\label{eq:2block_par_onesided}
\begin{aligned}
|\Delta_n(q_1^{(t+1)},q_2^{(t)})| & \le \frac{\gamma_n}{2} \big[ D_{\rm KL}(q_1^{(t+1)}\,||\,q_1^\star) +  D_{\rm KL}(q_2^{(t)}\,||\,q_2^\star) \big],  \\
|\Delta_n(q_1^{(t)},q_2^{(t+1)})| & \le \frac{\gamma_n}{2} \big[D_{\rm KL}(q_1^{(t)}\,||\,q_1^\star) + D_{\rm KL}(q_2^{(t+1)}\,||\,q_2^\star)\big].
\end{aligned}
\end{align}
If the initialization satisfies $D_{\rm KL}(q_j^{(0)}\,||\,q_j^\star) \le r_0$, $(j=1, 2)$, then for any $t \ge 1$, 
\be 
D_{\rm KL}(q^{(t+1)}\,||\,q^\star) \le \kappa_n D_{\rm KL}(q^{(t)}\,||\,q^\star),
\ee
where $\kappa_n = \frac{\gamma_n}{2-\gamma_n} \in (0,1)$. 
\end{corollary}
Corollary \ref{cor:2block_par_onesided} requires a bound on $|\Delta_n|$ only in terms of the one-sided KL divergences $D(q_j \,||\, q_j^\star)$. This is convenient in the present context since we make minimal assumptions on the iterates $\{q_j^{(t)}\}$. The assumption that $q_j^\star$ is sub-Gaussian allows us to exploit the transport inequality in Lemma \ref{lem:tce}. Using Corollary \ref{cor:2block_par_onesided}, we are prepared to state the following result. 
\begin{theorem}\label{thm:2d_bvm}
Assume the conditions on the likelihood surface around equation \eqref{eq:third_order_der_bd}. Also, assume Assumptions (A1) and (A2) hold. Suppose the parallel CAVI algorithm is initialized so that $D_{\rm KL}(q_j^{(0)} \,||\, q_j^\star) \le r_0 :\,= \omega^2 n$ for some sufficiently small constant $\omega$ independent of $n$. Then, we have, for any $t \ge 0$ that 
\be 
D_{\rm KL}(q^{(t+1)}\,||\,q^\star) \le \frac{\gamma_n}{2-\gamma_n} D_{\rm KL}(q^{(t)}\,||\,q^\star),
\ee
with 
\be \label{eq:gamn_bvm}
\gamma_n = \left[ 2 \frac{|\widehat{I}_{n,12}|}{ \sqrt{\widehat{I}_{n,11}} \, \sqrt{\widehat{I}_{n,22}}} + \frac{2C}{\sqrt{n}} + 3C \omega \right], 
\ee
for some global constant $C > 0$. 
\end{theorem}
Letting $\widehat{\rho}_n :\,= \frac{|\widehat{I}_{n,12}|}{ \sqrt{\widehat{I}_{n,11}} \, \sqrt{\widehat{I}_{n,22}}}$ denote the magnitude of the correlation in the asymptotic sampling distribution of $\widehat{\theta}_n$, it is evident from \eqref{eq:gamn_bvm} that a non-trivial upper bound on $\widehat{\rho}_n$ is needed to ensure $\gamma_n < 1$. 
This restriction on the correlation arises as an artifact of Corollary \ref{cor:2block_par_onesided} which only uses one-sided KL divergences. As a point of comparison, if we considered a Gaussian target distribution $\pi_n \equiv N(\theta_0, \widehat{I}_n^{-1}/n)$ and applied Corollary \ref{cor:2block_par_onesided}, we would get $\gamma_n = 2 \widehat{\rho}_n$ and thus require $\widehat{\rho}_n < 1/2$ for convergence\footnote{Whereas Proposition \ref{prop:two_block_g} informs that exponential convergence holds for any $\widehat{\rho}_n \in (0, 1)$.}. We only incur additional lower-order terms $2C/\sqrt{n} + 3 C \omega$ in the present context for a much wider family of target distributions, and under minimal assumptions on the iterates. \\[1ex]
\noindent{\bf Exponential-family latent variable model.} Now, consider the general setup of \cite{hoffman2013stochastic} where data $x_i$ and latent variable $z_i$ are conditionally independent given parameter $\beta \in \mb R^d$ for $i \in [n]$, with joint density in an exponential family
\be 
p(x_i, z_i \mid \beta) = h(x_i, z_i) \exp\big\{\beta' s(x_i, z_i) - a_{\blds x,\blds z}(\beta) \big\}, 
\ee
with $d$ dimensional sufficient statistics $s(x_i, z_i)$ and cumulant function $a_{\blds x, \blds z}: \mb R^d \to \mb R$. As in \cite{hoffman2013stochastic}, consider a conditionally conjugate Diaconis--Ylvisaker prior \citep{diaconis1979conjugate} on the parameter $\beta$ as 
\be 
\pi_{\DY}(\beta \mid \alpha)  = h_0(\beta) \, \exp \big\{\alpha' r(\beta) - a_{\blds \beta}(\alpha) \big\},
\ee
where $r(\beta) = [\beta, -a_{\blds x,\blds z}(\beta)]'$ denotes the sufficient statistics for the prior family, $\alpha = [\alpha_1; \alpha_2] \in \mb R^{d+1}$ designate fixed prior hyperparameters, and $h_0(\cdot)$ does not depend on $\alpha$. The cumulant generating function $a_{\blds \beta} : \mb R^{d+1} \to \mb R$ is a convex function. The bold subscripts in the cumulant generating functions are only used for indexing and should not be confused as variables. 

Consider now a mean-field approximation $q(\beta, z) = q_{\blds \beta}(\beta) \, q_{\blds z}(z)$ to the joint posterior $\pi_n(\beta, z) :\,= p(\beta, z \mid x) \, \propto \, \pi_{\DY}(\beta \mid \alpha) \, \prod_{i=1}^n p(x_i, z_i \mid \beta)$. To enable closed-form CAVI updates, it is customary to additionally assume that the distribution of $z_i \mid x_i, \beta$ lies in an exponential family, 
\be 
p(z_i \mid x_i, \beta) = \tilde{h}(z_i) \exp \{\eta(\beta, x_i)' u(z_i) - a_{\blds z}(\eta(\beta, x_i)) \}.
\ee
We shall denote a general density in the above density family by $p_{\LV}(z_i \mid \eta_i)$. The parallel updates can now be obtained as 
\be 
q_{\blds \beta}^{(t)}(\beta) = \pi_{\DY}(\beta \mid \alpha^{(t)}), \quad q_{\blds z}^{(t)}(z) = \prod_{i=1}^n q_i^{(t)}(z_i) \text{ with } q_i^{(t)}(z_i) = p_{\LV}(z_i \mid \eta_i^{(t)}), 
\ee
where 
\be 
\alpha^{(t+1)} = \bigg[\alpha_1 + \sum_{i=1}^n \mb E_{q_i^{(t)}} [s(x_i, z_i)]; \alpha_2 + n\bigg], \quad
\eta_i^{(t+1)} = \mb E_{q_{\blds \beta}^{(t)}} [\eta(\beta, x_i)], \quad i \in [n]. 
\ee
To simplify the ensuing analysis, we shall henceforth make the mild assumption that $h(x_i, z_i) = \tilde{h}(z_i) = 1$ for all $i \in [n]$. We also assume that there exists a positive function $c(\cdot)$ such that 
\be \label{eq:eta_th_x}
\|\eta(\theta, x)\| \le c(x) \|\theta\| \text{ for all } x, \theta. 
\ee

Given the form of the updates, it is clear that $q_{\blds \beta}^\star \equiv \pi_{\DY}(\cdot \mid \alpha^\star)$ with $\alpha^\star(d+1) = \alpha_2 + n$, and $q_{\blds z}^\star \equiv p_{\LV}(\cdot \mid \eta^\star)$ with $\eta^\star = [(\eta_1^\star)';\ldots; (\eta_n^\star)']$. We assume that the convex function $a_{\blds \beta}$ satisfies $\lambda_{\min}(a_{\blds \beta}(\alpha)) \ge 2L/\min\{C,\|\alpha\|\}$ for all $\alpha$, where $L > 0$ and $C \in (0, 1)$ are constants. This implies a local strong convexity condition
\be \label{eq:lsc_a_beta}
a_{\blds \beta}(\alpha) \ge a_{\blds \beta}(\alpha^\star) + \langle \alpha - \alpha^\star, \nabla a_{\blds \beta}(\alpha^\star) \rangle + \frac{L\|\alpha - \alpha^\star\|^2}{\min\{C,\|\alpha\|\}}.
\ee


We similarly assume, for any $i \in [n]$,
\be \label{eq:sc_a_z}
a_{\blds z}(\eta_i) \ge a_{\blds z}(\eta_i^\star) + \langle \eta_i - \eta_i^\star, \nabla a_{\blds z}(\eta_i^\star) \rangle + L \|\eta_i - \eta_i^\star\|^2. 
\ee
We now state a bound on $\mbox{\rm GCorr}(\pi_n)$ in the present setup. 
\begin{proposition}\label{prop:exp_family_lvm}
Assume \eqref{eq:eta_th_x}, \eqref{eq:lsc_a_beta} and \eqref{eq:sc_a_z} hold. Assume there exists a constant $\omega > 0$ such that 
\begin{itemize}
\item[(i)] For any $\alpha$ with $\|(\alpha[d]-\alpha^\star[d])/n\| < \omega \sqrt{d/n}$, we have $\ell_\alpha (d/n) \leq \lambda_{\min} \{H_{\blds \beta}(\alpha)\} \le \lambda_{\max} \{H_{\blds \beta}(\alpha) \} \le u_\alpha (d/n)$ where $0 < \ell_\alpha \le u_\alpha$ are constants free of $n$, and $H_{\blds \beta}(\alpha)$ denotes the first $d \times d$ block of $\nabla^2 a_{\blds \beta}(\alpha)$.

\item[(ii)] For any $\eta$ with $\big(n^{-1} \sum_{i=1}^n \|\eta_i - \eta_i^\star\|^2\big)^{1/2} < \omega \sqrt{d}$, we have $\ell_\eta \leq \lambda_{\min} \{\nabla^2 a_{\blds z}(\eta_i)\}  \leq \lambda_{\min}\{\nabla^2 a_{\blds z}(\eta_i)\} \leq u_\eta$ for all $i \in [n]$, where $0 < \ell_\eta \le u_\eta$ are constants free of $n$. 
\end{itemize}
Finally, assume that the function $c$ in \eqref{eq:eta_th_x} satisfies $\sum_{i=1}^n c^2(x_i) < (n/d) \frac{\ell_\alpha \ell_\eta}{(u_\alpha u_\eta)^2}$. Then, setting $r_0 = L \omega^2 d n$ in Theorem \ref{thm:CAVI_2block_par_alpha_init},
\be 
\mbox{\rm GCorr}(\pi_n; r_0) \le 2 \frac{u_\alpha u_\eta}{(\ell_\alpha \ell_\eta)^{1/2}} \, \sqrt{\frac{1}{n} \sum_{i=1}^n c^2(x_i)} < 2.
\ee
\end{proposition}
For $\beta \sim \pi_{\DY}(\cdot \mid \alpha)$, $\mbox{cov}(r(\beta)) = \nabla^2 a_{\blds \beta}(\alpha)$, and therefore, $\mbox{cov}(\beta) = H_{\blds \beta}(\alpha)$. In situations where the optimal variational solution $q_{\blds \beta}^\star = \pi_{\DY}(\cdot \mid \alpha^\star)$ for the parameter part has an optimal $\sqrt{d/n}$ asymptotic spread [CITE], all the eigenvalues of $H_{\blds \beta}(\alpha^\star)$ are of order $d/n$. Observe that in our parameterization, $\alpha[d]$ involves sum of sufficient statistics, and hence $\alpha[d]/n$ is a stable parameterization, i.e., one which does not grow with $n$. Thus, $\big\{ \|(\alpha[d] - \alpha^\star[d])/n\| \le \omega \sqrt{d/n} \big\}$ gives a $\sqrt{d/n}$-neighborhood around $\alpha^\star[d]$, and assumption (i) requires that the smallest and largest singular values of the Hessian matrix $H_{\blds \beta}$ remain of the order of $d/n$ in this neighborhood. A similar interpretation holds for assumption (ii) with the main difference that the latent variables do not typically exhibit concentration, and hence we assume that the eigenvalues of the Hessian are bounded by constants in a fixed small neighborhood of $\eta^\star$. 

\subsection{Example illustrating a feature of sequential update}\label{sec:seq_eg}
Here, we illustrate an interesting feature of the sequential update as promised earlier in the discussion after Theorem \ref{thm:CAVI_2block_seq_alpha_init}. We use a two-component Gaussian mixture model as an illustrative example, although this feature should find usage in more general latent variable models. \\
\noindent {\bf Two-component Gaussian mixture.} Consider a Gaussian mixture model $x_i \stackrel{\text{i.i.d.}} \sim \frac{1}{2} \, \m N(0, 1) + \frac{1}{2} \, \m N(\mu, 1)$ with prior $\mu \sim \mbox{N}(0, \tau_0^{-1})$. Consider the standard latent variable augmentation to write $x_i \mid z_i, \mu \stackrel{ind.} \sim N(\mu \ind(z_1=2), 1)$ and $\mbox{pr}(z_i = 1) = \mbox{pr}(z_i = 2) = 1/2$. Letting $z = (z_1, \ldots, z_n)'$, consider a mean-field decomposition $q(\mu, z) = q_{\blds \mu}(\mu) \, q_{\blds z}(z)$. The updates for $z$ lie in the family $q_{\blds z}(z) = \prod_{i=1}^n q_i(z_i)$, where each $q_i$ is a two-point distribution on $\{1, 2\}$ with probabilities $(1-p_i)$ and $p_i$ respectively. Also, the update for $\mu$ is of the form $N(m, \tau^{-1})$. Consider the sequential update which proceeds as 
\be \label{eq:gmm_seq}
(q_{\blds \mu}^{(t)}, q_{\blds z}^{(t)}) \mapsto (q_{\blds \mu}^{(t)}, q_{\blds z}^{(t+1)}) \mapsto (q_{\blds \mu}^{(t+1)}, q_{\blds z}^{(t+1)}).
\ee  
The update equations for the parameters are: 
\be
\mbox{logit}(p_i^{(t+1)}) &= m^{(t)} x_i - \frac{1}{2} \bigg((m^{(t)})^2 + \frac{1}{\tau^{(t)}}\bigg), \\
(m^{(t+1)}, \tau^{(t+1)}) &= \left( \frac{ \sum_{i=1}^n p_i^{(t+1)} x_i }{ \tau_0 + \sum_{i=1}^n p_i^{(t+1)} }, \tau_0 + \sum_{i=1}^n p_i^{(t+1)} \right). 
\ee
For $(m, \tau) \in \mb R \times (0, \infty)$, define $(m,\tau) \mapsto p_i(m, \tau) \in (0,1)$ through the identity
\be \label{eq:pimtau}
\mbox{logit}(p_i(m, \tau)) = m x_i - \frac{(m^2 + \tau^{-1})}{2},
\ee
so that we can write $p_i^{(t+1)} = p_i(m^{(t)}, \tau^{(t)})$ in the previous display. Also, we have $p_i^\star = p_i(m^\star, \tau^\star)$ from the self-consistency of the VB solution. 
\begin{proposition}\label{prop:two_comp_mix}
Suppose the data generating distribution is compactly supported, and there exists $\delta > 0$ sufficiently small such that $\min_{i \in [n]} \max\{p_i^\star, 1-p_i^\star\} \ge 1 - \delta$ holds with high probability under the data generating distribution. Also assume $\tau^\star \ge c_1 n$ for some $c_1 > 0$. Assume the sequential update \eqref{eq:gmm_seq} is initialized so that $D_{1/2}(q_{\blds \mu}^{(0)}\,||\,q_{\blds \mu}^\star) \le r_0 := \omega^2 n$ for some fixed $\omega > 0$. Then, there exists $\kappa_n \in (0, 1)$ such that for any $t \ge 1$, 
\be 
D_{\rm KL, 1/2}(q_{\blds \mu}^{(t+1)}\,||\,q_{\blds \mu}^\star)  \le \kappa_n D_{\rm KL, 1/2}(q_{\blds \mu}^{(t)}\,||\,q_{\blds \mu}^\star).
\ee
\end{proposition} 
The proof of Proposition \ref{prop:two_comp_mix} establishes 
\be 
D_{\rm KL, 1/2}(q_{\blds z}^{(t+1)}\,||\,q_{\blds z}^\star) &\le \kappa_{1n} D_{\rm KL, 1/2}(q_{\blds \mu}^{(t)}\,||\,q_{\blds \mu}^\star), \\
D_{\rm KL, 1/2}(q_{\blds \mu}^{(t+1)}\,||\,q_{\blds \mu}^\star) &\le \kappa_{2n} D_{\rm KL, 1/2}(q_{\blds z}^{(t+1)}\,||\,q_{\blds z}^\star),
\ee
where $\kappa_{2n}$ is bounded by a constant, not necessarily smaller than one. However, exploiting the initialization condition, $\kappa_{1n}$ is shown to be sufficiently small so that $\kappa_{1n} \, \kappa_{2n} < 1$. 

\section{More than two blocks}\label{sec:gtr_2b}
While a complete treatment of more than two blocks is beyond the scope of the present paper, we provide some initial results. In particular, we offer an expanded notion of generalized correlation, and derive a global contraction theorem for the parallel CAVI algorithm with $d$ blocks. Similar results can be derived for local contraction as well as for sequential and randomized updates, and are omitted. 

Given $j \in [d]$, and $q = q_j \otimes q_{-j}$, we define 
\be \label{eq:Deltaq_gen}
\Delta_{j,n}(q_j, q_{-j}) = \int (q_j - q_j^\star) (q_{-j} - q_{-j}^\star) \log \pi_n
\ee
in an analogous manner to \eqref{eq:Deltaq}, where $q^\star = q_j^\star \otimes q_{-j}^\star$ is a global minima of $F$. We state a partial extension of Lemma \ref{prop:LSC} first. 
\begin{lemma}\label{lem:LSC_gen}
The optimality of $q^\star$  implies
\be \label{eq:CAVI_opt}
q_j^\star \,\propto\, \exp\left\{\int q_{-j}^\star \log\pi_n\right\}, \ j \in [d].
\ee
Moreover, for any $j \in [d]$ and $q = q_j \otimes q_{-j}$,
\be 
F(q_j^\star \otimes q_{-j}) - F(q_j \otimes q_{-j}) 
= -D_{\rm KL}(q_j\,||\,q_j^\star) + \Delta_{j, n}(q_j, q_{-j}).
\ee
\end{lemma}
Next, we offer a generalized definition of GCorr for any $d \ge 2$, which coincides with our earlier definition when $d =2$. For any $j \in [d]$, define, 
\be\label{eq:Gcorr_j_alpha}
\mbox{\rm GCorr}_\alpha^{(j)}(\pi_n) = \sup_{q_j \in \m Q_j \backslash \{q_j^\star\}} \frac{|\Delta_{j,n}(q_j,\,q_{-j})|}{\sqrt{D_{\rm KL, \alpha}(q_j \,||\, q_j^\star) }\, \sqrt{D_{\rm KL, 1-\alpha}(q_{-j}\,||\, q_{-j}^\star) }},
\ee
and 
\be 
\mbox{\rm GCorr}^{(j)}(\pi_n) = \inf_{\alpha\in[0,1]} \bar{\gamma}_{n, \alpha}^{(j)}, \quad \bar{\gamma}_{n,\alpha}^{(j)} = \left[\max \left\{ \mbox{\rm GCorr}^{(j)}_\alpha(\pi_n), \mbox{\rm GCorr}^{(j)}_{1-\alpha}(\pi_n) \right\} \right]. 
\ee
Clearly, $\mbox{\rm GCorr}^{(j)}(\pi_n)$ is the generalized correlation between the $j$th block and the remaining blocks collapsed together. 
We then define the $d$-block version of the generalized correlation as the maximum of these $d$-many component-specific generalized correlations:
\be \label{eq:GCorr_d}
\mbox{\rm GCorr}_d(\pi_n) = \max_{j \in [d]} \mbox{\rm GCorr}^{(j)}(\pi_n). 
\ee
When $d = 2$, $\mbox{\rm GCorr}^{(1)}(\pi_n) = \mbox{\rm GCorr}^{(2)}(\pi_n) = \mbox{\rm GCorr}(\pi_n)$, recovering the older definition. 
\begin{remark}
Since $\inf_\alpha \max_j \bar{\gamma}_{n,\alpha}^{(j)} \ge \max_j \inf_\alpha \bar{\gamma}_{n,\alpha}^{(j)}$, it is natural to define the smaller quantity to be $\mbox{\rm GCorr}_d(\pi_n)$ and impose bounds on it. 
\end{remark}
We are now prepared to state a generalization of Theorem \ref{thm:CAVI_2block_par_alpha}. 
\begin{theorem}[$d$-block parallel CAVI]\label{thm:CAVI_dblock_par_alpha}
Suppose the target density $\pi_n$ satisfies $\mbox{\rm GCorr}_d(\pi_n) \in (0, 2/\sqrt{d-1})$.
Then, for any initialization $q^{(0)} = \otimes_{j=1}^d q_j^{(0)} \in \m Q$ of the parallel {\rm CAVI} algorithm, one has a contraction
\be \label{eq:par_cavi_exp_cgence}
D_{\rm KL,1/2}(q^{(t+1)}\,||\,q^\star)  \le \kappa_n D_{\rm KL, 1/2}(q^{(t)}\,||\,q^\star),
\ee
for any $t \ge 0$, where the contraction constant $\kappa_n :\,= (d-1) \, \mbox{\rm GCorr}_d^2(\pi_n)/4 \in (0, 1)$.
\end{theorem}
All the remarks after Theorem \ref{thm:CAVI_2block_par_alpha} naturally extend to the $d$-block setting. A key distinction from the $d = 2$ case is the more stringent requirement on $\mbox{\rm GCorr}_d(\pi_n)$, which now needs to be bounded by $2/\sqrt{d-1}$. This imposes stronger restrictions on the correlation structure of the target. This is not an artifact of our proof technique, and a feature of the parallel CAVI algorithm itself, as we exhibit below through an example. 

Consider a class of target distributions $\pi_n \equiv N_d(0, Q^{-1})$ where the precision matrix $Q$ has a compound symmetry form, i.e., $Q = (1-\rho) \mr I_d + \rho 1_d 1_d'$. To ensure positive definiteness of $Q$, assume $-(d-1)^{-1} < \rho < 1$. Consider a mean-field approximation $q(\theta) = \prod_{j=1}^d q_j(\theta_j)$. The parallel update proceeds as 
\be 
q_j^{(t+1)}(\theta_j) = \m N(\theta_j; m_j^{(t+1)}, 1), \ m_j^{(t+1)} = - \rho \sum_{k \ne j} m_k^{(t)}, \ j \in [d]. 
\ee
Thus, following Remark \ref{rem:restrict}, it is sufficient to consider $q_j \equiv N(m_j, 1)$ to bound $\Delta_{j, n}$. We have, 
\be 
\Delta_{j, n}(q_j, q_{-j}) 
& = - \int (q_j - q_j^\star) (q_{-j} - q_{-j}^\star) \frac{1}{2} \big[ \theta_j^2 + \|\theta_{-j}\|^2 +2 \rho \sum_{k \ne j} \theta_j \theta_k \big] \\
& = - \rho \sum_{k \ne j} (m_j - m_j^\star)(m_k - m_k^\star) 
= - \rho \, (m_j-m_j^\star) \, 1_{d-1}' (m_{-j} - m_{-j}^\star).
\ee
By Cauchy--Schwarz inequality,  
\be 
|\Delta_{j, n}(q_j, q_{-j})| 
& \le |\rho| \, |m_j-m_j^\star| \, \sqrt{d-1} \, \|m_{-j} - m_{-j}^\star\| \\
&= 2 |\rho| \sqrt{d-1} \, \sqrt{D_{\rm KL, \alpha}(q_j\,||\,q_j^\star )} \, \sqrt{D_{\rm KL,1-\alpha}(q_{-j}\,||\,q_{-j}^\star)},
\ee
for any $\alpha \in (0, 1)$. Therefore, the condition $\mbox{\rm GCorr}_d(\pi_n) \in (0, 2/\sqrt{d-1})$ is satisfied if and only if $|\rho| < 1/(d-1)$. Thus, unlike the $d = 2$ case which covered all non-singular Gaussian distributions, there is a non-trivial restriction on the correlation structure within the compound symmetry class. 

We illustrate the above restriction is a feature of the parallel CAVI algorithm itself by conducting a direct analysis for the dynamical system for the mean vector $m^{(t)}$. It is easy to see that 
\be \label{eq:lin_dyn}
m^{(t+1)} = \rho (\mr I_d - 1_d 1_d') m^{(t)}. 
\ee
Since this is a linear dynamical system with a fixed coefficient matrix $A :\,= \rho (\mr I_d - 1_d 1_d')$, the system \eqref{eq:lin_dyn} globally converges if and only if the spectral radius (largest absolute value of the eigenvalues) of $A$ is smaller than $1$. Since the matrix $\mr I_d - 1_d 1_d'$ has eigenvalues $d-1, 1, \ldots, 1$, the spectral radius of $A$ is $|\rho| (d-1)$ and hence we obtain the same condition $|\rho| < 1/(d-1)$ for convergence. Thus, the parallel scheme itself fails to converge if $1/(d-1) \le \rho < 1$.

\section{Proofs of main results}\label{sec:proofs}
In this section, we present the proofs of the main results about general convergence analysis of CAVI in Sections~\ref{sec:conv_two_block} and \ref{sec:gtr_2b}. Proofs of the applications in Section~\ref{sec:examples} are deferred to the appendix.

\subsection{Proof of Lemma \ref{prop:LSC}}
We begin with the assertion in \eqref{eq:CAVI_opt}. To see this, consider initializing the sequential updating scheme \eqref{eq:seq_CAVI_gen} with $q_1^\star \otimes q_2^\star$. Since $D_{\rm KL}(q_1^{(1)} \otimes q_2^\star \,||\, \pi_n) \le D_{\rm KL}(q_1^\star \otimes q_2^\star \,||\, \pi_n)$, we must have that $q_1^{(1)} = q_1^\star$. By a similar argument, $q_2^{(1)} = q_2^\star$. The conclusion now follows from \eqref{eq:cavi_gen_form}. 
\\[1ex]
Next, we establish \eqref{Eqn:LSC}. We have, 
\begin{align*}
& F(q_1 \otimes q_2) - F(q_1^\star \otimes q_2^\star) 
= D_{\rm KL}(q_1 \otimes q_2\,||\, \pi_n)- D_{\rm KL}(q_1^\star \otimes q_2^\star\,||\, \pi_n)
\\=&
\int_{\m X_1} q_1\log q_1 - \int_{\m X_1} q_1^\star \log q_1^\star + \int_{\m X_2} q_2\log q_2 - \int_{\m X_2} q_2^\star \log q_2^\star
- \int_{\m X} (q_1q_2-q_1^\star q_2^\star )\log \pi_n\\
=& \int_{\m X_1} (q_1\log q_1 - q_1^\star \log q_1^\star) + \int_{\m X_2} (q_2\log q_2 - q_2^\star \log q_2^\star) 
- \int_{\m X} (q_1-q_1^\star)(q_2- q_2^\star )\log \pi_n\\
- & \int_{\m X} q_1^\star(q_2-q_2^\star) \log \pi_n -  \int_{\m X} (q_1-q_1^\star)q_2^\star \log \pi_n.
\end{align*}
Equation \eqref{eq:CAVI_opt} implies $\int_{\m X_{-j}} q_{-j}^\star \log \pi_n = \log q_j^\star + Z_j$ for $j = 1, 2$, and 
\begin{align*}
\int_{\m X} (q_j-q_j^\star) q_{-j}^\star  \log \pi_n = \int_{\m X_j} (q_j-q_j^\star) \log q_j^\star,
\end{align*} 
since $\int(q_j-q_j^\star) Z_j = Z_j \int(q_j-q_j^\star) =0$. Consequently, for $j = 1, 2$, 
\begin{align*}
\int_{\m X_j} (q_j\log q_j - q_j^\star \log q_j^\star) - \int_{\m X} (q_j-q_j^\star)q_{-j}^\star \log \pi_n = D_{\rm KL}(q_j \,|| \, q_j^\star).
\end{align*}
By combining the above identities, we arrive at \eqref{Eqn:LSC}. Equation \eqref{Eqn:LSC_spec} follows immediately from \eqref{Eqn:LSC}. 

\subsection{Proof of Theorem \ref{thm:CAVI_2block_par_alpha}}
We begin by formulating the first-order conditions for the respective updates. To that end, we introduce some notation related to those in Section \ref{sec:cons_fn}. For a fixed $q_2$, we shall denote by $\partial_1 F(q_1 \otimes q_2)$ the element $\dot{G}(q_1)$ in $L^2(\m X_1)$, where $G(\rho) = F(\rho \otimes q_2)$. Define $\partial_2 F(q_1 \otimes q_2)$ as an element of $L^2(\m X_2)$ in a similar fashion. It is straightforward to check; see Appendix \ref{sec:pf_app_aux}; that 
\be \label{eq:partial_ders}
\partial_j F(q_1 \otimes q_2) = \log q_j - \int_{\m X_{-j}} q_{-j} \log \pi_n, \quad j = 1, 2. 
\ee
The first-order optimality condition \eqref{eq:foc} for the two-block parallel CAVI algorithm is now given by 
\begin{align}\label{eq:foc_2bpar}
\begin{aligned}
\bigg \langle \partial_1 F(q_1^{(t+1)} \otimes q_2^{(t)}), (q_1 - q_1^{(t+1)}) \bigg \rangle & \ge 0,  \\
\bigg \langle \partial_2 F(q_1^{(t)} \otimes q_2^{(t+1)}), (q_2 - q_2^{(t+1)}) \bigg \rangle & \ge 0. 
\end{aligned}
\end{align}
for any $q_1 \in \m Q_1, q_2 \in \m Q_2$, and $t \ge 0$. We now show that for any $q_j \in \m Q_j$, $j = 1, 2$, 
\be
\big \langle \partial_1 F(q_1^{(t+1)} \otimes q_2^{(t)}),  (q_1 - q_1^{(t+1)}) \big \rangle &= F(q_1 \otimes q_2^{(t)}) - F(q_1^{(t+1)} \otimes q_2^{(t)}) - D_{\rm KL}(q_1 \,||\, q_1^{(t+1)}), \\
\big \langle \partial_2 F(q_1^{(t)} \otimes q_2^{(t+1)}),  (q_2 - q_2^{(t+1)}) \big \rangle &= F(q_1^{(t)} \otimes q_2) - F(q_1^{(t)} \otimes q_2^{(t+1)}) - D_{\rm KL}(q_2 \,||\, q_2^{(t+1)}). 
\ee
We only prove the first identity; the second one is similar. Setting $q = q_1 \otimes q_2^{(t)}$ and $\tilde{q} = q_1^{(t+1)} \otimes q_2^{(t)}$, the right hand side is $F(q) - F(\tilde{q}) - D_{\rm KL}(q \,||\, \tilde{q}) = \langle \dot{F}(\tilde{q}), q - \tilde{q} \rangle$ as shown in \eqref{eq:conv_F}. Substituting the values of $q$ and $\tilde{q}$, we get 
\be 
\langle \dot{F}(\tilde{q}), q - \tilde{q} \rangle 
& = \int_{\m X} \big(q_1 - q_1^{(t+1)}\big) \, q_2^{(t)} \, \log \frac{q_1^{(t+1)} \, q_2^{(t)}}{\pi_n}  \\
& = \int_{\m X_1} \big(q_1 - q_1^{(t+1)}\big)  \, \bigg[\log q_1^{(t+1)} + C  - \int _{\m X_2} q_2^{(t)} \, \log \pi_n + \log Z \bigg] \\
& = \int_{\m X_1} \big(q_1 - q_1^{(t+1)}\big) \, \bigg[\log q_1^{(t+1)}  - \int _{\m X_2} q_2^{(t)} \, \log \pi_n \bigg] \\
& = \int_{\m X_1} \big(q_1 - q_1^{(t+1)}\big) \, \partial_1 F(q_1^{(t+1)} \otimes q_2^{(t)}). 
\ee
where at the last step, we invoked \eqref{eq:partial_ders}. In the second line, $C = \int_{\m X_2} q_2^{(t)} \, \log q_2^{(t)}$. Since we have a difference of densities outside, the constants get annihilated from the second to the third line. 

Getting back to \eqref{eq:foc_2bpar}, set $q_1 = q_1^\star$ in the first inequality, and rearrange terms in light of the subsequent identity to get
\be 
& D_{\rm KL}(q_1^\star \,||\, q_1^{(t+1)}) \le F(q_1^\star \otimes q_2^{(t)}) - F(q_1^{(t+1)} \otimes q_2^{(t)})  \\
= & \big[F(q_1^\star \otimes q_2^{(t)}) - F(q_1^\star \otimes q_2^\star)\big] - [F(q_1^{(t+1)} \otimes q_2^{(t)}) - F(q_1^\star \otimes q_2^\star)\big] \\
= & D_{\rm KL}(q_2^{(t)} \,||\, q_2^\star) - \big[D_{\rm KL}(q_1^{(t+1)}\,||\,q_1^\star) + D_{\rm KL}(q_2^{(t)}\,||\,q_2^\star) - \Delta_n(q_1^{(t+1)}, q_2^{(t)}) \big].
\ee
Rearranging, we obtain
\be \label{eq:deln_tplus1_t}
D_{\rm KL}(q_1^\star \,||\, q_1^{(t+1)}) + D_{\rm KL}(q_1^{(t+1)}\,||\,q_1^\star) \le \Delta_n(q_1^{(t+1)}, q_2^{(t)}).
\ee
Repeating the same program with the second inequality of \eqref{eq:foc_2bpar}, we also obtain 
\be \label{eq:deln_t_tplus1}
D_{\rm KL}(q_2^\star \,||\, q_2^{(t+1)}) + D_{\rm KL}(q_2^{(t+1)}\,||\,q_2^\star) \le \Delta_n(q_1^{(t)}, q_2^{(t+1)}). 
\ee
We now bound $\Delta_n$ in \eqref{eq:deln_tplus1_t} and \eqref{eq:deln_t_tplus1}. 
Fix $\alpha \in [0, 1]$. By definition of $\mbox{\rm GCorr}(\pi_n)$, we have, for any $q_j \in \m Q_j$, 
\begin{align}\label{eq:deltan_double}
\begin{aligned}
|\Delta_n(q_1,\,q_2)| & \le \bar{\gamma}_{n, \alpha} \, \big( D_{\rm KL, \alpha}(q_1 \,||\, q_1^\star) \big)^{1/2} \, \big( D_{\rm KL, 1-\alpha}(q_2\,||\, q_2^\star) \big)^{1/2}, \\
|\Delta_n(q_1,\,q_2)| & \le \bar{\gamma}_{n, \alpha} \big( D_{\rm KL, 1-\alpha}(q_1 \,||\, q_1^\star) \big)^{1/2} \, \big( D_{\rm KL, \alpha}(q_2\,||\, q_2^\star) \big)^{1/2}. 
\end{aligned}
\end{align}
We use the following inequality multiple times over: for $\gamma, a, b > 0$, 
\be \label{eq:ab_ineq}
\gamma \sqrt{ab} = (\sqrt{2 a}) (\gamma \sqrt{b/2}) \le a + \frac{b \gamma^2}{4}. 
\ee
Consider \eqref{eq:deln_tplus1_t} first. Apply \eqref{eq:ab_ineq} to the first and second inequality of \eqref{eq:deltan_double} respectively to bound
\begin{align*} 
\Delta_n(q_1^{(t+1)}, q_2^{(t)}) 
\le & D_{\rm KL, \alpha}(q_1^{(t+1)} \,||\, q_1^\star) + \frac{\bar{\gamma}_{n, \alpha}^2}{4} D_{\rm KL, 1-\alpha}(q_2^{(t)}\,||\, q_2^\star), \\
\Delta_n(q_1^{(t+1)}, q_2^{(t)}) 
\le & D_{\rm KL, 1-\alpha}(q_1^{(t+1)} \,||\, q_1^\star) + \frac{\bar{\gamma}_{n, \alpha}^2}{4} D_{\rm KL, \alpha}(q_2^{(t)}\,||\, q_2^\star).
\end{align*}
Adding and cascading the resulting inequality through \eqref{eq:deln_tplus1_t}, we get 
\be \label{eq:par_12_symm}
D_{\rm KL, 1/2}(q_1^{(t+1)}\,||\,q_1^\star) \le \frac{\bar{\gamma}_{n, \alpha}^2}{4} D_{\rm KL, 1/2}(q_2^{(t)}\,||\, q_2^\star). 
\ee
Similarly, starting from \eqref{eq:deln_t_tplus1} and bounding $\Delta_n(q_1^{(t)}, q_2^{(t+1)})$ using \eqref{eq:ab_ineq} inside the second and first inequality of \eqref{eq:deltan_double}, we obtain
\be \label{eq:par_21_symm}
D_{\rm KL, 1/2}(q_2^{(t+1)}\,||\,q_2^\star) \le \frac{\bar{\gamma}_{n, \alpha}^2}{4} D_{\rm KL, 1/2}(q_1^{(t)}\,||\, q_1^\star). 
\ee
Adding \eqref{eq:par_12_symm} and \eqref{eq:par_21_symm}, we get 
\be \label{eq:par_symm}
D_{\rm KL, 1/2}(q^{(t+1)}\,||\,q^\star) \le \frac{\bar{\gamma}_{n, \alpha}^2}{4} D_{\rm KL, 1/2}(q^{(t)}\,||\, q^\star).
\ee
Since $\alpha$ is arbitrary, taking an infimum over $\alpha \in [0, 1]$, we get the desired result. Note that the condition $\mbox{\rm GCorr}(\pi_n) < 2$ is only used to guarantee that $\kappa_n \in (0, 1)$. 

\subsection{Proof of Theorem \ref{thm:CAVI_2block_par_alpha_init}}
We prove this by induction. Suppose for some $t$, $q_j^{(t)}$ satisfies $D_{\rm KL,1/2}(q_j^{(t)}\,||\,q_j^\star) \le r_0/2$ for $j = 1,2$. Then, we show that 
\begin{align}\label{eq:induc_step}
\begin{aligned}
D_{\rm KL, 1/2}(q_1^{(t+1)}\,||\,q_1^\star) &\le \frac{\mbox{\rm GCorr}^2(\pi_n)}{4} D_{\rm KL, 1/2}(q_2^{(t)}\,||\, q_2^\star), \\
D_{\rm KL, 1/2}(q_2^{(t+1)}\,||\,q_2^\star) &\le \frac{\mbox{\rm GCorr}^2(\pi_n)}{4} D_{\rm KL, 1/2}(q_1^{(t)}\,||\, q_1^\star).
\end{aligned}
\end{align}
This in particular implies $D_{\rm KL,1/2}(q_j^{(t+1)}\,||\,q_j^\star) \le r_0/2$ for $j = 1,2$, implying the induction hypothesis continues from $t$ to $t+1$. Also, the induction hypothesis is clearly satisfied at $t=0$ by assumption. Thus, \eqref{eq:induc_step} is true for all $t \ge 0$. The conclusion of the theorem follows by adding the two equations. 

We now establish \eqref{eq:induc_step}. From $D_{\rm KL,1/2}(q_j^{(t)}\,||\,q_j^\star) \le r_0/2$, we can conclude that $D_{\rm KL}(q_j^{(t)} \,||\, q_j^\star) \le r_0$ and $D_{\rm KL}(q_j^{\star} \,||\, q_j^{(t)}) \le r_0$, $(j = 1, 2)$. The second inequality in particular implies $q_j^{(t)} \in Q_j^\star(r_0)$, $(j=1, 2)$. From Lemma \ref{prop:LSC}, we have $\Delta_n(q_1,\,q_2)\leq D_{\rm KL}(q_1\,||\,q_1^\star) + D_{\rm} (q_2\,||\,q_2^\star)$ for any $q_1$ and $q_2$, implying
\be 
D_{\rm KL}(q_1^\star \,||\, q_1^{(t+1)}) + D_{\rm KL}(q_1^{(t+1)}\,||\,q_1^\star) \le \Delta_n(q_1^{(t+1)}, q_2^{(t)}) \le D_{\rm KL}(q_1^{(t+1)}\,||\,q_1^\star) + D_{\rm KL}(q_2^{(t)}\,||\,q_2^\star),
\ee
where the first inequality is the same as equation \eqref{eq:deln_tplus1_t} (inside the proof of Theorem \ref{thm:CAVI_2block_par_alpha}) which only uses the first-order optimality of $q_1^{(t+1)}$. We then get $D_{\rm KL}(q_1^\star \,||\, q_1^{(t+1)}) \le D_{\rm KL}(q_2^{(t)}\,||\,q_2^\star) \le r_0$, implying $q_1^{(t+1)} \in \m Q_1^\star(r_0)$. Combined with the fact that $q_2^{(t)} \in \m Q_2^\star(r_0)$, we can now bound $|\Delta_n(q_1^{(t+1)}, q_2^{(t)})|$ as in \eqref{eq:deltan_double} (with the modified definition of $\bar{\gamma}_{n, \alpha}$) and repeat the same steps to arrive at the bound in \eqref{eq:par_12_symm}. Taking an infimum over $\alpha$ delivers the first inequality of \eqref{eq:induc_step}. The second inequality follows in a similar fashion.

\subsection{Proof of Theorem \ref{thm:CAVI_2block_rand_alpha_init}}
We have 
\begin{align*}
q^{(t+1)} = 
\begin{cases}
\rho_1^{(t+1)} \otimes q_2^{(t)} & \text{ with prob. } \frac{1}{2}, \\
q_1^{(t)} \otimes \rho_2^{(t+1)} & \text { with prob. } \frac{1}{2},
\end{cases}
\end{align*}
where $\rho_j^{(t+1)} :\,= \argmin{q_j} F(q_j \otimes q_{-j}^{(t)})$ is the usual parallel update for $q_j$ at time $(t+1)$. Define $\rho^{(t+1)} = \rho_1^{(t+1)} \otimes \rho_2^{(t+1)}$. We have, for any $\alpha \in [0, 1]$,
\begin{align*}
\mb E[D_{\rm KL,1/2}(q^{(t+1)}\,||\,q^\star) \mid \m F_t] 
& = \frac{1}{2} \bigg[ D_{\rm KL,1/2}(\rho_1^{(t+1)} \otimes q_2^{(t)} \,||\,q^\star) + D_{\rm KL,1/2}(q_1^{(t)} \otimes \rho_2^{(t+1)} \,||\,q^\star) \bigg] \\
& = \frac{1}{2} \bigg[ D_{\rm KL,1/2}(\rho^{(t+1)} \,||\, q^\star) + D_{\rm KL,1/2}(q^{(t)}\,||\,q^\star) \bigg] \\
& \le \bigg[\frac{1}{2} + \frac{1}{2} \, \frac{\bar{\gamma}_{n, \alpha}^2}{4} \bigg] \, D_{\rm KL,1/2}(q^{(t)}\,||\,q^\star). 
\end{align*}
We used the tensorization property of the KL divergence going from the first to the second line, and then invoked the bound \eqref{eq:par_symm} to bound $D_{\rm KL,1/2}(\rho^{(t+1)} \,||\, q^\star)$. The conclusion now follows by taking an infimum over $\alpha$. 

\subsection{Proof of Corollary \ref{cor:2block_par_onesided}}
We prove this by induction. Suppose for some $t$, $D_{\rm KL}(q_j^{(t)} \,||\, q_j^\star) \le r_0, (j=1,2)$. From the first-order condition \eqref{eq:deln_tplus1_t}, we know that $D_{\rm KL}(q_1^{(t+1)} \,||\, q_1^\star) + D_{\rm KL}(q_1^\star \,||\, q_1^{(t+1)}) \le \Delta_n(q_1^{(t+1)}, q_2^{(t)})$. Now, bound $\Delta_n(q_1^{(t+1)}, q_2^{(t)})$ using the first inequality of \eqref{eq:2block_par_onesided} and rearrange terms to get $(1 - \gamma_n/2) D_{\rm KL}(q_1^{(t+1)} \,||\, q_1^\star) + D_{\rm KL}(q_1^\star \,||\, q_1^{(t+1)}) \le (\gamma_n/2) D_{\rm KL}(q_2^{(t)} \,||\, q_2^\star)$, implying 
\be
D_{\rm KL}(q_1^{(t+1)} \,||\, q_1^\star) \le \frac{\gamma_n}{2-\gamma_n} \, D_{\rm KL}(q_2^{(t)} \,||\, q_2^\star).
\ee
Similarly, combining \eqref{eq:deln_t_tplus1} with the second inequality of \eqref{eq:2block_par_onesided}, get 
\be 
D_{\rm KL}(q_2^{(t+1)} \,||\, q_2^\star) \le \frac{\gamma_n}{2-\gamma_n} \, D_{\rm KL}(q_2^\star \,||\, q_2^{(t)})
\ee
Adding the two displayed equations, we get $D_{\rm KL}(q^{(t+1)} \,||\, q^\star) \le \kappa_n D_{\rm KL}(q^{(t)} \,||\, q^\star)$. This in particular implies $D_{\rm KL}(q^{(t+1)} \,||\, q^\star) \le r_0$, meaning the induction hypothesis continues from $t$ to $t+1$. The condition on the initialization means the induction hypothesis is satisfied at $t = 0$. This completes the proof. 
\subsection*{Proof of Lemma \ref{lem:LSC_gen}}
The proof of the stationary condition is essentially the same as the $d=2$ case. We therefore only prove the second part. 

Observe that 
\be 
F(q_j \otimes q_{-j}) 
& = \int_{\m X} (q_j \otimes q_{-j}) \log(q_j \otimes q_{-j}) - \int_{\m X} (q_j \otimes q_{-j}) \log \pi_n \\
& = \int_{\m X_j} q_j \log q_j + \int_{\m X_{-j}} q_{-j} \log q_{-j} - \int_{\m X} (q_j \otimes q_{-j}) \log \pi_n.
\ee
This gives, 
\be 
F(q_j^\star \otimes q_{-j}) -  F(q_j \otimes q_{-j}) = \int_{\m X_j} q_j^\star \log q_j^\star - \int_{\m X_j} q_j \log q_j + \int_{\m X} (q_j - q_j^\star)\otimes q_{-j} \log \pi_n. 
\ee
We also have, 
\be 
\Delta_{j.n}(q_j, q_{-j}) 
& = \int_{\m X} (q_j - q_j^\star)\otimes q_{-j} \log \pi_n - \int_{\m X} (q_j - q_j^\star)\otimes q_{-j}^\star \log \pi_n \\
& = \int_{\m X} (q_j - q_j^\star)\otimes q_{-j} \log \pi_n - \int_{\m X_j} (q_j - q_j^\star) \bigg[\int_{\m X_{-j}} q_{-j}^\star \log \pi_n\bigg] \\
& =  \int_{\m X} (q_j - q_j^\star)\otimes q_{-j} \log \pi_n - \int_{\m X_j} (q_j - q_j^\star) \log q_j^\star.
\ee
Here, going from the first to the second line, we used Tonelli's theorem to write the joint integral as an iterated one, and in the next step, used the stationarity condition $q_j^\star \,\propto\, \exp \big(\int q_{-j}^\star \log \pi_n \big)$. Equating the previous two identities, we get,
\be 
F(q_j^\star \otimes q_{-j}) -  F(q_j \otimes q_{-j})
& =  \int_{\m X_j} q_j^\star \log q_j^\star - \int_{\m X_j} q_j \log q_j  + \int_{\m X_j} (q_j - q_j^\star) \log q_j^\star + \Delta_{j.n}(q_j, q_{-j})   \\
& = -D_{\rm KL}(q_j \,||\, q_j^\star) + \Delta_{j.n}(q_j, q_{-j}).
\ee

\subsection*{Proof of Theorem \ref{thm:CAVI_dblock_par_alpha}}
Analogous to Theorem \ref{thm:CAVI_2block_par_alpha}, we define $\partial_j F(q_j \otimes q_{-j})$ to be $\dot{G}_j(\rho)$, where $G_j(\rho) = G(\rho \otimes q_{-j})$. It can be shown that 
\be \label{eq:partial_ders_d}
\partial_j F(q_j \otimes q_{-j}) = \log q_j - \int_{\m X_{-j}} q_{-j} \log \pi_n, \quad j \in [d]. 
\ee
See Appendix \ref{sec:pf_app_aux} for a derivation. The first-order optimality condition \eqref{eq:foc} for the sub-problem $q_j^{(t+1)} = \argmin_{q_j} F(q_j \otimes q_{-j}^{(t)})$ is given by
\be \label{eq:foo_d}
\bigg \langle \partial_j F(q_j^{(t+1)} \otimes q_{-j}^{(t)}), (q_j - q_j^{(t+1)}) \bigg \rangle \ge 0,
\ee
for any $q_j \in \m Q_j$. Extending a similar argument from the proof of Theorem \ref{thm:CAVI_2block_par_alpha}, see Appendix \ref{sec:pf_app_aux} for a detailed derivation, the left hand of \eqref{eq:foo_d} simplifies to
\begin{align} \label{eq:deljF_ip}
\left \langle \partial_j F(q_j^{(t+1)} \otimes q_{-j}^{(t)}),  (q_j - q_j^{(t+1)}) \right \rangle &= F(q_j \otimes q_{-j}^{(t)}) - F(q_j^{(t+1)} \otimes q_{-j}^{(t)}) - D_{\rm KL}(q_j \,||\, q_j^{(t+1)}).
\end{align}
Setting $q_j = q_j^\star$ in the previous display, 
\be 
D_{\rm KL}(q_j^\star \,||\, q_j^{(t+1)}) \le F(q_j^\star \otimes q_{-j}^{(t)}) - F(q_j^{(t+1)} \otimes q_{-j}^{(t)}). 
\ee
Now, from Lemma \ref{lem:LSC_gen}, 
\be \label{eq:LSC_d}
F(q_j^\star \otimes q_{-j}^{(t)}) - F(q_j^{(t+1)} \otimes q_{-j}^{(t)}) = - D_{\rm KL}(q_j^{(t+1)} \,||\, q_j^\star) + \Delta_{j, n}(q_j^{(t+1)}, q_{-j}^{(t)}). 
\ee
Combining, we get 
\be\label{eq:deljn_d}
D_{\rm KL}(q_j^\star \,||\, q_j^{(t+1)}) + D_{\rm KL}(q_j^{(t+1)} \,||\, q_j^\star)  \le  \Delta_{j, n}(q_j^{(t+1)}, q_{-j}^{(t)}), \quad j \in [d]. 
\ee
By definition of $\bar{\gamma}_{n, \alpha}^{(j)}$ in the display before \eqref{eq:GCorr_d}, we have, for any $\alpha \in [0, 1]$, $j \in [d]$, $q_j \in \m Q_j$, and $q_{-j} \in \m Q_{-j}$, 
\begin{align}\label{eq:delta_jn_double}
\begin{aligned}
|\Delta_{j,n}(q_j,\,q_{-j})| & \le \bar{\gamma}_{n, \alpha}^{(j)} \, \big( D_{\rm KL, \alpha}(q_j \,||\, q_j^\star) \big)^{1/2} \, \big( D_{\rm KL, 1-\alpha}(q_{-j}\,||\, q_{-j}^\star) \big)^{1/2}, \\
|\Delta_{j, n}(q_j,\,q_{-j})| & \le \bar{\gamma}_{n, \alpha}^{(j)} \big( D_{\rm KL, 1-\alpha}(q_j \,||\, q_j^\star) \big)^{1/2} \, \big( D_{\rm KL, \alpha}(q_{-j}\,||\, q_{-j}^\star) \big)^{1/2}. 
\end{aligned}
\end{align}
Fix $j \in [d]$. Apply \eqref{eq:ab_ineq} to either inequality of \eqref{eq:delta_jn_double} respectively to bound
\begin{align*} 
\Delta_{j,n}(q_j^{(t+1)}, q_{-j}^{(t)}) 
\le & D_{\rm KL, \alpha}(q_j^{(t+1)} \,||\, q_j^\star) + \frac{(\bar{\gamma}_{n, \alpha}^{(j)})^2}{4} D_{\rm KL, 1-\alpha}(q_{-j}^{(t)}\,||\, q_{-j}^\star), \\
\Delta_{j,n}(q_j^{(t+1)}, q_{-j}^{(t)}) 
\le & D_{\rm KL, 1-\alpha}(q_j^{(t+1)} \,||\, q_j^\star) + \frac{(\bar{\gamma}_{n, \alpha}^{(j)})^2}{4} D_{\rm KL, \alpha}(q_{-j}^{(t)}\,||\, q_{-j}^\star).
\end{align*}
Adding the two inequalities further gives, for any $j \in [d]$, 
\be 
\Delta_{j,n}(q_j^{(t+1)}, q_{-j}^{(t)}) \le D_{\rm KL, 1/2}(q_j^{(t+1)} \,||\, q_j^\star) + \frac{(\bar{\gamma}_{n, \alpha}^{(j)})^2}{4} D_{\rm KL, 1/2}(q_{-j}^{(t)}\,||\, q_{-j}^\star). 
\ee
Combining with \eqref{eq:deljn_d}, one obtains
\be 
D_{\rm KL, 1/2}(q_j^{(t+1)} \,||\, q_j^\star) \le \frac{(\bar{\gamma}_{n, \alpha}^{(j)})^2}{4} D_{\rm KL, 1/2}(q_{-j}^{(t)}\,||\, q_{-j}^\star). 
\ee
Since the above is true for any $\alpha \in [0, 1]$, we ca take an infimum over $\alpha$ to obtain
\be 
D_{\rm KL, 1/2}(q_j^{(t+1)} \,||\, q_j^\star) 
& \le \frac{\{\mbox{\rm GCorr}^{(j)}(\pi_n)\}^2}{4} D_{\rm KL, 1/2}(q_{-j}^{(t)}\,||\, q_{-j}^\star) \\& 
\le \frac{\{\mbox{\rm GCorr}_d(\pi_n)\}^2}{4} D_{\rm KL, 1/2}(q_{-j}^{(t)}\,||\, q_{-j}^\star), 
\ee
where the second inequality simply uses the definition of $\mbox{\rm GCorr}_d(\pi_n)$. Summing both sides over $j$, we get the desired conclusion, since 
\be 
\sum_{j=1}^d D_{\rm KL, 1/2}(q_{-j}^{(t)}\,||\, q_{-j}^\star) = \sum_{j=1}^d \sum_{k \ne j} D_{\rm KL, 1/2}(q_k^{(t)}\,||\,q_k^\star) = (d-1) D_{\rm KL, 1/2}(q^{(t)} \,||\, q^\star). 
\ee

\section{Discussion}\label{sec:discussion}
In this article, we studied the convergence of two-block CAVI for implementing MF variational approximation, and introduced the notion of a generalized correlation to quantify the contraction rate. We take a functional optimization perspective to view the CAVI algorithm as optimization on density space, and use specific parametric nature of the updates to bound the generalized correlation, instead of directly analyzing possibly highly non-linear dynamical systems for the parameter updates. Moreover, one can exploit statistical concentration properties of the VB posterior while bounding the generalized correlation. A number of illustrations have been provided for standard statistical models highlighting these features. 

We have also provided initial results for CAVI with more than two blocks, where we extend the notion of generalized correlation for characterizing the interactions among different blocks by considering the largest generalized correlation between any block and the collapsed block consisting of the rest.
Different from the (parallel) two-block CAVI that always converges for approximating any non-singular Gaussian distribution (Proposition~\ref{prop:two_block_g}), the general multi-block CAVI may fail to converge for some highly correlated multi-variate Gaussian distribution due to overly aggressive updates of the blocks that overshoot the highest probability regions. We illustrate this for the compound symmetry covariance structure in Section \ref{sec:gtr_2b}. An interesting future direction to pursue is to investigate whether some proper notion of step size can be incorporated to enforce the convergence of CAVI. In particular, we are investigating a ``lazy" version of the CAVI
\begin{align*}
    q_j^{(t+1)} \,\propto \,\,\big(q_j^{(t)}\big)^{1-\alpha} \cdot \big(q^{(t+1)}_{{\rm full},j}\big)^{\alpha}, \quad j\in[m], \quad t=0,1,\ldots,
\end{align*}
for some partial step size $\alpha\in(0,1)$, where $q^{(t+1)}_{{\rm full},j}$ denotes the full step size CAVI (i.e.~$\alpha=1$), such as the one given by formulation~\eqref{eq:seq_CAVI_gen} in the sequential CAVI. Introducing such a partial step size is particularly meaningful in the more than two blocks case as at least for multivariate Gaussian targets, we can show that the stringent condition (Theorem~\ref{thm:CAVI_dblock_par_alpha} with large $d$) for the full step size CAVI to convergence can be substantially relaxed. We wish to report more on this in a separate avenue. In addition, in the two-block case, the three updating schemes, i.e.~parallel, sequential, and randomized updates, behaves very similarly as their contraction rates usually differ by at most a constant order; it would be interesting to study and compare them in the multi-block case.


\bibliographystyle{plain}
\bibliography{CAVI}

\newpage
\appendix
\begin{center}
{\bf\Large Appendix}
\end{center}

\section{Some auxiliary results}\label{sec:app_aux}
We record some facts and auxiliary results which are used in the proofs. Proofs of lemmas stated in this section can be found in Appendix C.  
\\[1ex]
{\bf KL and symmetric KL divergences for some standard families.} \\[1ex]
(i) For $p_i \equiv \mbox{Gamma}(a, b_i), (i = 0, 1)$,  
\begin{align} \label{eq:gamma_kl}
\begin{aligned}
D_{\rm KL}(p_0 \,||\, p_1) & = a \big( b_1/b_0 - 1 - \log(b_1/b_0) \big), \\
D_{\rm KL, 1/2}(p_0\,||\,p_1) & = \frac{a}{2}\, \frac{(b_1-b_0)^2}{b_0 b_1}. 
\end{aligned}
\end{align}
\\[1ex]
(ii) For $p_i \equiv N(\mu_i, \tau_i^{-1}), (i=0,1)$, 
\begin{align} \label{eq:gauss_kl}
\begin{aligned}
D_{\rm KL}(p_0 \,||\, p_1) & = \frac{1}{2} \big[\tau_1/\tau_0 -  \log\big(\tau_1/\tau_0\big) - 1\big] + \frac{\tau_1(\mu_1 - \mu_0)^2}{2}, \\
D_{\rm KL,1/2}(p_0 \,||\, p_1) & = \frac{1}{4} \, \frac{(\tau_1 - \tau_0)^2}{\tau_0 \tau_1} + \frac{(\tau_0 + \tau_1)(\mu_1 - \mu_0)^2}{4}.  
\end{aligned}
\end{align}
\\[1ex]
(iii) For $p_i \equiv N_d(\mu_i, \Sigma_i), i = 0, 1$,
\begin{align} \label{eq:mvn_kl}
D_{\rm KL}(p_0 \,||\, p_1) 
& =  \frac{1}{2} \big( \mbox{tr}(\Sigma_1^{-1} \Sigma_0) - \log \mbox{det}(\Sigma_1^{-1} \Sigma_0) - d + (\mu_1 - \mu_0)' \Sigma_1^{-1} (\mu_1 - \mu_0) \big) \notag \\
& = \frac{1}{2} \bigg[ \sum_{j=1}^d \big(\lambda_j(\Delta) - 1 - \log \lambda_j(\Delta)\big) + (\mu_1 - \mu_0)' \Sigma_1^{-1} (\mu_1 - \mu_0) \big) \bigg],
\end{align}
where $\Delta = \Sigma_1^{-1/2} \Sigma_0 \Sigma_1^{-1/2}$ is pd and $\{\lambda_j(\Delta)\}_{j=1}^d$ denote its eigenvalues. 
\\[1ex]
(iv) Let $TN_{\m C}(\mu, \sigma^2)$ denotes a $N(\mu, \sigma^2)$ distribution truncated to region $\m C \subseteq \mb R$. Suppose $p_i \equiv TN_{\m C}(a_i, 1)$ for $i = 0, 1$ with density 
\be 
p_i(x) = \frac{\phi(x - a_i)}{Z_i} \, \ind_{\m C}(x), \quad Z_i = \int_{\m C - a_i} \phi(t) dt, 
\ee
where $\m C - a = \{x - a \,:\, x \in \m C\}$. Then, we have,
\be \label{eq:tmvn_kl}
D_{\rm KL}(p_0 \,||\, p_1) = (a_0 - a_1) \bigg[E_{p_0}(X) - \frac{a_0+a_1}{2} \bigg] - \log \frac{Z_0}{Z_1}.
\ee
It follows from the previous equation that  
\be \label{eq:tmvn_skl}
D_{\rm KL,1/2}(p_0 \,||\, p_1) = \frac{1}{2} \, (a_0 - a_1) [E_{p_0}(X) - E_{p_1}(X)]. 
\ee
As in the main document, let $TN_1$ and $TN_0$ denote truncated normal distributions with truncation region $\m C = (0, \infty)$ and $(-\infty, 0)$, respectively. We state a useful contraction result about means of TN distributions. 
\begin{lemma}\label{lem:tn_contract}
Fix $y \in \{0, 1\}$. Suppose $p_i \equiv TN_{y}(a_i, 1)$ and let $b_i = E_{X \sim p_i}(X)$ for $i = 0, 1$. Then, $|b_0 - b_1| \le |a_0 - a_1|$. 
\end{lemma}

\noindent {\bf An inequality for sub-Gaussian random variables.}
\begin{lemma}\label{lem:subg_mom}
Let $Z$ be a sub-Gaussian random variable with $\mb E [e^{t(Z-\mb E[Z])}] \le e^{\nu^2 t^2/2}$ for any $t \in \mb R$. Then, for any $m \in \mb R$ and $k \in \mb N$, $\mb E |Z - m|^{k} \le C \big[ |\mb E[Z] - m|^{k} + \nu^{k}]$, where $C$ is a global constant free of $m$ and $\nu$. 
\end{lemma}

\section{Proofs related to applications}

\subsection*{Proof of Proposition \ref{prop:two_block_g}}
Recall from the discussion after Proposition \ref{prop:two_block_g} that $\bar{\m Q}_j$ consists of densities $q_j \equiv N(m_j, (n Q_{jj})^{-1})$ with $m_j \in \mb R^{p_j} (j = 1, 2)$. We have, for any such $q_1, q_2$,  
\begin{align*}
\Delta_n(q_1, q_2) 
= & - n \int (\theta_1 - \theta_{01})' Q_{12} (\theta_2 - \theta_{02}) \, (q_1 - q_1^\star) (q_2 - q_2^\star)\\
= & -n \bigg[\int (\theta_1 - \theta_{01})' \, (q_1 - q_1^\star) \bigg] \, Q_{12} \bigg[\int (\theta_2 - \theta_{02}) \, (q_2 - q_2^\star)\bigg] \\
=  & - n \delta_1' Q_{12} \delta_2,
\end{align*}
where $\delta_j = \mb E_{q_j}[\theta_j] - \mb E_{q_j^\star}[\theta_j] = m_j - m_j^\star$ for $j = 1, 2$. Also, 
using identity \eqref{eq:mvn_kl}, 
\be 
D_{\rm KL}(q_j \,||\, q_j^\star) = 
D_{\rm KL}(q_j^\star \,||\, q_j) = 
\frac{ n \, \delta_j' Q_{jj} \delta_j }{2} = \frac{n \|Q_{jj}^{1/2} \delta_j\|^2}{2}, \quad j = 1, 2. 
\ee
This implies, for any $\alpha \in [0, 1]$, 
\be 
D_{\rm KL, \alpha}(q_j \,||\, q_j^\star) = \frac{n \|Q_{jj}^{1/2} \delta_j\|^2}{2}. 
\ee
We therefore have, for any $q_j \in \bar{\m Q}_j$, 
\begin{align}\label{eq:Gauss_case}
\begin{aligned}
|\Delta_n(q_1, q_2|  
& = n \left | \delta_1' Q_{11}^{1/2} \, (Q_{11}^{-1/2} Q_{12} Q_{22}^{-1/2}) \, Q_{22}^{1/2} \delta_2 \right | \\
& \le n \|Q_{11}^{1/2} \delta_1\| \, \|Q_{22}^{1/2} \delta_2\| \, \|Q_{11}^{-1/2} Q_{12} Q_{22}^{-1/2}\|_2 \\ 
& = 2 \|Q_{11}^{-1/2} Q_{12} Q_{22}^{-1/2}\|_2  \, \sqrt{ D_{\rm KL, \alpha}(q_1 \,||\, q_1^\star)  } \, \sqrt{ D_{\rm KL, 1-\alpha}(q_2\,||\,q_2^\star)  } \\
& = 2 \|Q_{11}^{-1/2} Q_{12} Q_{22}^{-1/2}\|_2  \, \sqrt{ D_{\rm KL, 1-\alpha}(q_1 \,||\, q_1^\star) } \, \sqrt{ D_{\rm KL, \alpha}(q_2\,||\,q_2^\star)  },
\end{aligned}
\end{align}
where the inequality follows from sub-multiplicativity of the spectral norm. Since this inequality is sharp, it follows from the last two lines of \eqref{eq:Gauss_case} that $\bar{\gamma}_{n, \alpha} = 2 \|Q_{11}^{-1/2} Q_{12} Q_{22}^{-1/2}\|_2$ for any $\alpha \in [0, 1]$.
We show below that for any $Q$ positive definite, $\|Q_{11}^{-1/2} Q_{12} Q_{22}^{-1/2}\|_2 < 1$, implying $\mbox{\rm GCorr}(\pi_n) \in (0, 2)$. 

To see why $\|Q_{11}^{-1/2} Q_{12} Q_{22}^{-1/2}\|_2 < 1$, consider $(X; Y) \sim N(0, Q)$, and define $U = Q_{11}^{-1/2} X$ and $V = Q_{22}^{-1/2} Y$. Then, $\mbox{cov}(U) = \mr I_{p_1}$, $\mbox{cov}(V) = \mr I_{p_2}$, and $\mbox{cov}(U, V) = Q_{11}^{-1/2} Q_{12} Q_{22}^{-1/2}$. The spectral norm of $\mbox{cov}(U, V)$ is the supremum of $\alpha' \mbox{cov}(U, V) \beta$ over all $\alpha \in \m S^{p_1-1}$ and $\beta \in \m S^{p_2-1}$. For any such $\alpha, \beta$, observe that $\alpha' \mbox{cov}(U, V) \beta = \mbox{cov}(\alpha'U, \beta'V) < \sqrt{\mbox{var}(\alpha' U)} \, \sqrt{\mbox{var}(\beta' V)} = 1$ by Cauchy--Schwarz inequality. This proves the claim. 

\subsection*{Proof of Proposition \ref{prop:gauss_conds}}
From the form of the updates, it is evident that $q_j^\star = N(0,1/\tau_j^\star)$. 
Directly solving the fixed point equation, we get $\tau_1^\star = \tau_2^\star = \tau^\star :\,= (1+\sqrt{5})/2 > 1$, the only positive solution to $x^2 - x - 1 = 0$. Also, we can take $\bar{\m Q}_j = \{q_j \equiv N(0,1/\tau_j) \,:\, \tau_j > 1\}$. We have, for $q_j \equiv N(0, 1/\tau_j) \in \bar{\m Q}_j$, 
\be 
\Delta_n(q_1, q_2) 
&= -\frac{1}{2} \bigg(\int u_1^2 (q_1(u_1) - q_1^\star(u_1)) \bigg) \,  \bigg(\int u_2^2 (q_2(u_2) - q_2^\star(u_2)) \bigg) \\
&= -\frac{1}{2} \, \bigg(\frac{1}{\tau_1} - \frac{1}{\tau_1^\star}\bigg) \,\bigg(\frac{1}{\tau_2} - \frac{1}{\tau_2^\star}\bigg). 
\ee
We also have from \eqref{eq:gauss_kl} that
\be 
D_{\rm KL, 1/2}(q_j\,||\,q_j^\star) = \frac{1}{4} \frac{(\tau_j - \tau_j^\star)^2}{\tau_j \tau_j^\star}. 
\ee
Then, 
\be 
\frac{|\Delta_n(q_1, q_2)|}{\sqrt{ D_{\rm KL, 1/2}(q_1\,||\,q_1^\star) } \, \sqrt{ D_{\rm KL, 1/2}(q_2\,||\,q_2^\star) }} = 2 \frac{1}{\sqrt{\tau_1 \tau_1^\star} \, \sqrt{\tau_2 \tau_2^\star}} = 2 \frac{1}{\tau^\star} \, \frac{1}{\sqrt{\tau_1 \tau_2}}. 
\ee
Since $\tau_j > 1$ for any $q_j \in \bar{\m Q}_j$, it follows that $\mbox{GCorr}_{1/2}(\pi_n) \le 2/\tau^\star = 4/(1+\sqrt{5})$.


\subsection*{Proof of Proposition \ref{prop:probit}}
Recall $\Sigma = (X'X + \kappa I_p)^{-1}$. Given the nature of the parallel updates in \eqref{eq:probit_params_update} and the preceding equation, it suffices to restrict to $q_{\blds \beta} \equiv N(m, \Sigma)$ with $m \in \mb R^d$ and $q_{\blds z} = \otimes_{i=1}^n q_i$ with $q_i \equiv TN_{y_i}(\alpha_i, 1)$ for $\alpha_i \in \mb R$ to bound $\Delta_n$, in light of Remark \ref{rem:restrict}.
We have, for any such $q_{\blds \beta}, q_{\blds z}$ that
\be 
\Delta_n(q_{\blds \beta}, q_{\blds z})
& = \int -\frac{1}{2} \sum_{i = 1}^n (z_i - x_i'\beta)^2 \, (q_{\blds \beta} - q_{\blds \beta}^\star) (q_{\blds z} - q_{\blds z}^\star) \\
& = \int \sum_{i=1}^n z_i x_i' \beta (q_{\blds \beta} - q_{\blds \beta}^\star) (q_{\blds z} - q_{\blds z}^\star) \\
& =  \sum_{i=1}^n \big(x_i' m - x_i' m^\star \big) \big(E_{q_i}(z_i) - E_{q_i^\star}(z_i) \big) \\
& = \langle X'(m-m^\star), \big(E_q(z) - E_{q^\star}(z)\big) \rangle.
\ee
Also, from \eqref{eq:mvn_kl} and \eqref{eq:tmvn_skl}, 
\be 
D_{\rm KL,1/2}(q_{\blds \beta} \,||\, q_{\blds \beta}^\star) & = \frac{(m - m^\star)' \Sigma^{-1} (m - m^\star)}{2}, \\
D_{\rm KL,1/2}(q_{\blds z} \,||\, q_{\blds z}^\star) & = \frac{1}{2} \sum_{i=1}^n (\alpha_i - \alpha_i^\star)\big(E_{q_i}(z_i) - E_{q_i^\star}(z_i) \big). 
\ee
Now, using Cauchy--Schwarz inequality, bound
\be \label{eq:cs_probit}
\Delta_n^2(q_{\blds \beta}, q_{\blds z}) \le \|X'(m-m^\star)\|^2 \, \|E_q(z) - E_{q^\star}(z)\|^2. 
\ee
Further, bound 
\be 
\|X(m-m^\star)\|^2 
& = (m - m^\star)' X'X (m - m^\star)  \\
& = (m - m^\star)' \Sigma^{-1/2} (\Sigma^{1/2} X'X \Sigma^{1/2}) \Sigma^{-1/2} (m - m^\star) \\
& \le \lambda_{\max}(\Sigma^{1/2} X'X \Sigma^{1/2}) (m - m^\star)'\Sigma^{-1}(m - m^\star) \\
& = 2 \lambda_{\max}(\Sigma^{1/2} (X'X) \Sigma^{1/2})  \, D_{\rm KL,1/2}(q_{\blds \beta} \,||\, q_{\blds \beta}^\star).
\ee
Here, going from the second to the third line, we used $x' A x \le \lambda_{\max}(A) \|x\|^2$ for any positive definite matrix $A$. 
Next, bound 
\be 
\|E_q(z) - E_{q^\star}(z)\|^2 
& = \sum_{i=1}^n |E_{q_i}(z_i) - E_{q_i^\star}(z_i)|^2 \\
& \le \sum_{i=1}^n |\alpha_i - \alpha_i^\star| \, |E_{q_i}(z_i) - E_{q_i^\star}(z_i)| \\
& = \sum_{i=1}^n (\alpha_i - \alpha_i^\star)\big(E_{q_i}(z_i) - E_{q_i^\star}(z_i) \big) \\
& = 2 D_{\rm KL,1/2}(q_{\blds z} \,||\, q_{\blds z}^\star). 
\ee
Here, going from the first to the second line, we invoked Lemma \ref{lem:tn_contract}. In the next step, we used the quantity $(\alpha_i - \alpha_i^\star)\, \big(E_{q_i}(z_i) - E_{q_i^\star}(z_i) \big)$ inside the absolute value is positive, since it equals $D_{\rm KL, 1/2}(q_i\,||\,q_i^\star)$ from \eqref{eq:tmvn_skl}. 

Substituting these bounds in \eqref{eq:cs_probit}, we get 
\be 
\Delta_n^2(q_{\blds \beta}, q_{\blds z}) \le 4 \lambda_{\max}(\Sigma^{1/2} (X'X) \Sigma^{1/2})  \, D_{\rm KL,1/2}(q_{\blds \beta} \,||\, q_{\blds \beta}^\star) \, D_{\rm KL,1/2}(q_{\blds z} \,||\, q_{\blds z}^\star). 
\ee
Thus, we have verified the symmetric KL condition \eqref{eq:Gcorr_alpha_half} with $\tilde{\gamma}_n = 2 \lambda_{\max}^{1/2}(\Sigma^{1/2} (X'X) \Sigma^{1/2})$. This proves the claimed bound on $\mbox{\rm GCorr}(\pi_n)$. 


\subsection*{Proof of Proposition \ref{prop:gauss_meanprec}}
Recall $\bar{\m Q}_{\blds \mu} = \{q_{\blds \mu} \equiv N(m,s^{-1}) \,:\, (m, s) \in \mb R \times \mb (0, \infty)\}$, and $\bar{\m Q}_{\blds \tau} = \{q_{\blds \tau} \equiv \mbox{Gamma}(n/2+a_0,b) \,:\, b \in (0, \infty)\}$. For any $q_{\blds \mu} \otimes q_{\blds \tau} \in \bar{\m Q}_{\blds \mu} \times \bar{\m Q}_{\blds \tau}$, we have
\be 
\Delta_n(q_{\blds \mu}, q_{\blds \tau})
= & - \int (q_{\blds \mu}(\mu) - q_{\blds \mu}^\star(\mu)) (q_{\blds \tau}(\tau) - q_{\blds \tau}^\star(\tau)) \,\frac{n \tau (\mu - \bar{x})^2}{2} \, d\mu \, d\tau.
\ee
Here, we made use of the identity \eqref{eq:int_survives} and that the only interaction term in $\log \pi_n(\mu, \tau)$ is $-n \tau (\mu - \bar{x})^2/2$. Using the distributional forms of $q_{\blds \mu}$ and $q_{\blds \tau}$, we can simplify 
\begin{align*}
\Delta_n(q_{\blds \mu}, q_{\blds \tau})
= & -\frac{n}{2} \bigg[(m-\bar{x})^2 + \frac{1}{s} - (m^\star-\bar{x})^2 - \frac{1}{s^\star}\bigg] \,\int \tau \, (q_{\blds \tau}(\tau) - q_{\blds \tau}^\star(\tau)) \, d\tau \\
= & \frac{n(n/2 + a_0) }{2} \, \bigg[2\bigg(\bar{x} - \frac{m + m^\star}{2}\bigg)(m - m^\star) + \bigg(\frac{1}{s^\star} - \frac{1}{s}\bigg) \bigg] \,\bigg(\frac{1}{b} - \frac{1}{b^\star}\bigg) \\
= & n(n/2 + a_0) \, \alpha_m \, (m - m^\star) \, \bigg(\frac{b^\star-b}{b b^\star} \bigg) + \frac{n(n/2 + a_0) }{2} \, \bigg( \frac{s - s^\star}{s s^\star} \bigg) \, \bigg(\frac{b^\star-b}{b b^\star} \bigg), 
\end{align*}
where $\alpha_m = \bar{x} - \frac{m + m^\star}{2}$. 
Also, from \eqref{eq:gauss_kl} and \eqref{eq:gamma_kl}, we have, 
\begin{align*}
\begin{aligned}
D_{\rm KL,1/2}(q_{\blds \mu} \,||\, q_{\blds \mu^\star}) & = \frac{(s+s^\star)(m-m^\star)^2}{4} + \frac{1}{4} \, \frac{(s-s^\star)^2}{s s^\star}, \\
D_{\rm KL,1/2}(q_{\blds \tau} \,||\, q_{\blds \tau^\star}) & = \frac{n}{2} \, \frac{(b-b^\star)^2}{b b^\star}. 
\end{aligned}
\end{align*}
Using $\sqrt{|a| + |b|} \ge (\sqrt{|a|} +\sqrt{|b|})/\sqrt{2}$, bound 
\be 
\sqrt{ D_{\rm KL,1/2}(q_{\blds \mu} \,||\, q_{\blds \mu^\star}) }  & \ge \frac{\sqrt{s^\star} |m - m^\star|}{2 \sqrt{2}} + \frac{1}{2 \sqrt{2}} \frac{|s - s^\star|}{(s s^\star)^{1/2}}, \\
\sqrt{D_{\rm KL,1/2}(q_{\blds \tau} \,||\, q_{\blds \tau^\star})} & \ge \frac{\sqrt{n}}{2} \, \frac{|b-b^\star|}{(b b^\star)^{1/2}}. 
\ee
Combining, we get 
\be \label{eq:gmp_nbyd}
\frac{ |\Delta_n(q_{\blds \mu}, q_{\blds \tau})| }{\sqrt{ D_{\rm KL,1/2}(q_{\blds \mu} \,||\, q_{\blds \mu^\star}) } \, \sqrt{D_{\rm KL,1/2}(q_{\blds \tau} \,||\, q_{\blds \tau^\star})} } = \frac{N_1 + N_2}{D_1 + D_2} \le \max \bigg\{ \frac{N_1}{D_1}, \frac{N_2}{D_2} \bigg\},
\ee
with 
\be 
\frac{N_1}{D_1} = \frac{2\sqrt{2}\, |\alpha_m| \, n(n+2a_0)}{\sqrt{n s^\star}} \, \frac{1}{\sqrt{b b^\star}}, \quad \frac{N_2}{D_2} = \frac{\sqrt{2n (n+2a_0)}}{ \sqrt{s s^\star} \sqrt{b b^\star} }.
\ee
Defining $\widebar{\m Q^\star}_{\blds \mu}(r_0) :\,= \{q_{\blds \mu} \in \bar{\m Q}_{\blds \mu} : D(q_{\blds \mu}^\star \,||\, q_{\blds \mu}) \le r_0\}$ and $\widebar{\m Q^\star}_{\blds \tau}(r_0) :\,= \{q_{\blds \tau} \in \bar{\m Q}_{\blds \tau} : D(q_{\blds \tau}^\star \,||\, q_{\blds \tau}) \le r_0\}$ as in Remark \ref{rem:restrict_init}, we now bound the quantity in \eqref{eq:gmp_nbyd} over $\widebar{\m Q^\star}_{\blds \mu}(r_0) \times \widebar{\m Q^\star}_{\blds \tau}(r_0)$ to apply Theorem \ref{thm:CAVI_2block_par_alpha_init}. To that end, we first show that for the prescribed choice of $r_0$ in Proposition \ref{prop:gauss_meanprec},
\begin{align}\label{eq:gmp_nbd}
\begin{aligned}
\widebar{\m Q^\star}_{\blds \mu}(r_0) & \subseteq \bigg\{ (m, s) : s \ge \frac{c_1}{e}, \ |m - m^\star| \le \omega \sqrt{c_1/e} \bigg\}, \\
\widebar{\m Q^\star}_{\blds \tau}(r_0) & \subseteq \big\{ b : b \ge c_3 n/e^2 \big\}. 
\end{aligned}
\end{align}
Then, it is straightforward to see that inside $\widebar{\m Q^\star}_{\blds \mu}(r_0) \times \widebar{\m Q^\star}_{\blds \tau}(r_0)$,  
\be 
\frac{N_2}{D_2} \le \frac{C}{\sqrt{n}}, \quad \frac{N_1}{D_1} \le \frac{2 \sqrt{2} \, e}{\sqrt{c_1} \, c_3} \, \bigg(1 + \frac{2a_0}{n}\bigg) \, \bigg(\frac{C}{\sqrt{n}} + \frac{1}{2} \, \sqrt{\frac{c_1}{e}} \, \omega \bigg) \le \bigg(\frac{C_1}{\sqrt{n}} + C_2\omega\bigg). 
\ee
Here, we used $|\alpha_m| \le |\bar{x} - m^\star| + |m - m^\star|/2 \le C n^{-1/2} + |m - m^\star|/2$, and substituted the upper bound for $|m-m^\star|$ along with the lower bounds for $s$ and $b$ from 
\eqref{eq:gmp_nbd}. For any $n$ large enough so that $2a_0/n < 1$, the constant $C_2$ can be chosen to be $2 \sqrt{2e}/c_3$. This proves the claimed bound on $\mbox{GCorr}(\pi_n; r_0)$.

It now only remains to establish \eqref{eq:gmp_nbd}. Recall that $2r_0 = W_0(\omega^2 n)$, where for $x > 0$, $W_0(x)$ is the unique solution to the equation $y e^y = x$. Since $y \mapsto y e^y$ is an increasing function for $y > 0$, $y e^y \vert_{y=1} = e$, and $y e^y \vert_{y=\log x} = x \log x$, it follows that for $x \ge e$, $W_0(x) \in [1, \log x]$. Thus, for $n$ large enough so that $\omega^2 n \ge e$, we have $2 r_0 \in [1, \log (\omega^2 n)]$. We have from \eqref{eq:gauss_kl} and \eqref{eq:gamma_kl} that for $q_{\blds \mu} \in \bar{\m Q}_{\blds \mu}$ and $q_{\blds \tau} \in \bar{\m Q}_{\blds \tau}$, 
\be 
D_{\rm KL}(q_{\blds \mu}^\star \,||\, q_{\blds \mu}) 
& = \frac{1}{2} \big[s/s^\star -  \log\big(s/s^\star\big) - 1\big] + \frac{s(m - m^\star)^2}{2}, \\
D_{\rm KL}(q_{\blds \tau}^\star \,||\, q_{\blds \tau}) 
& = (n/2+a_0) \big[b/b^\star - \log \big(b/b^\star\big) - 1\big].
\ee
Thus, $D_{\rm KL}(q_{\blds \mu}^\star \,||\, q_{\blds \mu}) \le r_0$ implies $s(m-m^\star)^2/2 \le r_0$ and $g(s/s^\star) \le 2r_0$, where $g(x) := x - 1 - \log x$ for $x \in (0, \infty)$. Since $g(x) \ge 0$ (with equality if and only if $x = 1$), $\lim_{x \to 0} g(x) = \lim_{x \to \infty} g(x) = \infty$, it follows that for any $c > 0$, the set $\{g \le c\}$ is an interval $[\ell_c, u_c]$ with $\ell_c < 1 < u_c$. Using $g(\ell_c) = c$, we have after some algebra that $\ell_c e^{-\ell_c} = e^{-(1+c)}$, implying $\ell_c = e^{\ell_c} e^{-(1+c)} \in (e^{-(1+c)}, e^{-c})$. Thus, $g(s/s^\star) \le 2 r_0$ implies $s/s^\star \ge e^{-(1+2r_0)}$, and therefore $s \ge (c_1 n/e) e^{-2 r_0}$. Moreover, using $2r_0 \le \log(\omega^2n)$, we can further bound $s \ge c_1/(e \omega^2)$. Next, $s(m-m^\star)2/2 \le r_0$ implies $(m-m^\star)^2 \le 2 r_0/s \le e/(c_1 n) 2 r_0 e^{2 r_0} = \omega^2 e/c_1$. Thus, we have proved that $D_{\rm KL}(q_{\blds \mu}^\star \,||\, q_{\blds \mu}) \le r_0$ implies $|m - m^\star| \le (e/c_1)^{1/2} \omega$ and $s \ge c_1/(e \omega^2) \ge c_1/e$, establishing the first part of \eqref{eq:gmp_nbd}. 

Next, $D_{\rm KL}(q_{\blds \tau}^\star \,||\, q_{\blds \tau}) \le r_0$ implies $g(b/b^\star) \le r_0/(n/2 + a_0) \le 2 r_0/n \le \log (\omega^2 n)/n < 1$. From the calculations in the previous paragraph, this implies $b/b^\star \ge e^{-2}$, or $b \ge (c_3/e^2) n$. This establishes the second part of \eqref{eq:gmp_nbd}.

\subsection*{Proof of Theorem \ref{thm:2d_bvm}}
We verify the conditions of Corollary \ref{cor:2block_par_onesided} to prove Theorem \ref{thm:2d_bvm}. To that end, we first obtain a general bound on $|\Delta_n(q_1, q_2)|$ for any $q_j \in \bar{\m Q}_j(C)$ in equation \eqref{eq:deln_bvm_finalbd}. Then, we show that if for any $t \ge 0$, $D_{\rm KL}(q_j^{(t)} \,||\, q_j^\star) \le \omega^2 n, (j = 1, 2)$, then \eqref{eq:2block_par_onesided} is satisfied with the $\gamma_n$ in \eqref{eq:gamn_bvm}.

We have $\log \pi_n(\theta) = n \ell_n(\theta) + \log \pi_1(\theta_1) + \log \pi_2(\theta_2) + \text{constant}$, and hence by \eqref{eq:int_survives}, 
\be \label{eq:deln_elln}
\Delta_n(q_1, q_2) = n \int \ell_n(\theta) (q_1(\theta_1) - q_1^\star(\theta_1))(q_2(\theta_2) - q_2^\star(\theta_2)) \,d\theta_1 d\theta_2. 
\ee
Define a centered version of the log-likelihood function (around $\widehat{\theta}_n$) as 
\be\label{eq:ln_c} 
\ell_n^c(\theta_1, \theta_2) = \ell_n(\theta_1, \theta_2) - \ell_n(\widehat{\theta}_{1n}, \theta_2) - \ell_n(\theta_1, \widehat{\theta}_{2n}) + \ell_n(\widehat{\theta}_{1n}, \widehat{\theta}_{2n}). 
\ee
Another application of \eqref{eq:int_survives} in \eqref{eq:deln_elln} gives
\be \label{eq:deln_ellnc}
\Delta_n(q_1, q_2) = n \int \ell_n^c(\theta) (q_1(\theta_1) - q_1^\star(\theta_1))(q_2(\theta_2) - q_2^\star(\theta_2)) \,d\theta_1 d\theta_2. 
\ee
Observe that $\ell_n^c$ satisfies $\frac{\partial^m \ell_n^c}{\partial \theta_j^m}(\widehat{\theta}_n) = 0$ for $j = 1, 2$ and any positive integer $m$. Also, $\frac{\partial^2 \ell_n^c}{\partial \theta_1 \partial \theta_2} = \frac{\partial^2 \ell_n}{\partial \theta_1 \partial \theta_2}$, implying 
\be \label{eq:ell_nc_der}
\frac{\partial^3 \ell_n^c}{\partial \theta_1^2 \partial \theta_2} = \frac{\partial^3 \ell_n}{\partial \theta_1^2 \partial \theta_2}, \quad \frac{\partial^3 \ell_n^c}{\partial \theta_1 \partial \theta_2^2} = \frac{\partial^3 \ell_n}{\partial \theta_1 \partial \theta_2^2}. 
\ee
Let \label{eq:til_ellnc}
\be
\tilde{\ell}_n^c(\theta) :\,= \ell_n^c(\widehat{\theta}_n) + (\theta - \widehat{\theta}_n)'\nabla\ell_n^c(\widehat{\theta}_n) + (\theta - \widehat{\theta}_n)' \nabla^2 \ell_n^c(\widehat{\theta}_n) (\theta - \widehat{\theta}_n)/2
\ee
be the second-order Taylor expansion of $\ell_n^c(\cdot)$ around $\widehat{\theta}_n$. From \eqref{eq:ell_nc_der} and the assumption that the third-order mixed partial derivatives of $\ell_n$ are bounded, we can bound the remainder $R_n(\theta) :\,= \ell_n^c(\theta) - \tilde{\ell}_n^c(\theta)$ as 
\be \label{eq:rem_bd} 
|R_n(\theta)| \le C \big[ |\theta_1 - \widehat{\theta}_{1n}|^2 \, |\theta_2 - \widehat{\theta}_{2n}| +  |\theta_1 - \widehat{\theta}_{1n}| \, |\theta_2 - \widehat{\theta}_{2n}|^2 \big]. 
\ee
Since $\frac{\partial^3 \ell_n^c}{\partial \theta_j^3}(\widehat{\theta}_n) = 0$, the terms $|\theta_j - \widehat{\theta}_{jn}|^3, (j=1,2)$ do not appear in \eqref{eq:rem_bd}, and only the cross-terms appear in the bound \eqref{eq:rem_bd}. Substituting in \eqref{eq:deln_ellnc}, we get 
\be \label{eq:deln_ellnc_bd1}
|\Delta_n(q_1, q_2)| \le n \left| \int \tilde{\ell}_n^c (q_1 - q_1^\star)(q_2 - q_2^\star) \right| + n \int |R_n| |q_1 - q_1^\star| \, |q_2 - q_2^\star|. 
\ee
We bound these terms separately, beginning with the first. Using properties of $\ell_n^c$, we can simplify to obtain $\tilde{\ell}_n^c(\theta) = -\widehat{I}_{n,12} \, (\theta_1 - \widehat{\theta}_{1n})(\theta_2 - \widehat{\theta}_{2n})$. Here, we used that the diagonal entries of $\nabla^2 \ell_n^c(\widehat{\theta}_n)$ are zero, and the off-diagonal entries equal $\frac{\partial^2 \ell_n}{\partial \theta_1 \partial \theta_2}(\widehat{\theta}_n) = -\widehat{I}_{n,12}$. We can then write
\be 
n \left| \int \tilde{\ell}_n^c (q_1 - q_1^\star)(q_2 - q_2^\star) \right| 
&= n |\widehat{I}_{n,12}| |\mb E_{q_1}(\theta_1) - \mb E_{q_1^\star}(\theta_1)| \, 
|\mb E_{q_2}(\theta_2) - \mb E_{q_2^\star}(\theta_2)| \\
&\le 2 \frac{|\widehat{I}_{n,12}|}{ \sqrt{\widehat{I}_{n,11}} \, \sqrt{\widehat{I}_{n,22}}} \, \sqrt{ D_{\rm KL}(q_1\,||\,q_1^\star)} \sqrt{ D_{\rm KL}(q_2\,||\,q_2^\star)} \\
&:\,= 2 \widehat{\rho}_n \sqrt{ D_{\rm KL}(q_1\,||\,q_1^\star)} \sqrt{ D_{\rm KL}(q_2\,||\,q_2^\star)}.\label{eq:deln_ellnc_bd1a}
\ee
Here, going from the first to the second line, we used the transport inequality in Lemma \ref{lem:tce} along with Assumption (A1) that the squared sub-Gaussian norm of $q_j^\star$ is bounded by $1/(n \widehat{I}_{n,jj})$ to bound $|\mb E_{q_j}(\theta_j) - \mb E_{q_j^\star}(\theta_j)| \le \sqrt{2 (n \widehat{I}_{n,jj})^{-1} \, D_{\rm KL}(q_j \,||\, q_j^\star)}, (j=1,2)$. 

Next, we bound the second term in the right hand side of \eqref{eq:deln_ellnc_bd1}. By \eqref{eq:rem_bd}, it amounts to bound 
\begin{align}
& n \int |q_1(\theta_1) - q_1^\star(\theta_1)| \, |q_2(\theta_2) - q_2^\star(\theta_2)|  \, |\theta_1 - \widehat{\theta}_{1n}|^2 \, |\theta_2 - \widehat{\theta}_{2n}|  \, d\theta_1 d \theta_2  \notag\\
=& n \bigg(\int |\theta_1 - \widehat{\theta}_{1n}|^2 |q_1(\theta_1) - q_1^\star(\theta_1)| d\theta_1\bigg) \, \bigg(\int |\theta_2 - \widehat{\theta}_{2n}| |q_2(\theta_2) - q_2^\star(\theta_2)| d \theta_2 \bigg), \label{eq:cross_term1}
\end{align}
and 
\begin{align} 
& n \int |q_1(\theta_1) - q_1^\star(\theta_1)| \, |q_2(\theta_2) - q_2^\star(\theta_2)|  \, |\theta_1 - \widehat{\theta}_{1n}| \, |\theta_2 - \widehat{\theta}_{2n}|^2  \, d\theta_1 d \theta_2 \notag \\
= & n \bigg(\int |\theta_1 - \widehat{\theta}_{1n}| |q_1(\theta_1) - q_1^\star(\theta_1)| d\theta_1\bigg) \, \bigg(\int |\theta_2 - \widehat{\theta}_{2n}|^2 |q_2(\theta_2) - q_2^\star(\theta_2)| d \theta_2 \bigg). \label{eq:cross_term2}
\end{align}
It now suffices to bound $\int |\theta_j - \widehat{\theta}_{jn}|^{k} |q_j(\theta_j) - q_j^\star(\theta_j)| d\theta_j$ for $k = 1, 2$. Let us gather some pieces below to that end. First, by consistency of VB, with high probability, 
\be \label{eq:vb_cons}
|\mb E_{q_j^\star}(\theta_j) - \widehat{\theta}_{jn} | \lesssim \frac{1}{\sqrt{n}}, \quad j = 1, 2. 
\ee
Next, following the discussion after \eqref{eq:deln_ellnc_bd1a}, 
\be \label{eq:trans_appl}
|\mb E_{q_j}(\theta_j) - \mb E_{q_j^\star}(\theta_j)| \lesssim \sqrt{ \frac{D_{\rm KL}(q_j \,||\, q_j^\star)}{n}}, \quad j = 1, 2. 
\ee
Using Lemma \ref{lem:subg_mom} and \eqref{eq:vb_cons}, for any $k \in \mb N$, 
\be \label{eq:subg_mom_bd}
\int |\theta_j - \widehat{\theta}_{jn}|^{2k} q_j^\star(d\theta_j) \lesssim \frac{1 + [D_{\rm KL}(q_j \,||\, q_j^\star)]^k}{n^k}, \quad j = 1, 2. 
\ee
Next, set $Q=q_j$, $P=q_j^\star$, and $Z_j=cn|\theta_j - \mb E_{q_j^\star}(\theta_j)|^2$ for some sufficiently small constant $c$ (independent of $n$) in the variational characterization of the KL divergence \eqref{eq:KL_var_char}. Note that since $q_j^\star$ is $n^{-1/2}$-sub-Gaussian, we have $\mb E_P[e^{Z_j}] \leq 2$ for some sufficiently small $c>0$. Therefore, by \eqref{eq:KL_var_char} we obtain
\begin{align*}
\mb E_Q[Z_j] = cn \int  |\theta_j - \mb E_{q_j^\star}(\theta_j)|^2 q_j(d\theta_j)  \leq D_{\rm KL}(q_j \,||\, q_j^\star)  + 1,
\end{align*}
implying
\be\label{eq:change_mes}
\int |\theta_j - \mb E_{q_j^\star}(\theta_j)|^2 q_j(d\theta_j) \leq c^{-1} n^{-1}  \big[D_{\rm KL}(q_j \,||\, q_j^\star) + 1 \big], \quad j = 1, 2. 
\ee
With these ingredients, return to bounding \eqref{eq:cross_term1} and \eqref{eq:cross_term2}. Using Cauchy--Schwarz inequality, bound, for $j = 1, 2$, 
\begin{align}
& \int |\theta_j - \widehat{\theta}_{jn}|^{k} |q_j(\theta_j) - q_j^\star(\theta_j)| d\theta_j \notag \\
= & \int |\theta_j - \widehat{\theta}_{jn}|^{k} |\surd{q_j}(\theta_j) + \surd{q_j^\star}(\theta_j)| \, |\surd{q_j}(\theta_j) - \surd{q_j^\star}(\theta_j)| d\theta_j \notag \\
\lesssim & \bigg( \int |\theta_j - \widehat{\theta}_{jn}|^{2k} (q_j+q_j^\star)(d\theta_j)\bigg)^{1/2} \, \big(h^2(q_j, q_j^\star)\big)^{1/2}. \label{eq:bvmcalcs_bd}
\end{align}
For $k = 1$, it is immediate from \eqref{eq:subg_mom_bd} and \eqref{eq:change_mes} that 
\be \label{eq:bvmcalcs_bd1}
\bigg(\int |\theta_j - \widehat{\theta}_{jn}|^2 \, (q_j + q_j^\star)(d\theta_j) \bigg)^{1/2} \lesssim \frac{1 + \sqrt{ D_{\rm KL}(q_j \,||\, q_j^\star) } }{\sqrt{n}}, \quad j = 1,2. 
\ee
Also, we show that for $k = 2$,
\be \label{eq:bvmcalcs_bd2}
\bigg(\int |\theta_j - \widehat{\theta}_{jn}|^4 \, (q_j + q_j^\star)(d\theta_j) \bigg)^{1/2}  \lesssim \frac{1 +  D_{\rm KL}(q_j \,||\, q_j^\star)}{n}, \quad j = 1, 2. 
\ee
To see this, bound using previous inequalities \eqref{eq:vb_cons} -- \eqref{eq:change_mes} and Assumption A2, 
\begin{align*}
& \int |\theta_j - \widehat{\theta}_{jn}|^4 \, (q_j + q_j^\star)(d\theta_j) \\
\lesssim &  |\mb E_{q_j}(\theta_j) - \mb E_{q_j^\star}(\theta_j)|^4 + |\mb E_{q_j^\star}(\theta_j) - \widehat{\theta}_{jn}|^4 + \int |\theta_j - \mb E_{q_j}(\theta_j)|^4 q_j(d\theta_j) + \frac{1 + [D_{\rm KL}(q_j\,||\, q_j^\star)]^2}{n^2} \\
\lesssim & \bigg(\int |\theta_j - \mb E_{q_j}(\theta_j)|^2 q_j(d\theta_j) \bigg)^2 +  \frac{1 + [D_{\rm KL}(q_j\,||\, q_j^\star)]^2}{n^2} \\
\lesssim & \bigg(|E_{q_j}(\theta_j) - \mb E_{q_j^\star}(\theta_j)|^2 + \int |\theta_j - \mb E_{q_j^\star}(\theta_j)|^2 q_j(d\theta_j) \bigg)^2 +  \frac{1 + [D_{\rm KL}(q_j\,||\, q_j^\star)]^2}{n^2} \\
\lesssim & \frac{1 + [D_{\rm KL}(q_j\,||\, q_j^\star)]^2}{n^2}. 
\end{align*}
Now, substituting \eqref{eq:bvmcalcs_bd1} and \eqref{eq:bvmcalcs_bd2} in \eqref{eq:bvmcalcs_bd} to bound the quantity in \eqref{eq:cross_term1} by 
\begin{align} 
& C n \, \frac{1 + D_{\rm KL}(q_1\,||\, q_1^\star)}{n} \, \frac{1 + \sqrt{ D_{\rm KL}(q_2 \,||\, q_2^\star) } }{\sqrt{n}} \, \big(h^2(q_1, q_1^\star)\big)^{1/2} \, \big(h^2(q_2, q_2^\star)\big)^{1/2} \notag \\
\le & C n^{-1/2} \big(2 + 2\sqrt{D_{\rm KL}(q_1 \,||\, q_1^\star)}\big) \, \sqrt{ D_{\rm KL}(q_1 \,||\, q_1^\star) } \sqrt{ D_{\rm KL}(q_2 \,||\, q_2^\star) }. \label{eq:cross_term1a}
\end{align}
Going from the first to the second line, we expanded the product into a sum of four terms and judiciously used $h^2(p, q) \le \min\{1, D_{\rm KL}(p\,||\,q)\}$. Along similar lines, we can bound the quantity in \eqref{eq:cross_term2} by 
\be 
C n^{-1/2} \big(2 + 2\sqrt{D_{\rm KL}(q_2 \,||\, q_2^\star)}\big) \, \sqrt{ D_{\rm KL}(q_1 \,||\, q_1^\star) } \sqrt{ D_{\rm KL}(q_2 \,||\, q_2^\star) }. \label{eq:cross_term2a}
\ee
Collecting the bounds \eqref{eq:deln_ellnc_bd1a}, \eqref{eq:cross_term1a} and \eqref{eq:cross_term2a}, we can finally bound, for any $q_j \in \bar{\m Q}_j(C)$, 
\be 
|\Delta_n(q_1, q_2)| 
\le 2 \widehat{\rho}_n \sqrt{ D_{\rm KL}(q_1\,||\,q_1^\star)} \sqrt{ D_{\rm KL}(q_2\,||\,q_2^\star)} + \\
\frac{2C}{\sqrt{n}} \big(2 + \sqrt{D_{\rm KL}(q_1\,||\,q_1^\star)} + \sqrt{ D_{\rm KL}(q_2\,||\,q_2^\star)} \big) \sqrt{ D_{\rm KL}(q_1\,||\,q_1^\star)} \sqrt{ D_{\rm KL}(q_2\,||\,q_2^\star)}. \label{eq:deln_bvm_finalbd}
\ee
We now verify \eqref{eq:2block_par_onesided} in Corollary \ref{cor:2block_par_onesided}. Suppose $D_{\rm KL}(q_j^{(t)} \,||\, q_j^\star) \le \omega^2 n, (j=1, 2)$. We can then bound
\begin{align*}
|\Delta_n(q_1^{(t+1)}, q_2^{(t)})| 
& \le \big(2 \widehat{\rho}_n + 4C n^{-1/2} + 2C \omega\big) \sqrt{ D_{\rm KL}(q_1^{(t+1)}\,||\,q_1^\star)} \sqrt{ D_{\rm KL}(q_2^{(t)}\,||\,q_2^\star)} \\ & \quad \quad  + 2C \omega  D_{\rm KL}(q_1^{(t+1)}\,||\,q_1^\star) \\
& \le \bigg(\widehat{\rho}_n + \frac{2C}{\sqrt{n}} + 3C \omega \bigg)\, \big[D_{\rm KL}(q_1^{(t+1)}\,||\,q_1^\star) + D_{\rm KL}(q_2^{(t)}\,||\,q_2^\star) \big]. 
\end{align*}
In the first step, we substitute $q_1 = q_1^{(t+1)}$ and $q_2 = q_2^{(t)}$ in \eqref{eq:deln_bvm_finalbd} and use $\sqrt{D_{\rm KL}(q_2^{(t)} \,||\, q_2^\star)/n} \le \omega$. In the second step, we use $\sqrt{|ab|} \le (|a|+|b|)/2$. This verifies the first inequality of \eqref{eq:2block_par_onesided}. The bound on $|\Delta_n(q_1^{(t)}, q_2^{(t+1)})|$ follows similarly from \eqref{eq:deln_bvm_finalbd}.

\subsection*{Proof of Proposition \ref{prop:exp_family_lvm}}
Given the nature of the updates, it suffices to restrict to $q_{\blds \beta} \equiv \pi_{\DY}(\cdot \mid \alpha)$ where $\alpha(d+1) = \alpha_2 + n$, and $q_{\blds z} \equiv p_{\LV}(\cdot \mid \eta)$.
Using a standard fact for KL divergences in exponential families, we have 
\be 
D_{\rm KL}( q_{\blds \beta}^\star \,||\, q_{\blds \beta}) 
& = a_{\blds \beta}(\alpha) - a_{\blds \beta}(\alpha^\star) - (\alpha - \alpha^\star)' \nabla a_{\blds \beta}(\alpha^\star), 
\ee
and 
\be 
D_{\rm KL}( q_{\blds z}^\star \,||\, q_{\blds z}) 
= \sum_{i=1}^n \big[a_{\blds z}(\eta_i) - a_{\blds z}(\eta_i^\star) - (\eta_i - \eta_i^\star)' \nabla a_{\blds z}(\eta_i^\star) \big],
\ee
From \eqref{eq:lsc_a_beta} and \eqref{eq:sc_a_z}, we have, 
\be 
D_{\rm KL}( q_{\blds \beta}^\star \,||\, q_{\blds \beta}) \ge \frac{L\|\alpha-\alpha^\star\|^2}{\min\{C,\|\alpha\|\}}, \quad D_{\rm KL}( q_{\blds z}^\star \,||\, q_{\blds z}) \ge L \sum_{i=1}^n \|\eta_i - \eta_i^\star\|^2.
\ee
Let $\m Q_{\blds \beta}(r_0) = \big\{D_{\rm KL}( q_{\blds \beta}^\star \,||\, q_{\blds \beta}) \le r_0\big\}$ and $\m Q_{\blds z}(r_0) = \big\{D_{\rm KL}( q_{\blds z}^\star \,||\, q_{\blds z}) \le r_0\big\}$ for $r_0 > 0$ be the neighborhoods defined in Theorem \ref{thm:CAVI_2block_par_alpha_init}. Also, define $\m U :\,= \big\{\alpha \,:\, \|(\alpha[d]-\alpha^\star[d])/n\| < \omega \sqrt{d}\big\}$ and $\m V :\,= \big\{\eta \,:\, \big(n^{-1} \sum_{i=1}^n \|\eta_i - \eta_i^\star\|^2\big)^{1/2} < \omega \sqrt{d} \big\}$ to be the neighborhoods from (i) and (ii) in the statement of the proposition. Now, with $r_0 = L \omega^2 d n$ and using the inequalities in the previous display, we get 
\be 
D_{\rm KL}( q_{\blds \beta}^\star \,||\, q_{\blds \beta}) \subseteq \m U, \quad D_{\rm KL}( q_{\blds z}^\star \,||\, q_{\blds z}) \subseteq \m V. 
\ee
We shall show below that 
\be \label{eq:exp_lvm_deltan_bd}
\frac{ |\Delta_n(q_{\blds \beta}, q_{\blds z})| }{ \sqrt{ D_{\rm KL, 1/2}( q_{\blds \beta} \,||\, q_{\blds \beta}^\star) } \, \sqrt{ D_{\rm KL,1/2}( q_{\blds z} \,||\, q_{\blds z}^\star) } } 
\le 2 \frac{u_\alpha u_\eta}{(\ell_\alpha \ell_\eta)^{1/2}} \, \sqrt{\frac{d}{n} \sum_{i=1}^n c^2(x_i)},
\ee
on $\m U \times \m V$. Since the KL neighborhoods are contained in $\m U$ and $\m V$ respectively, \eqref{eq:exp_lvm_deltan_bd} implies the claimed bound on $\mbox{\rm GCorr}(\pi_n)$. 

We now establish \eqref{eq:exp_lvm_deltan_bd}. Observe that we can write
\be 
\log \pi_n(\beta, z) = [\widehat{\alpha}(x, z)]' \, r(\beta) + C, \quad \widehat{\alpha}(x, z) = \bigg[\alpha_1 + \sum_{i=1}^n s(x_i, z_i); \alpha_2 + n\bigg]. 
\ee
Therefore, we have,  
\be 
\Delta_n(q_{\blds \beta}, q_{\blds z}) 
= & \int (q_{\blds \beta}(\beta) - q_{\blds \beta}^\star(\beta)) (q_{\blds z}(z) - q_{\blds z}^\star(z)) \, \widehat{\alpha}'(x, z) \, r(\beta) \\
= & \bigg[ \int r(\beta) \,  (q_{\blds \beta}(\beta) - q_{\blds \beta}^\star(\beta)) \, d\beta \bigg]' \, \bigg[ \int \widehat{\alpha}(x, z) \, (q_{\blds z}(z) - q_{\blds z}^\star(z)) \, dz \bigg] \\
= & \big[ \nabla a_{\blds \beta}(\alpha) - \nabla a_{\blds \beta}(\alpha^\star) \big]' \, \bigg[ \sum_{i=1}^n \big( \mb E_{q_i} s(x_i, z_i) - \mb E_{q_i^\star} s(x_i, z_i) \big); 0\bigg].
\ee
In the last step, we used that the expectation of the sufficient statistic in an exponential family is the gradient of the cumulant function. 
Let $\theta$ denote the vector formed by the first $d$ coordinates of $\nabla a_{\blds \beta}(\alpha) - \nabla a_{\blds \beta}(\alpha^\star)$. 
Then, the quantity in the above display equals
\be 
\sum_{i=1}^n \mb E_{q_i} [\theta' s(x_i, z_i)] - \mb E_{q_i^\star} [\theta' s(x_i, z_i)]. 
\ee
For $q_i \equiv p_{\LV}(\cdot \mid \eta_i)$ and $\theta \in \mb R^d$, we have, 
\be 
\theta' s(x_i, z_i) 
& = a_{\blds x, \blds z}(\theta) + \log p(x_i, z_i \mid \theta) \\
& = a_{\blds x, \blds z}(\theta) + \log p(x_i \mid \theta) + \log p(z_i \mid x_i, \theta) \\
& = \eta(\theta, x_i)' u(z_i) + \text{ terms only involving } x_i \text{ and } \theta. 
\ee
We then have 
\be 
\sum_{i=1}^n \mb E_{q_i} [\theta' s(x_i, z_i)] - \mb E_{q_i^\star} [\theta' s(x_i, z_i)] = \sum_{i=1}^n \eta(\theta, x_i)' [\nabla a_{\blds z}(\eta_i) - \nabla a_{\blds z}(\eta_i^\star)].
\ee
Cascading back, we have so far obtained that 
\be \label{deltan_ot}
\Delta_n(q_{\blds \beta}, q_{\blds z}) = \sum_{i=1}^n \eta(\theta, x_i)' [\nabla a_{\blds z}(\eta_i) - \nabla a_{\blds z}(\eta_i^\star)].
\ee
Using Cauchy--Schwarz inequality, now bound 
\be 
| \Delta_n(q_{\blds \beta}, q_{\blds z} |
& \le \sum_{i=1}^n \big\| \eta(\theta, x_i) \big\| \big \| \nabla a_{\blds z}(\eta_i) - \nabla a_{\blds z}(\eta_i^\star) \big \| \\
& \le \|\theta\| \, \sum_{i=1}^n c(x_i)  \big \| \nabla a_{\blds z}(\eta_i) - \nabla a_{\blds z}(\eta_i^\star) \big \|,
\ee
where in the second step, we used the assumption \eqref{eq:eta_th_x} to get $\|\eta(\theta, x_i)\| \le c(x_i) \|\theta\|$. 
We can further bound, using $\alpha(d+1) = \alpha^\star(d+1)$, 
\be 
\|\theta\| 
& \le \|\alpha - \alpha^\star\| \, \| H_{\blds \beta}(\tilde{\alpha}) \|_2
\ee
for some $\tilde{\alpha}$ on the line segment joining $\alpha$ and $\alpha^\star$, and 
\be
\big \| \nabla a_{\blds z}(\eta_i) - \nabla a_{\blds z}(\eta_i^\star) \big \| 
= \big \| \nabla^2 a_{\blds z}( \tilde{\eta}_i) (\eta_i - \eta_i^\star) \| 
\le  \| \nabla^2 a_{\blds z}( \tilde{\eta}_i) \|_2 \, \|\eta_i - \eta_i^\star\|,
\ee
for some $\tilde{\eta}_i$ in the line segment joining $\eta_i$ and $\eta_i^\star$. 
Putting everything together, we have,
\be 
| \Delta_n(q_{\blds \beta}, q_{\blds z}) | 
& \le \|\alpha - \alpha^\star\| \, \| H_{\blds \beta}(\tilde{\alpha}) \|_2 \sum_{i=1}^n c(x_i) \| \nabla^2 a_{\blds z}( \tilde{\eta}_i) \|_2 \, \|\eta_i - \eta_i^\star\| \\
& \le u_\alpha u_\eta \, (d/n) \,  \|\alpha - \alpha^\star\| \, \sum_{i=1}^n c(x_i) \, \|\eta_i - \eta_i^\star\|, \label{deltan_ot_1}
\ee
where in the second line, we used the assumptions on the Hessian matrices inside $\m U$ and $\m V$. 

Using the KL divergence formula, we have, on $\m U$,
\be 
D_{\rm KL}( q_{\blds \beta} \,||\, q_{\blds \beta}^\star) + D_{\rm KL}(q_{\blds \beta}^\star\,||\,q_{\blds \beta}) 
& = (\alpha - \alpha^\star)'  \big(\nabla a_{\blds \beta}(\alpha) - \nabla a_{\blds \beta}(\alpha^\star) \big) \\
& = (\alpha - \alpha^\star)' \nabla^2 a_{\blds \beta}(\bar{\alpha}) (\alpha - \alpha^\star) \ge \ell_\alpha \, (d/n) \, \|\alpha - \alpha^\star\|^2,
\ee
for some $\bar{\alpha}$ on the line segment joining $\alpha$ and $\alpha^\star$.
Similarly, on $\m V$,
\be 
D_{\rm KL}( q_{\blds z} \,||\, q_{\blds z}^\star)  + D_{\rm KL}(q_{\blds z}^\star \,||\, q_{\blds z} )
& = \sum_{i=1}^n (\eta_i - \eta_i^\star)'  \big( \nabla a_{\blds z}(\eta_i) - \nabla a_{\blds z}(\eta_i^\star) \big) \\
& = \sum_{i=1}^n (\eta_i - \eta_i^\star)'  \nabla^2 a_{\blds z}(\bar{\eta}_i) (\eta_i - \eta_i^\star) 
\ge \ell_\eta \sum_{i=1}^n \|\eta_i - \eta_i^\star\|^2. 
\ee
From the previous two displays and \eqref{deltan_ot_1}, we have that 
\be 
\frac{ |\Delta_n(q_{\blds \beta}, q_{\blds z})| }{ \sqrt{ D_{\rm KL, 1/2}( q_{\blds \beta} \,||\, q_{\blds \beta}^\star) } \, \sqrt{ D_{\rm KL,1/2}( q_{\blds z} \,||\, q_{\blds z}^\star) } } 
\le 2 \frac{u_\alpha u_\eta}{(\ell_\alpha \ell_\eta)^{1/2}} \, \sqrt{\frac{d}{n} \sum_{i=1}^n c^2(x_i)},
\ee
on $\m U \times \m V$, establishing \eqref{eq:exp_lvm_deltan_bd}.

\subsection*{Proof of Proposition \ref{prop:two_comp_mix}}
Recall that the updates for $z$ lie in the family $q_{\blds z}(z) = \prod_{i=1}^n q_i(z_i)$, where each $q_i$ is a two-point distribution on $\{1, 2\}$ with probabilities $(1-p_i)$ and $p_i$ respectively. Also, the update for $\mu$ is of the form $N(m, \tau^{-1})$. For any such $q_{\blds \mu}$ and $q_{\blds z}$, we can compute 
\be 
\Delta_n(q_{\blds \mu}, q_{\blds z}) = \sum_{i=1}^n (p_i - p_i^\star) \left[ z_i(m) \, (m-m^\star) + \frac{1}{2} \bigg(\frac{1}{\tau^\star} - \frac{1}{\tau}\bigg) \right],
\ee
where $z_i(m) = x_i - (m+m^\star)/2$. We also have, 
\be 
2 D_{\rm KL, 1/2}(q_\mu \,||\, q_\mu^\star) & = \frac{\tau+\tau^\star}{2} (m-m^\star)^2 + \frac{1}{2} \frac{(\tau - \tau^\star)^2}{\tau \tau^\star}, \\
2 D_{\rm KL, 1/2}(q_z \,||\, q_z^\star) & = \sum_{i=1}^n (p_i - p_i^\star) \, (\mbox{logit}(p_i) - \mbox{logit}(p_i^\star)). 
\ee
We drop the KL subscript and write $D_{1/2}$ in the rest of the proof. We first establish a preliminary bound to $\Delta_n(q_{\blds \mu}, q_{\blds z})$ in \eqref{eq:deltan_prelbd3} below. Using $(a+b)^2 \le 2 (|a|^2 + |b|^2)$ and the Cauchy--Schwarz inequality, bound 
\be 
\Delta_n^2(q_{\blds \mu}, q_{\blds z}) \le \underbrace{ 2 (m-m^\star)^2 \sum_{i=1}^n z_i^2(m) \sum_{i=1}^n (p_i - p_i^\star)^2}_{ \m N_1} + \underbrace{ \frac{ (\tau - \tau^\star)^2 }{ 2 (\tau \tau^\star)^2 } \, n \sum_{i=1}^n (p_i - p_i^\star)^2}_{\m N_2}. 
\ee
Then, we can write 
\be \label{eq:deltan_prelbd}
\frac{\Delta_n^2(q_{\blds \mu}, q_{\blds z})}{ (2 D_{1/2}(q_{\blds \mu} \,||\, q_{\blds \mu}^\star)) (2 D_{1/2}(q_{\blds z}\,||\, q_{\blds z}^\star)) } \le \frac{\m N_1 + \m N_2}{\m D_1 + \m D_2} \le \max \bigg\{ \frac{\m N_1}{\m D_1}, \frac{\m N_2}{\m D_2} \bigg\},
\ee
where 
\be 
\m D_1 & = \frac{\tau+\tau^\star}{2} (m-m^\star)^2 \, \sum_{i=1}^n (p_i - p_i^\star) \, (\mbox{logit}(p_i) - \mbox{logit}(p_i^\star)), \\
\m D_2 & = \frac{1}{2} \frac{(\tau - \tau^\star)^2}{\tau \tau^\star}  \, \sum_{i=1}^n (p_i - p_i^\star) \, (\mbox{logit}(p_i) - \mbox{logit}(p_i^\star)). 
\ee
Now, using the mean-value theorem, write $(\mbox{logit}(p_i) - \mbox{logit}(p_i^\star)) = (p_i - p_i^\star)/\{\tilde{p}_i (1 - \tilde{p}_i)\}$ for some $\tilde{p}_i$ between $p_i$ and $p_i^\star$. Substituting in the previous displays, and bounding $\tilde{p}_i (1 - \tilde{p}_i) \le \max_{k \in [n]} [\tilde{p}_k (1 - \tilde{p}_k)]$ for each $i \in [n]$, we obtain 
\be \label{eq:deltan_prelbd2}
\max \bigg\{ \frac{\m N_1}{\m D_1}, \frac{\m N_2}{\m D_2} \bigg\} \le \max \bigg\{ \frac{4 \sum_{i=1}^n z_i^2(m)}{\tau + \tau^\star}, \frac{n}{\tau \tau^\star} \bigg\} \, \max_{k \in [n]} [\tilde{p}_k (1 - \tilde{p}_k)].
\ee
Since the data-generating distribution is compactly supported, there exists $M > 0$ such that $|x_i| \le M$ for all $i \in [n]$. Inspecting the update equation for $(m, \tau)$, it is evident that for any $(m, \tau)$ in the solution path, $|m| \le \max_{k \in [n]} |x_k| \le M$ and $\tau \ge \tau_0$. The first assertion in particular implies that for any $i \in [n]$, $|z_i(m)| = |x_i - (m+m^\star)/2| \le M$. Use these to bound $(\tau + \tau^\star)^{-1} \sum_{i=1}^n z_i^2(m) \le n M^2/\tau^\star \le M^2/c_1$. Also, $n/(\tau \tau^\star) \le 1/(c_1 \tau_0)$. Thus, from \eqref{eq:deltan_prelbd} \& \eqref{eq:deltan_prelbd2}, we have 
\be \label{eq:deltan_prelbd3}
\frac{\Delta_n^2(q_{\blds \mu}, q_{\blds z})}{ D_{1/2}(q_{\blds \mu} \,||\, q_{\blds \mu}^\star) \, D_{1/2}(q_{\blds z}\ \,||\, q_{\blds z}^\star) } \le 4 \zeta^2 \, \max_{k \in [n]} [\tilde{p}_k (1 - \tilde{p}_k)],
\ee
where $\zeta^2 :\,= \max\big\{4M^2/c_1, 1/(\tau_0 c_1)\big\}$. 

We now turn to the main proof, which we prove by induction. Suppose for some $t \ge 1$, $D_{1/2}(q_{\blds \mu}^{(t)}\,||\,q_{\blds \mu}^\star) \le \omega^2 n$. With $\gamma^2 = 4 \zeta^2 b \delta$ for some constant $b$ and $\delta$ as in the statement of the proposition, we will show using the induction hypothesis that
\be \label{eq:pcontr1}
|\Delta_n(q_{\blds \mu}^{(t)}, q_{\blds z}^{(t+1)})| \le \gamma \sqrt{ D_{1/2}(q_{\blds \mu}^{(t)}\,||\,q_{\blds \mu}^\star) } \, \sqrt{ D_{1/2}(q_{\blds z}^{(t+1)}\,||\,q_{\blds z}^\star) }.
\ee
Then, from our general analysis (proof of Theorem 2), it follows that 
\be \label{eq:contr1}
D_{1/2}(q_{\blds z}^{(t+1)}\,||\,q_{\blds z}^\star) \le \frac{\gamma^2}{4} D_{1/2}(q_{\blds \mu}^{(t)}\,||\,q_{\blds \mu}^\star).
\ee
Next, using the trivial bound $p(1-p) \le 1/4$ for any $p \in (0, 1)$ in \eqref{eq:deltan_prelbd3}, we immediately have
\be \label{eq:pcontr2}
|\Delta_n(q_{\blds \mu}^{(t+1)}, q_{\blds z}^{(t+1)})| \le \zeta \sqrt{ D_{1/2}(q_{\blds \mu}^{(t+1)}\,||\,q_{\blds \mu}^\star) } \, \sqrt{ D_{1/2}(q_{\blds z}^{(t+1)}\,||\,q_{\blds z}^\star) }. 
\ee
Again our general analysis (proof of Theorem 2) gives us 
\be \label{eq:contr2}
D_{1/2}(q_{\blds \mu}^{(t+1)}\,||\,q_{\blds \mu}^\star) \le \frac{\zeta^2}{4} D_{1/2}(q_{\blds z}^{(t+1)}\,||\,q_{\blds z}^\star).
\ee
This in itself need not be a contraction, as $\zeta$ could be a large constant. However, combining \eqref{eq:contr1} and \eqref{eq:contr2}, we obtain 
\be 
D_{1/2}(q_{\blds \mu}^{(t+1)}\,||\,q_{\blds \mu}^\star) \le \kappa D_{1/2}(q_{\blds \mu}^{(t)}\,||\,q_{\blds \mu}^\star)
\ee 
with $\kappa = (\zeta^4 b/4) \, \delta \ll 1$ for $\delta$ sufficiently small. This in particular implies $D_{1/2}(q_{\blds \mu}^{(t+1)}\,||\,q_{\blds \mu}^\star) \le \omega^2 n$ and hence the induction hypothesis can be continued. 

It only remains to establish \eqref{eq:pcontr1}. From \eqref{eq:deltan_prelbd3}, we have 
\be \label{eq:gamma_zeta}
\gamma^2 = 4 \zeta^2 \max_{k \in [n]} [\tilde{p}_k (1 - \tilde{p}_k)],
\ee
where $\tilde{p}_k$ lies between $p_k^{(t+1)}$ and $p_k^\star$. We now obtain a refined bound on $\max_{k \in [n]} [\tilde{p}_k (1 - \tilde{p}_k)]$ using the induction hypothesis. The condition $D_{1/2}(q_{\blds \mu}^{(t)}\,||\,q_{\blds \mu}^\star) \le \omega^2 n$ implies $|m^{(t)} - m^\star| \le 2 \omega (n/\tau^\star)^{1/2} \le c_1' \omega$, where $c_1' = 2/\sqrt{c_1}$. Now, for any $i \in [n]$, 
\be 
\left| \mbox{logit}(p_i^{(t+1)}) - \mbox{logit}(p_i^\star) \right| 
& = \left | \mbox{logit}\big(p_i(m^{(t)}, \tau^{(t)})\big) - \mbox{logit}\big(p_i(m^\star, \tau^\star)\big) \right| \\
& = \left | (m^{(t)} - m^\star) z_i(m^{(t)}) - \frac{1}{2} \bigg(\frac{1}{\tau^{(t)}} - \frac{1}{\tau^\star}\bigg) \right| \\
& \le |m^{(t)} - m^\star| \, |z_i(m^{(t)})| + \frac{1}{2 \tau_0} \\
& \le c_1' \omega M +  \frac{1}{2 \tau_0} :\, = c_2.
\ee
Here, we used the global bound on $|z_i(m)|$ from earlier, and used that any $\tau$ on the solution path is $\ge \tau_0$. 

Now, we invoke the assumption on the $p_i^\star$s. Suppose $i$ is such that $p_i^\star \ge (1 - \delta)$. Then, $\mbox{logit}(p_i^\star) \ge \log \{(1-\delta)/\delta\}$. From the bound in the previous display, we obtain $\mbox{logit}(p_i^{(t+1)}) \ge \log \{(1-\delta)/\delta\} - c_2 \ge \log(\{1/(b \delta)\}$ for some $b > 1$. Since $\mbox{logit}$ is an increasing function and $\tilde{p}_i$ lies between $p_i^{(t+1)}$ and $p_i^\star$, we conclude that $\mbox{logit}(\tilde{p}_i) \ge \log(\{1/(b \delta)\}$, implying $\tilde{p}_i \ge 1/(b \delta)/\{1 + 1/(b \delta)\}$. This implies $\tilde{p}_i (1 - \tilde{p}_i) \le 1/\{1 + 1/(b \delta)\} \le b \delta$. The case when $p_i^\star \le \delta$ is similarly handled by noting that $|\mbox{logit}(p_1) - \mbox{logit}(p_2)| = |\mbox{logit}(1 - p_1) - \mbox{logit}(1 - p_2)|$. Thus, $\max_{k \in [n]} [\tilde{p}_k (1 - \tilde{p}_k)] \le b \delta$ in \eqref{eq:gamma_zeta}. This proves the claim in \eqref{eq:pcontr1} and completes the proof. 

\section{Additional proofs and derivations}\label{sec:pf_app_aux}
\subsection*{Proof of Lemma \ref{lem:tn_contract}}
A straightforward calculation shows
\be \label{eq:tn_mean}
E_{X \sim TN_1(a,1)}(X) = a + H(-a), \quad E_{X \sim TN_0(a,1)}(X)(X) = a - H(a),
\ee
where $H(t) = \phi(t)/(1-\Phi(t))$ for $t \in \mb R$ is the standard normal Hazard function. We recall some of its standard properties. $H$ is an increasing function with $\lim_{t \to -\infty} H(t) = 0$ and $\lim_{t \to \infty} H(t) = 1$. 
For any $t \in \mb R$, we have $H'(t) = H(t)(H(t)-t)$. $H'$ is an increasing function, with $\lim_{t \to -\infty} H'(t) = 0$ and $\lim_{t \to \infty} H'(t) = 1$. In particular, $H'(t) \in (0, 1)$ for all $t \in \mb R$. 

We first prove the lemma for the case $y = 1$. Without loss of generality, assume $a_1 > a_0$. Then, 
\be 
\frac{|b_1 - b_0|}{|a_1 - a_0|} 
& = \left \vert \frac{(a_1 - a_0) - \{H(-a_0) - H(-a_1)\}}{a_1-a_0} \right\vert \\
& = \left \vert 1 - \frac{ \{H(-a_0) - H(-a_1)\} }{\{(-a_0) - (-a_1)\}} \right \vert \\
& = |1 - H'(\xi)| 
\ee
for some $\xi \in (-a_1, -a_0)$. Since $H'(t) \in (0, 1)$ for all $t$, it follows that $|b_1 - b_0| \le |a_1 - a_0|$. 

We next consider the case $y = 0$. Again, without loss of generality, assume $a_1 > a_0$. Then, 
\be 
\frac{|b_1 - b_0|}{|a_1 - a_0|} 
& = \left \vert \frac{(a_1 - a_0) - \{H(a_1) - H(a_0)\}}{a_1-a_0} \right\vert \\
& = |1 - H'(\xi)| 
\ee
for some $\xi \in (a_0, a_1)$. The conclusion again follows using $H'(t) \in (0, 1)$ for all $t$. 

\subsection*{Proof of Lemma \ref{lem:subg_mom}}
Bound $\mb E |Z-m|^k \le C \left[ \mb E |Z - \mb E [Z]|^k + |\mb E[Z] - m|^k \right]$. Using an equivalent notion of sub-Gaussianity (see \cite[Definition 5.7]{vershynin2010introduction}), we have $\mb E |Z - \mb E [Z]|^k \lesssim \|Z\|_{\psi_2}^k \le \nu^k$. 

\subsection*{Derivations related to functional derivatives}
We first establish the identity in \eqref{eq:KL_frechet}. Fix $q \in \m Q$ and $h \in T_{\m Q}(q)$. We have, for scalar $t$,
\be 
F(q + th) - F(q) = \int (q + th) \log (q + th) - \int q \log q  - t \int h \log \pi_n. 
\ee
Thus, 
\be 
\frac{ F(q + th) - F(q) }{t} 
& = \frac{1}{t} \int q \log \left(1 + t \, \frac{h}{q} \right) + \int h \frac{\log(q + th)}{\pi_n} \\
& = \int q \, \frac{h}{q} \, \frac{\log (1 + t \, \frac{h}{q})}{t \, \frac{h}{q}} + \int h \frac{\log (q + th)}{\pi_n}. 
\ee
Taking the limit as $t \to 0$, and interchanging the limit and the integrals using DCT, we obtain 
\be
\lim_{t \to 0} \frac{ F(q + th) - F(q) }{t} = \int h + \int h \log \frac{q}{\pi_n} = \langle h, \log \frac{q}{\pi_n} \rangle
\ee
where at the last step we use $\int h = 0$ since $h \in T_{\m Q}(q)$. 
\\[3ex]
Next, we verify \eqref{eq:partial_ders_d}, which automatically implies \eqref{eq:partial_ders}. Recall $G_j(q_j) = F(q_j \otimes q_{-j})$. Fix $h \in T_{\m Q_j}(q_j)$. Call $\tilde{q} = (q_j + th) \otimes q_{-j}$ and $q = q_j \otimes q_{-j}$ to write
\be 
G_j(q_j + th) - G_j(q_j) 
& = \left( \int_{\m X} \tilde{q} \log \tilde{q} - \int_{\m X} q \log q \right) -  \int_{\m X} (\tilde{q} - q) \log \pi_n \\
& = \left( \int_{\m X} \tilde{q} \log \tilde{q} - \int_{\m X} q \log q \right) - t \int_{\m X} (h \otimes q_{-j}) \, \log \pi_n
\ee
Simple algebra yields
\be 
\tilde{q} \log \tilde{q} - q \log q = (q_j \otimes q_{-j}) \, \log\left(1 + t \, \frac{h}{q_j} \right) + t (h \otimes q_{-j}) \, \log(q_j + th) + t (h \otimes q_{-j}) \, \log q_{-j}. 
\ee
Substituting in the previous display, dividing both sides by $t$, and proceeding as in the previous case to interchange limits and integrals, one gets 
\be 
\lim_{t \to 0} \frac{ G_j(q_j + th) - G_j(q_j) }{t} 
& = \int_{\m X} h \, q_{-j} + \int_{\m X} h \, q_{-j} \, \log \left( \frac{q_j \otimes q_{-j}}{\pi_n} \right) \\
& = \int_{\m X} h \, (q_{-j} + q_{-j} \log q_{-j}) + \int_{\m X} h \, q_{-j} \, \log q_j - \int_{\m X} h \, q_{-j} \, \log \pi_n \\
& = \int_{\m X_j} h \log q_j - \int_{\m X_j} h \left(\int_{\m X_{-j}} q_{-j} \log \pi_n \right) \\
& = \int_{\m X_j} h \, \left( \log q_j - \int_{\m X_{-j}} q_{-j} \log \pi_n \right),
\ee
which proves the identity. Going from the second to the third line, the first integral vanishes since it splits into a product of two terms, one of which is $\int_{\m X_j} h = 0$. 
\\[3ex]
Finally, we establish the identity in \eqref{eq:deljF_ip}. Using the expression for $\partial_j F(q_j \otimes q_{-j})$ from \eqref{eq:partial_ders_d}, the left hand side of \eqref{eq:deljF_ip} 
\be 
\left \langle \partial_j F(q_j^{(t+1)} \otimes q_{-j}^{(t)}),  (q_j - q_j^{(t+1)}) \right \rangle = \int_{\m X_j} (q_j - q_j^{(t+1)}) \log q_j - \int_{\m X} (q_j \otimes q_{-j}^{(t)}) \log \pi_n + \int (q_j^{(t+1)} \otimes q_{-j}^{(t)}) \log \pi_n. 
\ee
Note also the right hand side of \eqref{eq:deljF_ip} can be written as $ 
F(q) - F(\tilde{q}) - D_{\rm KL}(q \,||\, \tilde{q})$ with $q = q_j \otimes q_{-j}^{(t)}$ and $\tilde{q} = q_j^{(t+1)} \otimes q_{-j}^{(t)}$. Using \eqref{eq:conv_F}, this further equals 
\be 
\langle \dot{F}(\tilde{q}), q-\tilde{q} \rangle 
& = \int_{\m X} (q - \tilde{q}) \log \left(\frac{\tilde{q}}{\pi_n} \right) \\
& = \int_{\m X} (q - \tilde{q}) \log \tilde{q} - \int_{\m X} q \log \pi_n + \int_{\m X} \tilde{q} \log \pi_n.
\ee
The proof is completed by noting that 
\be 
\int_{\m X} (q - \tilde{q}) \log \tilde{q} 
& = \int_{\m X} (q - \tilde{q}) \log q_j^{(t+1)} + \sum_{k \ne j} \int_{\m X} (q - \tilde{q}) \log q_k^{(t)} \\
& = \int_{\m X} (q_j - q_j^{(t+1)}) \otimes q_{-j}^{(t)} \, \log q_j^{(t+1)} + \sum_{k \ne j} \int_{\m X} (q_j - q_j^{(t+1)}) \otimes q_{-j}^{(t)} \log q_k^{(t)} \\
& = \int_{\m X_j} (q_j - q_j^{(t+1)}) \log q_j^{(t+1)} + \sum_{k \ne j} \left(\int_{\m X_j} (q_j - q_j^{(t+1)})\right) \, \left( \int_{\m X_k} q_k^{(t)} \log q_k^{(t)} \right) \\
& =  \int_{\m X_j} (q_j - q_j^{(t+1)}) \log q_j^{(t+1)} .
\ee

\end{document}